\documentclass[lettersize,journal]{IEEEtran}
\usepackage{amsmath,amsfonts}
\usepackage{amssymb}
\usepackage{algorithmic}
\usepackage{algorithm}
\usepackage{array}
\usepackage[utf8]{inputenc}
\usepackage[small]{caption}
\usepackage{subcaption}
\usepackage{textcomp}
\usepackage{stfloats}
\usepackage{xcolor}
\usepackage{soul}
\usepackage{cite}
\usepackage{url}
\usepackage{verbatim}
\usepackage{graphicx}
\usepackage{times}
\usepackage{graphicx}
\usepackage{booktabs}
\usepackage{tabularx}
\usepackage[switch]{lineno}
\usepackage{multirow}

\usepackage[colorlinks=true, linkcolor=blue, citecolor=blue, urlcolor=blue, pdfborder={0 0 0}]{hyperref}

\newcommand{\newtext}[1]{\textcolor{black}{#1}}

\makeatletter
\renewcommand\p@subfigure{\thefigure~} 
\makeatother

\newcommand{\Tr}{\operatorname{Tr}}
\newcommand{\Sum}{\operatorname{sum}}
\newcommand{\ve}{\operatorname{vec}}
\newcommand{\vol}{\operatorname{vol}}
\newtheorem{definition}{Definition}

\begin{document}

\title{SEHFS: Structural Entropy-Guided High-Order Correlation Learning for Multi-View Multi-Label Feature Selection}

\author{
Cheng~Peng,
Yonghao~Li\textsuperscript{*},
Wanfu~Gao\textsuperscript{*},~\IEEEmembership{Senior Member,~IEEE,}
Jie~Wen,~\IEEEmembership{Senior Member,~IEEE,}%
~and~Weiping~Ding,~\IEEEmembership{Senior Member,~IEEE}%
\thanks{Received 30 January 2026; revised 9 June 2026; accepted 30 September 2026. This work was supported in part by the National Natural Science Foundation of China under Grant 62606412 and Grant U2433216, in part by the Science Foundation of Jilin Province of China under Grant YDZJ202501ZYTS286, in part by the Natural Science Foundation of Sichuan Province under Grant 2025ZNSFSC1497. The associate eaitor coor-dinating the review of this article and approving it for publication was Prof.Wanaing Li. \textit{(Corresponding authors: Yonghao Li; Wanfu Gao.)}}%
\thanks{Cheng Peng is with the College of Software, Jilin University, Changchun 130012, China; Wanfu Gao is with the College of Computer Science and Technology, Jilin University, Changchun 130012, China. Both authors are also with the Key Laboratory of Symbolic Computation and Knowledge Engineering of the Ministry of Education, Changchun 130012, China (e-mail: pengcheng5523@mails.jlu.edu.cn; gaowf@jlu.edu.cn).}%
\thanks{Yonghao Li is with the School of Computing and Artificial Intelligence, Southwestern University of Finance and Economics, Chengdu 611130, China, and also with the Engineering Research Center of Intelligent Finance, Ministry of Education, Chengdu 611130, China (e-mail: liyonghao@swufe.edu.cn).}%
\thanks{Jie Wen is with the Shenzhen Key Laboratory of Visual Object Detection and Recognition, Harbin Institute of Technology, Shenzhen 518055, China (e-mail: jiewen\_pr@126.com)}%
\thanks{Weiping Ding is with the School of Artificial Intelligence and Computer Science, Nantong University, Nantong, 226019, China. (e-mail: dwp9988@163.com)}%
}

\markboth{IEEE TRANSACTIONS ON IMAGE PROCESSING,~Vol.~35,~2026}%
{Shell \MakeLowercase{\textit{et al.}}: A Sample Article Using IEEEtran.cls for IEEE Journals}


\maketitle

\begin{abstract}
In recent years, multi-view multi-label learning (MVML) has attracted extensive attention due to its close alignment with real-world scenarios. Information-theoretic methods have gained prominence for learning nonlinear correlations. However, two key challenges persist: first, features in real-world data commonly exhibit high-order structural correlations, but existing information-theoretic methods struggle to learn such correlations; second, commonly relying on heuristic optimization, information-theoretic methods are prone to converging to local optima. To address these two challenges, we propose a novel method called Structural Entropy Guided High-Order Correlation Learning for Multi-View Multi-Label Feature Selection (SEHFS). The core idea of SEHFS is to convert the feature graph into a structural-entropy-minimizing encoding tree, quantifying the information cost of high-order dependencies and thus learning high-order feature correlations beyond pairwise correlations. Specifically, features exhibiting strong high-order redundancy are grouped into a single cluster within the encoding tree, while inter-cluster feature correlations are minimized, thereby eliminating redundancy both within and across clusters. Furthermore, a new framework based on the fusion of information theory and matrix methods is adopted, which learns a shared semantic matrix and view-specific contribution matrices to reconstruct a global view matrix, thereby enhancing the information-theoretic method and balancing the global and local optimization. The ability of structural entropy to learn high-order correlations is theoretically established, and both experiments on eight datasets from various domains and ablation studies demonstrate that SEHFS achieves superior performance in feature selection.
\end{abstract}


\section{Introduction}
Multi-label feature selection has been widely applied in image processing, such as medical image recognition~\cite{huang2024label} and image classification~\cite{aksoy2022multi}. Recently, with the increasing prevalence of multi-modal data collection, multi-label feature selection is progressively applied in multi-view scenarios~\cite{yu2025review}. The introduction of multiple views provides more comprehensive and enriched feature semantics, but also introduces additional complexity for feature selection in MVML.

In multi-view scenarios, the core is to explore the relationships among data to balance consistency and complementarity~\cite{xie2024feature}. Data from different views often exhibit shared structure and yield a consensus that captures the most salient information~\cite{zhu2020global}. However, due to inter-view heterogeneity, each view contributes distinctive information, necessitating explicit consideration of complementarity~\cite{lin2022hierarchical}.

A major strand of consistency-focused methods aims to identify commonality across views~\cite{zhang2018latent,liu2023label}. RRFS~\cite{han2024feature} measures both inter-view and intra-view information overlap through a view redundancy coefficient, and preserves shared information across views based on information gain. In contrast, research on complementarity emphasizes the differing importance of each view~\cite{lyu2022beyond,hao2025embedded}. I$^2$VSLC~\cite{hao2024exploring} utilizes the graph Laplacian to capture the local geometric structure within each view, thereby preserving view-specific characteristics.

However, most existing methods adopt linear matrix operations as the final objective function, which makes it difficult for models to capture nonlinear relationships in the space, resulting in weak generalization ability. Moreover, these methods lack effective measures for feature selection~\cite{fan2024learning,dai2025multi}, making it challenging to directly quantify feature correlations and redundancy. To address this, some recent methods begin to use information-theoretic methods to select important features.

Most methods utilize mutual information to explore relationships within data and measure nonlinear correlations and redundancy~\cite{tan2024two,ma2025mi}, but these methods only capture pairwise and low-order relationships, resulting in limited effectiveness and high computational cost, which are further exacerbated in multi-view scenarios. FSLSFE~\cite{yang2025robust} first obtains the overall fuzzy entropy of the label space through label augmentation, and then determines feature importance by evaluating the change in fuzzy entropy before and after adding each feature. However, information-theoretic methods typically rely on heuristic search strategies for feature selection~\cite{gao2023unified}, which can easily lead to local optima.

\begin{figure}[ht]
    \centering
    \includegraphics[width=0.47\textwidth]{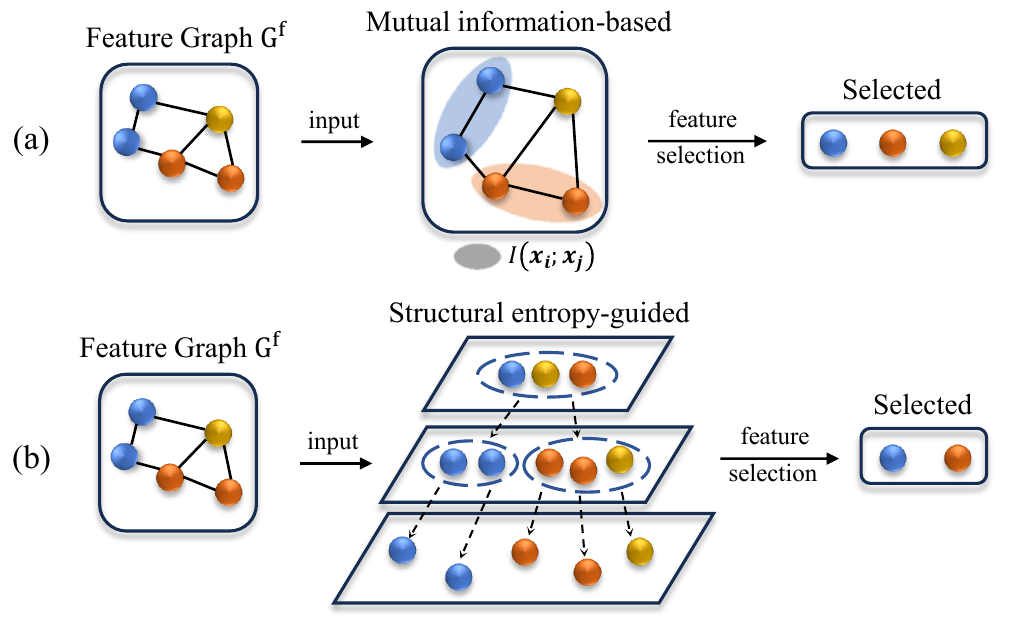}
    \caption{Concept maps of two types of information-theoretic methods. Each ball denotes a feature, and greater color similarity indicates higher correlation. The yellow and red balls show high-order correlation. (a) is existing methods based on mutual information, which are limited to learning low-order correlations; (b) is SEHFS guided by structural entropy, which learns high-order correlations.}
    \label{fig:moti}
\end{figure}
As illustrated in Figure~\ref{fig:moti}, existing information-theoretic methods are predominantly mutual information–based and thus largely restricted to learning low-order feature correlations. Moreover, because they commonly rely on heuristic optimization, they are prone to converging to local optima. To address these two challenges, we propose \textbf{S}tructural \textbf{E}ntropy Guided \textbf{H}igh-Order Correlation Learning for Multi-View Multi-Label \textbf{F}eature \textbf{S}election (SEHFS). The core idea of SEHFS is to convert the feature graph into a structural-entropy–minimizing encoding tree~\cite{li2016structural}, thereby quantifying the information cost of high-order dependencies and enabling the learning of high-order feature correlations beyond pairwise correlations. Concretely, the encoding tree aggregates features characterized by strong high-order redundancy into one cluster and attenuates cross-cluster feature correlations, thus eliminating redundancy inside and between clusters. In addition, we adopt a fusion framework that integrates information theory with matrix methods: a shared semantic matrix and view-specific contribution matrices are learned to reconstruct a global view matrix, which strengthens the information-theoretic modeling and balances global and local optimization. The main contributions of the proposed method are as follows:

\begin{itemize}
    \item A structural entropy-guided feature selection method based on minimization is proposed, which learns high-order feature correlations and eliminates redundancy, thereby addressing the challenge that existing information-theoretic methods typically learn only low-order correlations and achieve superior performance in feature selection.
    \item A novel information-matrix fusion framework is introduced, which reconstructs a global view matrix to balance multi-view consistency and complementarity, while mitigating the challenge that information-theoretic methods are prone to converging to local optima.
    \item An effective solution to the alternative optimization problem of SEHFS is proposed. Experimental results on eight cross-domain datasets demonstrate that the proposed method outperforms seven state-of-the-art methods.
\end{itemize}

The structure of this paper is as follows: Section \ref{sec:related} reviews the related works from previous studies. Section \ref{sec:proposed} provides a detailed explanation and optimization strategies of SEHFS. Section \ref{sec:experiments} presents an analysis of the experimental results. Finally, Section \ref{sec:conclusion} concludes the paper with a comprehensive summary.

\section{Related Work}
\label{sec:related}
\subsection{Multi-label Feature Selection}

In multi-label scenarios, each sample is associated with multiple labels, which makes the label space semantically richer~\cite{bogatinovski2022comprehensive,peng2026dual}. Research on multi-label feature selection (MLFS) achieves significant progress and mainly follows three directions:

First, some methods are based on manifold learning~\cite{yin2023multi,wang2025multi}. They use manifold learning to capture the underlying geometric structures of the feature matrix or label matrix, and perform feature selection via sparse regularization. MIFS~\cite{jian2016multi} decomposes multi-label information into a low-dimensional latent matrix and integrates manifold constraints reflecting local geometry with $\ell_{2,1}$-norm. MSSL~\cite{cai2018multi} incorporates feature manifold learning via graph regularization and $\ell_{2,1}$-norm sparsity into a least-squares regression model, obtaining sparse regression coefficients through iterative optimization. SPLDG~\cite{zhang2025sparse} uses pseudo-label learning to mitigate the incompatibility between true labels and linear mapping, and alternately optimizes feature weights using sparse regularization, dynamic graphs, and manifold learning.

Second, an innovative fusion of sparse norms has been proposed to overcome the limitations of conventional sparse regularization in multi-label settings~\cite{li2023multi}. SRFS~\cite{li2025fusion} fuses inner product regularization and the $\ell_{2,1}$-norm to construct a sparse complementary norm, which avoids LASSO’s tendency to eliminate useful zero-weight features and reduces feature redundancy caused by the $\ell_{2,1}$-norm.

Third, some methods utilize information-theoretic methods to evaluate features~\cite{qian2020mutual,dai2025noise}. SRLG-LMA~\cite{dai2024multi} utilizes fuzzy conditional mutual information to select optimal features based on strong relevant label gain. However, these methods rely on mutual information as the evaluation measure and can only capture pairwise low-order relationships.

There are many similarities between MLFS and multi-view multi-label feature selection (MVMLFS). The primary difference lies in whether multiple views are incorporated. The first two types of MLFS methods can be directly applied to multi-view scenarios by simply merging multiple views into a single view~\cite{zhang2023sparse}, the third type of MLFS methods, due to the high computational cost of mutual information calculations, is difficult to apply to large-scale MVML data~\cite{vinh2012novel}.

\subsection{Multi-view Multi-label Learning}
In multi-view scenarios, data come from different views that present diverse semantics, which increases both data diversity and scale~\cite{wu2019multi}. Existing MVML methods strive to balance consistency and complementarity to preserve comprehensive information. Some methods use reconstruction for view fusion~\cite{jiang2025global,hao2025tensor}. DHLI~\cite{hao2024double} focuses on reconstructing the label matrix, decomposing labels into dual-layer hybrid labels, and designing a regularization paradigm incorporating logical operations to establish the mapping between hybrid labels and features. GRAFS~\cite{hao2024anchor} focuses on reconstructing the global view, enhancing the completeness of the global view through anchors and unique contributions, and establishing the mapping from the global view to the label matrix. In addition, embedding techniques are employed to map data into a low-dimensional latent matrix, enabling the learning of salient information~\cite{zhu2018multi,lu2023distance}. MSFS~\cite{zhang2020multi} decomposes multi-label information into low-dimensional latent label representations to learn label correlations, constructs local geometric structures within each view to capture intra-view similarity, and applies sparse regularization to achieve mapping.

However, when facing more complex and large-scale data, these methods lack generalization ability in feature selection and struggle to learn high-order feature correlations~\cite{sanghavi2022multi,hu2023deep}.

\begin{figure*}[ht]
    \centering
    \includegraphics[width=\textwidth]{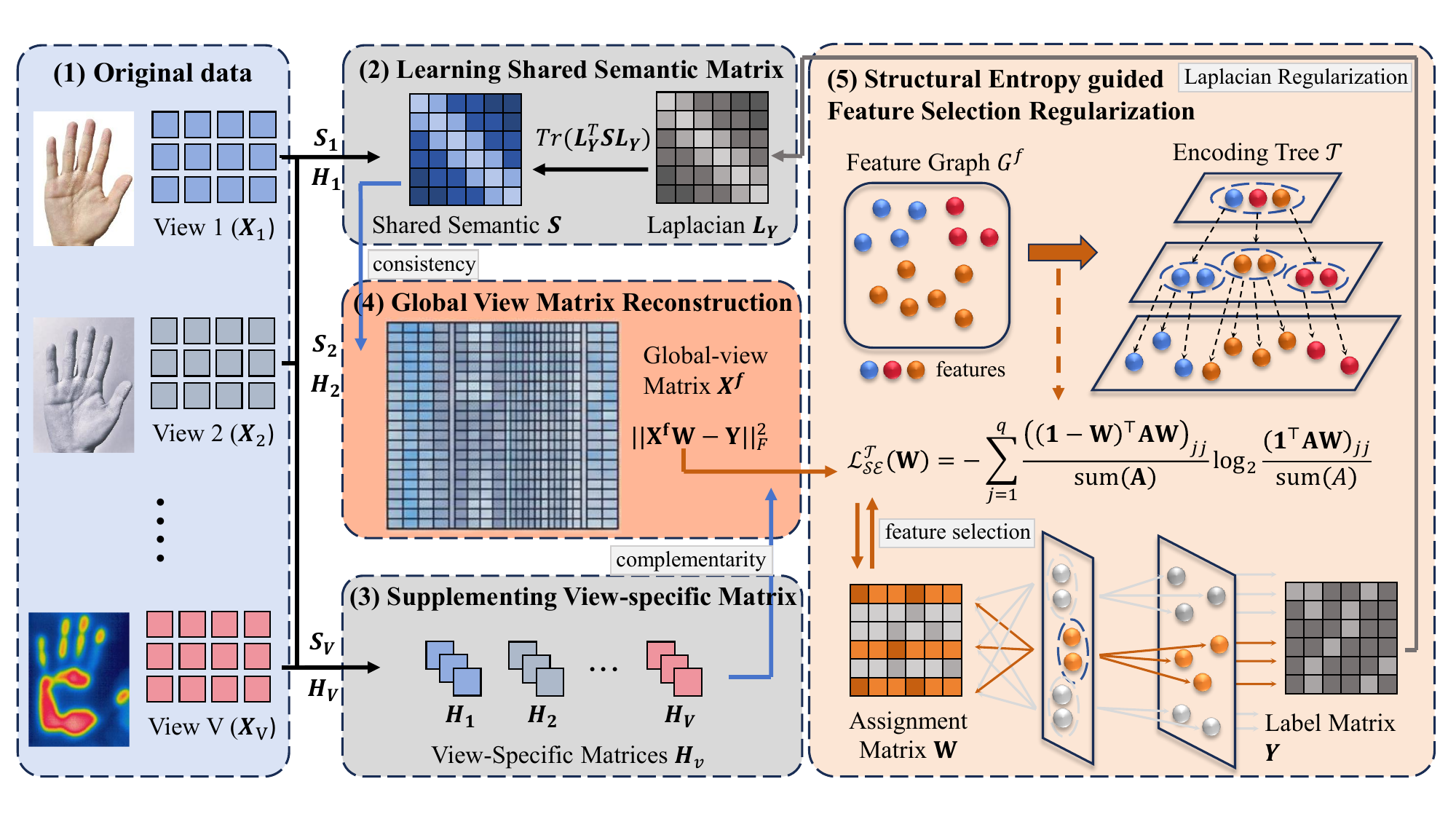}
    \caption{The framework of SEHFS. Given a multi-view multi-label dataset $[\mathbf{X}_1,\mathbf{X}_2,\dots,\mathbf{X}_V]$, SEHFS first learns the semantic matrix $\mathbf{S}_v$ and view-specific matrix $\mathbf{H}_v$ from the $v$-th view $\mathbf{X}_v$. Then, semantic matrices from all views are integrated to form a shared semantic matrix $\mathbf{S}$, whose graph structure is regularized by a graph Laplacian $\mathbf{L_Y}$ learned from the label matrix $\mathbf{Y}$. \newtext{Here, $\mathbf{S}$ captures cross-view consensus information, whereas $\mathbf{H}_v$ compensates for view-specific complementary information that is not fully preserved in the shared space. Their joint effect reconstructs the global view matrix $\mathbf{X^f}$}, from which a global feature graph $G^f$ is derived. Minimizing the structural entropy of $G^f$ yields an optimal encoding tree $\mathcal{T}$ that learns higher-order correlations and removes redundancy. Mapping $\mathbf{Y}$ onto $\mathcal{T}$ produces a more effective, low-redundancy feature selection.}
    \label{fig:framework}
\end{figure*}
\section{The Proposed Method}
\label{sec:proposed}
Given a multi-view multi-label dataset containing $n$ samples and $V$ views $(\mathbf{X}, \mathbf{Y})$, the feature matrix $\mathbf{X}$ consists of $[\mathbf{X}_1,\mathbf{X}_2,\dots,\mathbf{X}_v;\cdots;\mathbf{X}_V]$, where $v$ indexes the views ($v \in \{1,\cdots,V\}$), $d(v)$ represents the feature dimension of $v$-th view, and $\mathbf{X}_v \in \mathbb{R}^{n\times d(v)}$ denotes each instance for $v$-th view. The label matrix $\mathbf{Y} = [y_1,y_2,\dots,y_q]\in \{0,1\}^{n\times q}$ is denoted for all views, where $q$ is the dimension of label vector. In the label matrix, $\mathbf{Y}_{i,j} = 1$ means that the $j$-th label is the label of $i$-th sample. In this section, SEHFS is introduced with its framework shown in Figure~\ref{fig:framework}, which consists of two main components: (1) structural entropy regularization is used to learn high-order feature correlations, aiming to eliminate redundancy and achieve feature selection; (2) a global view matrix is reconstructed through matrix operations, aiming to balance global and local optimization.

\subsection{Structural Entropy-Guided Feature Selection Regularization}
Most existing information-theoretic methods use mutual information to compute correlations either between features or between features and labels. However, mutual information is fundamentally limited to pairwise correlations and can only learn low-order feature correlations. SEHFS utilizes the information-theoretic concept of structural entropy via matrix operations, introducing a structural entropy regularization term that can convert the feature graph into a structural-entropy–minimizing encoding tree and thus learn high-order feature correlations and eliminate redundancy among features.

The definition of encoding tree and structural entropy are presented as follows:
\begin{definition}[Encoding Tree]
    Given a graph $G = (X, A)$, $X$ is the set of input data points, $A$ is the adjacency matrix, denoting the weights of edges. The encoding tree $\mathcal{T}$ for $G$ is a hierarchical partition of $X$, and satisfies: (1) Each tree node $\alpha \in \mathcal{T}$ corresponds to a subset of data points $T_\alpha \subseteq X$. Specifically, for the root node $\lambda$ of $\mathcal{T}$, we define the points set associated with as $T_\lambda = X$; For the leaf node $\gamma$ of $\mathcal{T}$, $T_\gamma$ contains only one data points (2) For each non-leaf tree node $\alpha$ in $\mathcal{T}$, its parent node is denoted as $\alpha^-$; (3) For each non-leaf tree node $\alpha$ in $\mathcal{T}$, the union of the subsets associated with all its child nodes forms a complete partition of $T_\alpha$. Each tree node describes a partition of the data points $X$, thereby capturing the connectivity patterns of the graph.
\end{definition}

\begin{definition}[Structural Entropy]
    The structural entropy~\cite{li2016structural} is defined under the graph $G$ and the encoding tree $\mathcal{T}$ as follows:
    \begin{align}
        H^{\mathcal{T}}(G) &= \sum_{\alpha\in\mathcal{T},\alpha\neq\lambda} H^{\mathcal{T}}(G, \alpha), \\
        H^{\mathcal{T}}(G, \alpha) &= - \frac{g_\alpha}{\vol(\lambda)}\log_2\frac{\vol(\alpha)}{\vol(\alpha^-)},
    \end{align}
    where the cut $g_\alpha$ is the weight sum of edges that connect points inside $T_\alpha$ and points outside, and the volume $\vol(\alpha)$, $\vol(\alpha^-)$, and $\vol(\lambda)$ denote the degrees sum of all data points within $T_\alpha$, $T_{\alpha^-}$, and $\mathcal{T}$.
\end{definition}

To generalize structural entropy for feature selection, a feature graph $G^f = (\mathbf{D}, \mathbf{A})$ is constructed, where $\mathbf{D}$ is the node set corresponding to all $d$ features, with each feature represented as a node, and $\mathbf{A}$ is the adjacency matrix, where the edge weights reflect the feature correlations. Initially, mutual information between features is calculated and aggregated into a $d \times d$ square matrix $\mathbf{A}$:
\vspace{-0.2em}
\begin{align}
    \mathbf{A}_{i,j} = I(\boldsymbol{x_{:,i}};\boldsymbol{x_{:,j}}).
\label{eq:mutual}
\end{align}
\vspace{-0.2em}
\newtext{It should be noted that the pairwise mutual information in Formula~(\ref{eq:mutual}) serves only as a local observation of feature dependency. In this paper, the term ``high-order correlations'' does not mean that pairwise mutual information directly recovers the complete multivariate dependency. Instead, it refers to the group-wise and hierarchical correlation structures that emerge after structural entropy minimization over the entire feature graph. This interpretation can be further understood through Theorem 25 (Difference Principle of the Positioning Entropy and the Structural Information) from Li and Pan~\cite{li2016structural}. For a partition $\mathcal{P}=\{X_1,\dots,X_L\}$ of the feature graph $G^f$, the difference between the one-dimensional structural entropy and the partition-based structural entropy can be written as
\vspace{-0.2em}
\begin{align}
H^1(G^f)-H^\mathcal{P}(G^f) =& -\sum_{j=1}^L \frac{V_j-g_j}{\mathrm{vol}(G^f)}\log_2\frac{V_j}{\mathrm{vol}(G^f)} \nonumber \\
=& -\sum_{j=1}^L \frac{(1-\phi(X_j))V_j}{\mathrm{vol}(G^f)}\log_2\frac{V_j}{\mathrm{vol}(G^f)},
\label{eq:theorem25}
\end{align}
\vspace{-0.2em}
where $V_j$ denotes the volume of module $X_j$, $g_j$ is the cut weight between $X_j$ and the rest of the graph, and $\phi(X_j)$ is its conductance. This relation shows that the reduction of structural entropy is determined by module-level internal connectivity, external cuts, and module volume, rather than by any isolated pairwise edge alone. Therefore, although the graph is constructed from pairwise mutual information, structural entropy minimization performs a global topological optimization and favors feature groups that are densely connected internally and sparsely connected externally, thereby revealing high-order correlations at the subset and hierarchical levels.}

\newtext{Based on the above interpretation, we next theoretically evaluate the effectiveness of structural entropy in capturing high-order correlations. It is important to note that the true joint entropy and the structural entropy are not quantities of the same nature. For a feature set $S=\{X_1,\dots,X_d\}$, the former is a static statistical quantity directly determined by the joint distribution of the random variables, whereas the latter is the coding cost of the feature graph under a given encoding tree $\mathcal{T}$. Therefore, unlike the true joint entropy, structural entropy is not a fixed quantity and only becomes meaningful after an optimization process over the tree structure. To make the comparison theoretically consistent in the feature selection context, we define
\vspace{-0.2em}
\begin{align}
\Phi(S)=\min_{\mathcal{T}} H^{\mathcal{T}}(G^f(S)),
\label{eq:phi}
\end{align}
\vspace{-0.2em}
where $G^f(S)$ is the mutual-information-induced feature graph of $S$. Under this definition, $\Phi(S)$ represents the optimized structural complexity of the feature system in graph space, while the corresponding statistical complexity in probability space is denoted by $H^*(S)$, and its second-order approximation is denoted by $H^{(2)}(S)$. In this sense, the comparison is not between a raw coding cost and a statistical entropy, but between three complexity measures associated with the same feature system from different perspectives. Moreover, the minimization process defining $\Phi(S)$ is supported by the theoretical properties of structural entropy minimization in Li and Pan~\cite{li2016structural}. In particular, in non-extreme scenarios, Theorem 18 (Locality Theorem) guarantees that structural entropy minimization does not degenerate into a trivial one-cluster solution of the whole graph, while Theorem 19 (Separation Theorem) guarantees that disconnected modules remain strictly separated under minimization. These properties ensure that the optimized quantity $\Phi(S)$ preserves meaningful structural organization in the feature graph and thus serves as a well-defined topological complexity proxy for feature selection. Accordingly, we define the consistency gaps
\vspace{-0.2em}
\begin{align}
\Delta^{\Phi} &= |\Phi(S)-H^*(S)|,\qquad \nonumber \\
\Delta^{(2)} &= |H^{(2)}(S)-H^*(S)|,
\label{eq:delta}
\end{align}
\vspace{-0.2em}
and use them to compare the consistency of the optimized structural complexity and the second-order approximation with the true dependency complexity. The definitions of joint entropy and its second-order approximation are as follows:}
\begin{align}
    H^*(\cdot) &= \sum_{i=1}^nH(X_i)\; - \sum_{1 \leq i < j \leq n} I(X_i;X_j) \nonumber \\
    &\quad + \sum_{1 \leq i < j < k \leq n}I(X_i;X_j;X_k) - \cdots \nonumber \\
    &\quad + (-1)^{n-1}\,H(X_1;\cdots;X_n), \nonumber \\
    H^{(2)}(\cdot) &= \sum_{i=1}^nH(X_i)\; - \sum_{1 \leq i < j \leq n} I(X_i;X_j).
\label{eq:defh}
\end{align}
To rigorously validate the effectiveness of the optimized structural entropy in capturing high-order feature correlations, we examine two limiting cases that represent the extremes of information dependency. According to information decomposition theory~\cite{timme2014synergy}, the dependency structure of any complex system can be fundamentally decomposed into a linear combination of synergistic and redundant components. Therefore, a robust method for high-order correlation learning must demonstrate consistency across the entire spectrum between these two extremes. In this analysis, we define Scenario 1 ($S_1$) to represent the theoretical limit of maximum synergy, and Scenario 2 ($S_2$) to represent the limit of maximum redundancy. These scenarios serve as critical benchmarks to compare $\Phi(S)$ with the conventional second-order approximation.

The performance of a method under these extremes is critical: any robust high-order method must maintain consistency in both regimes. By analyzing these two scenarios, we show that the optimized structural complexity is more consistent with the true dependency complexity than standard low-order approximations. \newtext{Although real-world feature dependencies are usually complex mixtures of synergy and redundancy rather than pure extremes, the locality and separation properties established by Theorem 18 and Theorem 19 imply that structural entropy minimization remains structurally robust under such dynamic mixtures, since it avoids trivial collapse and preserves meaningful module boundaries throughout the optimization process.}

\begin{figure}[t]
    \centering
    \hspace{0.015\textwidth}
    \begin{subfigure}[b]{0.22\textwidth}
        \centering
        \includegraphics[width=\textwidth]{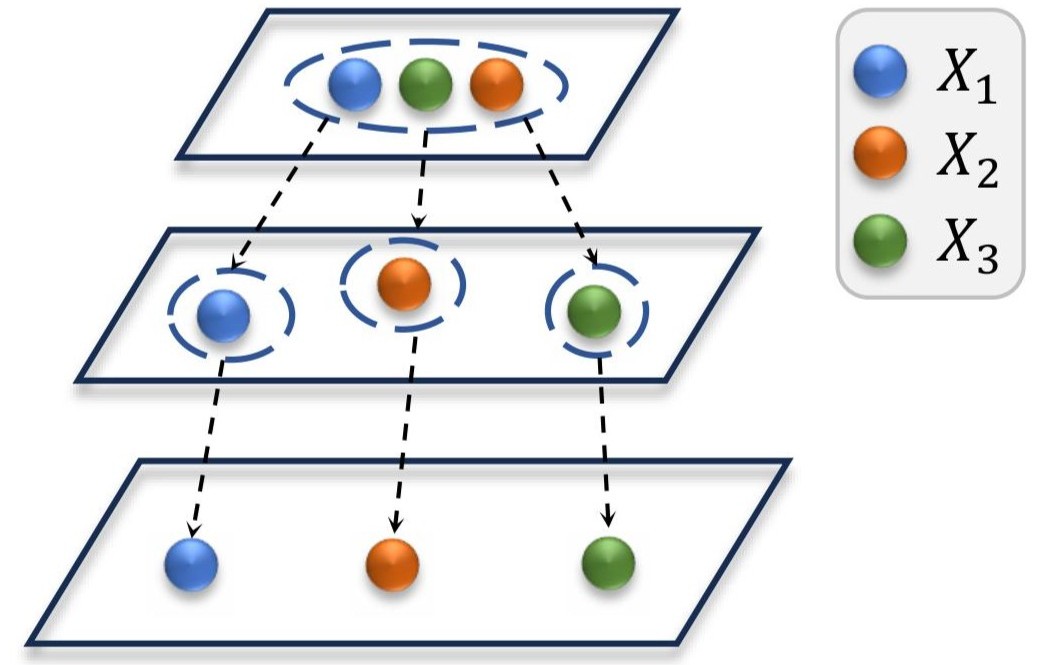}
        \caption{$S_1$}
        \label{fig:s1}
    \end{subfigure}
    \hspace{0.005\textwidth}
    \begin{subfigure}[b]{0.22\textwidth}
        \centering
        \includegraphics[width=\textwidth]{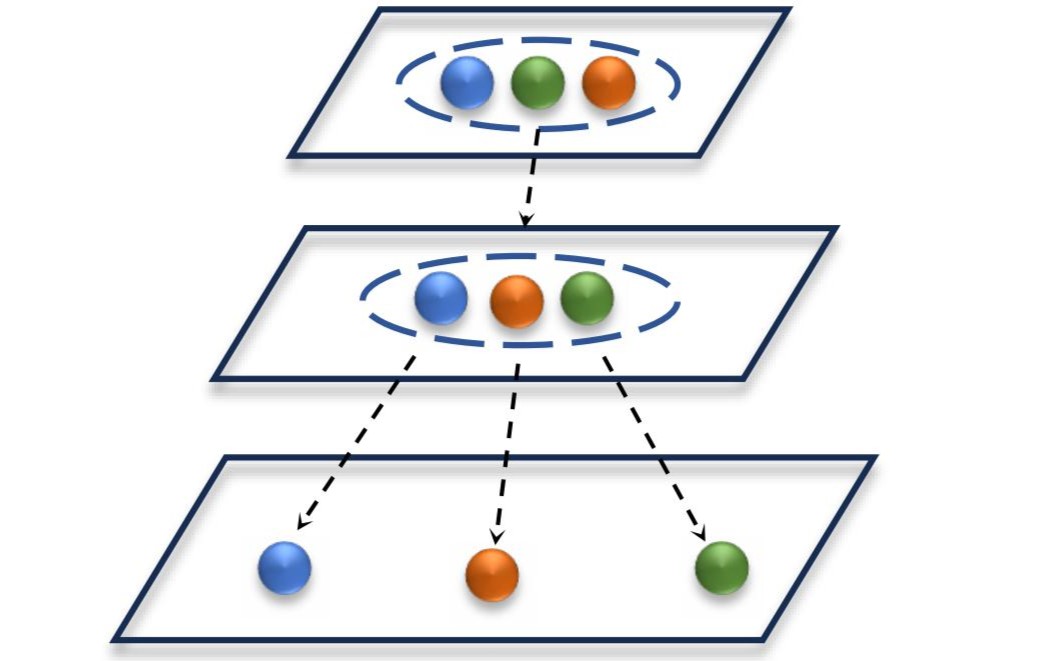}
        \caption{$S_2$}
        \label{fig:s2}
    \end{subfigure}
    \hfill
    
    \caption{Structural-entropy–minimizing encoding tree in Scenario 1 and Scenario 2.}
    \label{fig:encoding}
\end{figure}

\noindent\textbf{Scenario 1: Maximum Synergy (The XOR).}
In this scenario ($S_1$), we consider a system of three features where $X_1$ and $X_2$ are independent uniformly distributed binary variables, and $X_3 = X_1 \oplus X_2$. This configuration represents a state of pure synergy, where information exists solely in the joint distribution of all three variables, with no pairwise correlations.
\begin{itemize}
    \item \textit{Ground-truth Joint Entropy Analysis:}
    Since $X_1$ and $X_2$ are independent and $X_3$ is fully determined by them, any pair of variables is sufficient to determine the state of the entire system. Consequently, the true joint entropy is:
    \begin{equation}
        H^*(S_1) = H(X_1, X_2) = H(X_1) + H(X_2) = 1 + 1 = 2. \nonumber
    \end{equation}
    Despite high-order dependence, the pairwise mutual information vanishes for all pairs: $\forall i \neq j, \; I(X_i; X_j) = 0$.

    \item \textit{Second-order Approximation Solution:}
    Standard low-order methods rely on pairwise statistics. Since all pairwise mutual information terms are zero, the second-order approximation effectively treats the variables as independent:
    \begin{equation}
        H^{(2)}(S_1) = \sum_{i} H(X_i) - \sum_{i<j} I(X_i; X_j) = 3 - 0 = 3. \nonumber
    \end{equation}
    This results in a substantial approximation gap of $\Delta^{(2)} = |H^{(2)}(S_1) - H^*(S_1)| = |3 - 2| = 1$, indicating a complete inability to detect the synergistic structure.

    \item \textit{Structural Entropy Solution:}
    In the feature graph $G^f$, the edge weights are determined by mutual information. Since $I(X_i; X_j) = \varepsilon \rightarrow 0$, the optimal encoding tree $\mathcal{T}_1$ \newtext{which is obtained in the process of structural entropy minimization} partitions the features into three distinct singleton clusters (as illustrated in Figure~\ref{fig:s1}). The structural entropy \newtext{on the optimal encoding tree} is derived as:
    \newtext{
    \begin{align}
        \Phi(S_1) =  H^{\mathcal{T}_1}\big(G^f(S_1)\big) = 4\log 3 - 4 \approx 2.338. \nonumber
    \end{align}
    The resulting metric error is $\Delta^{\Phi} = |2.338 - 2| \approx 0.338 < \Delta^{(2)} = 1$.} This demonstrates that structural entropy yields a smaller consistency gap than the second-order approximation, even when explicit pairwise signals are absent.
\end{itemize}
\noindent\textbf{Scenario 2: Maximum Redundancy (The Equality Case).}
In this scenario ($S_2$), we define a system with extreme redundancy involving three identical features: $X_1 = X_2 = X_3$. In this regime, knowledge of any single feature is sufficient to resolve the uncertainty of the entire system.
\begin{itemize}
    \item \textit{Ground-truth Joint Entropy Analysis:}
    Since the variables are identical, the joint entropy collapses to the entropy of a single variable. With each feature having unit entropy ($H(X_i)=1$), the true joint entropy is:
    \begin{equation}
        H^*(S_2) = H(X_1) = 1. \nonumber
    \end{equation}
    Furthermore, the pairwise mutual information is maximized: $\forall i, j, \; I(X_i; X_j) = H(X_i) = 1$.

    \item \textit{Second-order Approximation Solution:}
    The second-order approximation attempts to correct for independence by subtracting pairwise overlaps. However, in cases of high redundancy, it over-corrects:
    \begin{equation}
        H^{(2)}(S_2) = \sum_{i} H(X_i) - \sum_{i<j} I(X_i; X_j) = 3 - 3 = 0. \nonumber
    \end{equation}
    This leads to a physically impossible entropy estimate of zero (implying determinism without observation) and a significant error of $\Delta^{(2)} = |0 - 1| = 1$.

    \item \textit{Structural Entropy Solution:}
    Given that all edge weights in $G^f$ equal $1$, strong connectivity causes the optimal encoding tree $\mathcal{T}_2$ \newtext{derived during structural entropy minimization} to aggregate all three features into a single cluster (as shown in Figure~\ref{fig:s2}). The structural entropy \newtext{on the optimal encoding tree} is derived as:
    \newtext{
    \begin{align}
        \Phi(S_2) = H^{\mathcal{T}_2}\big(G^f(S_2)\big) = 2\log 3 - 2 \approx 1.169. \nonumber
    \end{align}
    The corresponding error is $\Delta^{\Phi} = |1.169 - 1| \approx 0.169 < \Delta^{(2)} = 1$.} This result validates that structural entropy effectively compresses redundant information, maintaining high accuracy where low-order expansions collapse.
\end{itemize}

Based on the feature graph $G^f$, a three-layer encoding tree $\mathcal{T}$ is designed to describe correlation patterns among features. The root layer consists of a single root node containing all features; the leaf layer contains $d$ leaf nodes, each representing an individual feature; \newtext{and the intermediate layer contains $q$ semantic nodes aligned with the $q$-dimensional label space. Here, the choice of $q$ should not be interpreted as assuming that the true number of higher-order redundant feature groups is exactly equal to the number of labels. Instead, the intermediate layer is introduced as a task-guided semantic bottleneck, so that the structural organization learned from the feature graph can be directly aligned with the downstream supervision space without introducing additional mapping parameters.} \newtext{An assignment matrix $\mathbf{W}$ is introduced to connect the intermediate layer and the leaf layer, where $\mathbf{W}_{i,j}$ measures the association strength between the $i$-th feature and the $j$-th label-related semantic node. In this sense, $\mathbf{W}$ serves as a feature selection matrix. Since $\mathbf{W}$ is learned together with the classifier (e.g., $\mathbf{XW}=\mathbf{Y}$), the intermediate semantic nodes are encouraged to align with the label space while still being determined by the structural organization of the feature graph. This design reduces optimization difficulty and provides an inductive bias toward feature combinations that are most relevant to the prediction task.} To enhance the representational ability of structural entropy, the row sums of the feature selection matrix $\mathbf{W}$ are constrained to one, thereby ensuring both selection effectiveness and interpretability. The structural entropy of the intermediate layer nodes for the three-tier encoding tree $\mathcal{T}$ is as follows:
\begin{align}
    \mathcal{L}_{\mathrm{SE}}^\mathcal{T}(\mathbf{W}) &= -\sum_{j=1}^{q} \frac{((\mathbf{1} - \mathbf{W})^\top \mathbf{A W})_{jj}}{\Sum(\mathbf{A})} \log_2 \frac{(\mathbf{1^\top A W})_{jj}}{\Sum(\mathbf{A})}, \nonumber \\
    & s.t. \quad \mathbf{W}\boldsymbol{1}_q = \boldsymbol{1}_d,
\label{eq:lsew}
\end{align}
where $\boldsymbol{1}_q$ and $\boldsymbol{1}_d$ are the $q$-dimensional and $d$-dimensional all-ones column vectors, respectively, $\Sum(\cdot)$ denotes the element-wise summation of a matrix.

\newtext{Under the constraint $\mathbf{W}\boldsymbol{1}_q = \boldsymbol{1}_d$, each row of $\mathbf{W}$ lies on the probability simplex. As a result, the learned matrix represents a continuous soft assignment of features rather than a hard binary partition. This flexibility allows multiple topologically distinct feature groups to contribute to the same semantic node when they are related to the same label, while a single feature group may also distribute its weights across multiple semantic nodes. Therefore, setting the intermediate dimension to $q$ does not force a one-to-one correspondence between feature groups and labels, nor does it replace the intrinsic topology of $G^f$; rather, it controls the task-relevant compression granularity under label supervision.}

\begin{figure}[t]
    \centering
    \includegraphics[width=0.45\textwidth]{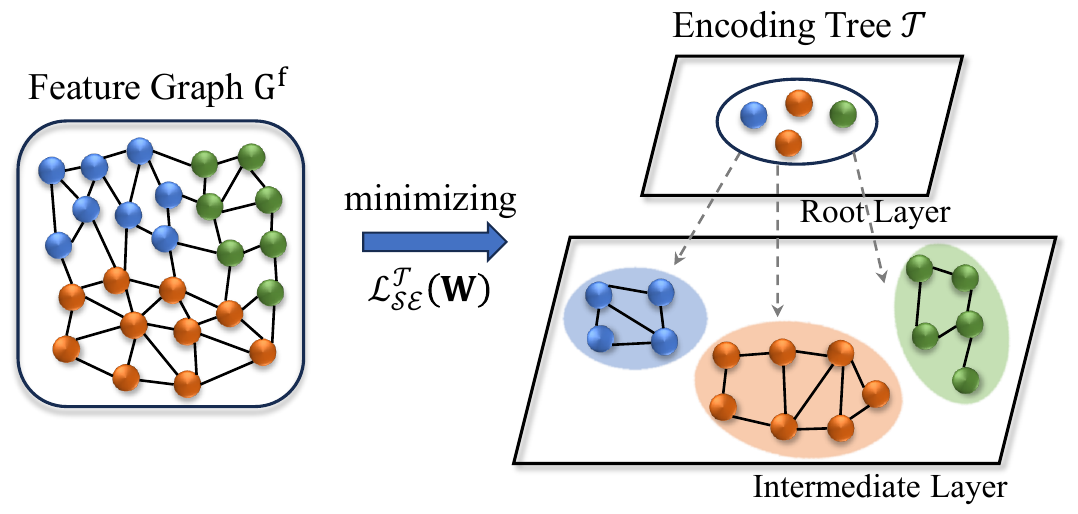}
    \caption{Subgraphs formed by nodes of the same color are highly densely connected, which is defined as ``redundancy''. By minimizing structural entropy, the feature graph $G^f$ is converted into an optimal encoding tree $\mathcal{T}$, which increases intra-layer edge weights while decreasing the weights from nodes to the root of their parent layer.}
    \label{fig:redu}
\end{figure}

``Redundancy'' is characterized as high-density connected subgraphs based on the definition of the feature graph $G^f$. By minimizing structural entropy, the feature graph $G^f$ is converted into an optimal encoding tree $\mathcal{T}$, \newtext{which increases intra-group connectivity while decreasing the connections from local groups to higher-level parent nodes.} Unlike mutual information-based methods, SEHFS explicitly models intra-layer and node–root correlations, achieving higher-order correlation learning. \newtext{In this way, structurally related features are organized into low-redundancy groups.} The principle of redundancy elimination in SEHFS is shown in Figure~\ref{fig:redu}. \newtext{Through the continuous feature selection matrix $\mathbf{W}$, these groups are softly projected onto the $q$-dimensional label semantic space, resulting in task-relevant and low-redundancy feature selection.}
\subsection{Global View Matrix Reconstruction}
In multi-view multi-label learning, information-theoretic methods are challenged in optimizing data from different views. To address this, global view matrix reconstruction is proposed to obtain a unified feature matrix with cross-view consistency, facilitating global feature selection. The reconstruction of the global view matrix mainly consists of two components: (1) learning a shared semantic matrix to capture consistency; (2) learning view-specific contribution matrices to capture complementarity. By integrating these two components, the global view matrix achieves a balance between shared consistency and unique complementarity across views.

\subsubsection{Learning Shared Semantic Matrix}
The semantic matrix of $v$-th view $\mathbf{S}_v \in \mathbb{R}^{n\times n}$ represents the semantics of $v$-th view by calculating the relationships among samples. $\mathbf{S}_v$ is calculated as follows:
\begin{align}
    s_{ij}^{(v)} = 
    \begin{cases} 
    \exp\left(-\dfrac{\|x_i^{(v)} - x_j^{(v)}\|_2^2}{\sigma^2}\right) & 
    \begin{aligned}
    &\text{if } x_i^{(v)} \in \mathcal{N}_k(x_j^{(v)}) \text{ or} \\
    &\quad x_j^{(v)} \in \mathcal{N}_k(x_i^{(v)})
    \end{aligned}, \\
    0 & \text{otherwise},
    \end{cases}
\label{eq:sv}
\end{align}
where $\sigma$ represents a predefined parameter, and $\mathcal{N}_k(x_j^{(v)})$ represents the set of $k$ nearest neighbors of $x_j^{(v)}$. \newtext{In all experiments, the number of nearest neighbors is fixed to $k=50$. For each view, the Gaussian kernel parameter $\sigma$ is set as the mean of all non-zero pairwise distances in that view, so that the scale of the local similarity graph is adaptively determined by the data distribution.}

The shared semantic matrix $\mathbf{S} \in \mathbb{R}^{n\times n}$ is obtained by integrating the semantic matrix $\mathbf{S}_v$ of each view, and the similarity between $\mathbf{S}_v$ and $\mathbf{S}$ guides the learning of the view weight $\boldsymbol{\alpha}_v\in \mathbb{R}^{+}$. Serving as the core component of the global view matrix $\mathbf{X^f}\in \mathbb{R}^{n\times d}$, the shared semantic matrix $\mathbf{S}$, together with view weight vector $\boldsymbol{\alpha}\in \mathbb{R}^{V\times 1}$, directly guides the integration of each view matrix $\mathbf{X}_v\in \mathbb{R}^{n\times d(v)}$. To learn each view matrix under the global view dimensionality, a transformation matrix $\mathbf{P}_v\in \mathbb{R}^{d(v)\times d}$ is defined. When concatenating the views in sequence, $\mathbf{P}_v$ is constructed as a diagonal matrix with all ones and a specific offset determined by the number of features $d(v)$ in each view. In this way, $\mathbf{X}_v\mathbf{P}_v \in \mathbb{R}^{n\times d}$ enables the unified integration of all views. The following objective function is formulated as the main component for global view matrix reconstruction:
\begin{align}
    \min_{\mathbf{X^f},\mathbf{S},\boldsymbol{\alpha}_v}  &\; \beta\,\{\|\mathbf{X^f} - \mathbf{S} \sum_{v=1}^V\boldsymbol{\alpha}_v\mathbf{X}_v \mathbf{P}_v\|_F^2 + \|\mathbf{S} - \sum_{v=1}^V\boldsymbol{\alpha}_v\mathbf{S}_v\|_F^2 \} \nonumber \\
    & + \lambda\,\Tr(\mathbf{L_Y}^\top \mathbf{S}\,\mathbf{L_Y}), \qquad s.t. \;\; \mathbf{X^f},\mathbf{S},\boldsymbol{\alpha}_v\geq 0,
\label{eq:tr}
\end{align}
where a graph Laplacian regularization term $\Tr(\mathbf{L_Y}^\top\mathbf{S}\,\mathbf{L_Y})$ is introduced to ensure consistency between the shared semantic matrix and the label matrix. Here, $\mathbf{L_Y} = \mathbf{D_Y} - \mathbf{S_Y} \in \mathbb{R}^{n \times n}$ denotes the graph Laplacian matrix of the label matrix, $\mathbf{S_Y}$ is a sample-wise cosine similarity matrix computed from the label matrix $\mathbf{Y}$, and $\mathbf{D_Y}$ is the degree matrix with diagonal elements equal to the row sums of $\mathbf{S_Y}$. Minimizing this regularization term effectively enhances the intrinsic correlation between features and labels, thereby improving the discriminability of feature representations.

\subsubsection{Supplementing View-specific Contribution Matrix}
To compensate for the missing view-specific information in $\mathbf{S}$, the view-specific contribution matrix of each view needs to be supplemented to $\mathbf{X^f}$. By utilizing the transpose of the transformation matrix $\mathbf{P}_v^\top$, $\mathbf{X^f}$ under the dimensionality $d(v)$ of $v$-th view can be calculated. Accordingly, for $v$-th view, a unique matrix $\mathbf{H}_v$ is learned. $\mathbf{X^f}$ under the corresponding view dimension is encouraged to be similar to the view-specific contribution matrix $\mathbf{X}_v\mathbf{H}_v$. Together with Formula~(\ref{eq:tr}), the following objective function is formulated as follows:
\begin{align}
\hspace{-0.5em}
    \min_{\mathbf{X^f},\mathbf{S},\mathbf{H}_v,\boldsymbol{\alpha}_v}  & \beta\{\|\mathbf{X^f} - \mathbf{S} \sum_{v=1}^V\boldsymbol{\alpha}_v\mathbf{X}_v \mathbf{P}_v\|_F^2 + \|\mathbf{S} - \sum_{v=1}^V\boldsymbol{\alpha}_v\mathbf{S}_v\|_F^2\} \nonumber \\
    & + \lambda\Tr(\mathbf{L_Y}^\top \mathbf{S}\,\mathbf{L_Y}) + \gamma\sum_{v=1}^V\|\mathbf{X^f P}_v^\top - \mathbf{X}_v \mathbf{H}_v\|_F^2, \nonumber\\
    &\; s.t. \quad \mathbf{X^f},\mathbf{S},\mathbf{H}_v,\boldsymbol{\alpha}_v\geq 0.
\label{eq:h}
\end{align}

\newtext{The above formulation shows that $\mathbf{S}$ and $\mathbf{H}^v$ play complementary rather than redundant roles in global view reconstruction. Specifically, $\mathbf{S}$ captures the cross-view consensus structure that is consistently supported by multiple views, while $\mathbf{H}^v$ compensates for the view-specific information that cannot be fully preserved in the shared semantic space. Meanwhile, the graph Laplacian regularization term $\mathrm{Tr}(\mathbf{L}_Y^\top \mathbf{S}\mathbf{L}_Y)$ further aligns the shared semantic structure with the label manifold. Therefore, the reconstructed global view is obtained through the joint action of shared semantic modeling, view-specific compensation, and label-aware regularization, which together balance consistency, complementarity, and discriminability. This collaborative design underlies the reconstruction of the global view matrix.}

\subsection{Information-Matrix Fusion Framework}
To reconcile global and local objectives, we propose a novel multi-view, multi-label framework. It reconstructs a global view matrix to optimize the overall structure and, in conjunction with a structural entropy–guided feature selection module, performs sample-level optimization by selecting features from the reconstructed global matrix. The final objective function is as follows:
\begin{align}
    &\min_{\mathbf{X^f},\mathbf{W},\mathbf{S},\mathbf{H}_v,\boldsymbol{\alpha}_v}  \; \|\mathbf{X^f W} - \mathbf{Y}\|_F^2 + \alpha\, \mathcal{L}_{\mathrm{SE}}^\mathcal{T}(\mathbf{W}) \nonumber \\
    & + \beta\, \{\|\mathbf{X^f} - \mathbf{S} \sum_{v=1}^V\boldsymbol{\alpha}_v\mathbf{X}_v \mathbf{P}_v\|_F^2 + \|\mathbf{S} - \sum_{v=1}^V\boldsymbol{\alpha}_v\mathbf{S}_v\|_F^2\} \nonumber \\
    & + \lambda \Tr(\mathbf{L_Y}^\top \mathbf{S}\,\mathbf{L_Y}) + \gamma \sum_{v=1}^V\|\mathbf{X^f P}_v^\top - \mathbf{X}_v \mathbf{H}_v\|_F^2, \nonumber \\
    & \quad\quad\quad s.t. \quad \mathbf{X^f},\mathbf{W},\mathbf{S},{H}_v,\boldsymbol{\alpha}_v\geq 0,\; \mathbf{W}\boldsymbol{1}_q = \boldsymbol{1}_d.
\label{eq:overall}
\end{align}

\subsection{Optimization}
\newtext{Since the overall objective in Formula~(\ref{eq:overall}) is non-convex with respect to all variables jointly, we adopt an alternating optimization strategy that updates one variable block at a time while fixing the others. The overall objective formula has a total of five variables $\mathbf{X^f,W,S,H}_v,\boldsymbol{\alpha}_v$ to be optimized.} The optimization details of the five variables are presented as follows:

\subsubsection{Update $\mathbf{X^f},\mathbf{S}$ and $\mathbf{H}_v$}
The equivalence formula for the Frobenius norm is defined as $\|\mathbf{X}\|_F^2 = \Tr(\mathbf{X^\top X})$. So the objective function can be rewritten as:
\begin{align}
    & \Theta(\mathbf{X^f,S,H}_v) = \Tr((\mathbf{X^f W} - \mathbf{Y})^\top (\mathbf{X^f W} - \mathbf{Y})) \nonumber\\
    &+ \beta\, \Tr((\mathbf{X^f} - \mathbf{S} \sum_{v=1}^V\boldsymbol{\alpha}_v\mathbf{X}_v \mathbf{P}_v)^\top(\mathbf{X^f} - \mathbf{S} \sum_{v=1}^V\boldsymbol{\alpha}_v\mathbf{X}_v \mathbf{P}_v)) \nonumber\\
    &+ \beta\, \Tr((\mathbf{S} - \sum_{v=1}^V\boldsymbol{\alpha}_v\mathbf{S}_v)^\top(\mathbf{S} - \sum_{v=1}^V\boldsymbol{\alpha}_v\mathbf{S}_v)) + \lambda \Tr(\mathbf{L_Y^\top S L_Y}) \nonumber\\
    &+ \gamma \sum_{v=1}^V \Tr((\mathbf{X^f P}_v^\top - \mathbf{X}_v \mathbf{H}_v)^\top(\mathbf{X^f P}_v^\top - \mathbf{X}_v \mathbf{H}_v)) .
\label{eq:update}
\end{align}

Multiplicative gradient descent strategy~\cite{beck2009fast} is adopted to solve Formula~(\ref{eq:update}). Meanwhile, to maintain the validity of the constraints, the following update rules are derived according to the KKT conditions:
\begin{align}
    & \mathbf{X^f} \xleftarrow{} \mathbf{X^f}\cdot \dfrac{\mathbf{Y W^\top}+ \sum_{v=1}^V\beta\,\boldsymbol{\alpha}_v\mathbf{S}\mathbf{X}_v\mathbf{P}_v+\gamma \mathbf{X}_v \mathbf{H}_v \mathbf{P}_v}{\mathbf{X^f W W^\top} + \beta\, \mathbf{X^f} + \gamma \sum_{v=1}^V\mathbf{X^f P}_v^\top \mathbf{P}_v }
\label{eq:updateX}, \\
    & \mathbf{S} \xleftarrow{} \mathbf{S}\cdot \dfrac{2\beta\, \mathbf{X^f}\sum_{v=1}^V\boldsymbol{\alpha}_v\mathbf{P}_v^\top\mathbf{X}_v^\top + 2\beta\, \sum_{v=1}^V\boldsymbol{\alpha}_v\mathbf{S}_v}{2\beta\, \mathbf{S} (\mathbf{I} + \sum_{v=1}^V\boldsymbol{\alpha}_v^2\mathbf{X}_v\mathbf{P}_v\mathbf{P}_v^\top\mathbf{X}_v^\top) + \lambda\mathbf{L_Y} \mathbf{L_Y^\top}}
\label{eq:updateS}, \\
    & \mathbf{H}_v \xleftarrow{} \mathbf{H}_v \cdot \dfrac{\mathbf{X}_v^\top\mathbf{X^f}\mathbf{P}_v^\top}{\mathbf{X}_v^\top\mathbf{X}_v\mathbf{H}_v}.
\label{eq:updateH}
\end{align}

When implementing the optimization, a small positive value $\varepsilon \xrightarrow{} 0^{+}$ is added to the denominator to prevent division-by-zero errors.

\subsubsection{Update $\mathbf{W}$}
Given $\mathbf{X^f}\in\mathbb{R}^{n\times d}$, $\mathbf{Y}\in\mathbb{R}^{n\times r}$, and a fixed matrix $\mathbf{A}\in\mathbb{R}^{d\times d}$, we estimate $\mathbf{W}\in\mathbb{R}^{d\times r}$ by
\begin{align}
\min_{\mathbf{W}} \;&  J(W) \;=\; \| \mathbf{X^f W} - \mathbf{Y}\|_F^2 + \alpha\,\mathcal{L}_{\mathrm{SE}}^{\mathcal{T}}(W), \nonumber \\
& s.t. \quad  \mathbf{W} \geq \mathbf{0},\; \mathbf{W}\boldsymbol{1}_q = \boldsymbol{1}_d.
\label{eq:objectiveW}
\end{align}

Setting $S = \Sum(\mathbf{A}) = \sum_{i=1}^d\sum_{k=1}^d \mathbf{A}_{ik} = \mathbf{1}^\top \mathbf{A}\mathbf{1}, \boldsymbol{q_j} = \mathbf{1}^\top A w_j, \boldsymbol{g_j} = ( \mathbf{1 - w_j})^\top A \boldsymbol{w_j}$, where $\mathbf{1}\in \mathbb{R}^{d}$ denote the all-ones vector, $\boldsymbol{w_j}$ represents the $j$-th column of $\mathbf{W}$, thus the optimization subproblem Formula~(\ref{eq:objectiveW}) is transformed into:
\begin{algorithm}[t]
    \caption{Pseudo code of SEHFS}
    \label{alg:algorithm}
     \textbf{Input}: Data matrices $\{\mathbf{X}_v\}_{v=1}^V$; Label matrix $\mathbf{Y}$. \\
     \textbf{Parameter}: Hyperparameters $\alpha,\,\beta,\,\gamma,\,\lambda$. \\
     \textbf{Output}: Selected features $fea\_idx$. \\ \vspace{-1.2em}
    \begin{algorithmic}[1] 
        \STATE Initialize $\mathbf{X^f}, \mathbf{W}, \mathbf{H}_v, \boldsymbol{\alpha}_v$;
        \FOR{$v = 1$ to $V$}
            \STATE Calculate $\mathbf{S}_v$ by Formula~(\ref{eq:sv});
        \ENDFOR
        \STATE Calculate $\mathbf{L_Y} = \mathbf{D_Y} - \mathbf{S_Y}$;
        \STATE Calculate the mutual information matrix $\mathbf{A}$ from input;
        \WHILE{not converged}
            \STATE Update $\mathbf{X^f}$ by Formula~(\ref{eq:updateX});
            \STATE Update $\mathbf{W}$ by Formula~(\ref{eq:updateW});
            \STATE Update $\mathbf{S}$ by Formula~(\ref{eq:updateS});
            \STATE Update $\mathbf{H}_v$ by Formula~(\ref{eq:updateH});
            \STATE Update $\boldsymbol{\alpha}_v$ by Formula~(\ref{eq:updateA});
        \ENDWHILE
        \STATE 
        \STATE Compute $s_i=\|\mathbf{W}_{i,\cdot}\|_2$ for all features;
        \STATE Set $fea\_idx$ as the ranked feature index in descending order according to $s_i$;
        \STATE \textbf{return} list of selected features $fea\_idx$.
    \end{algorithmic}
\end{algorithm}
\begin{align}
\mathcal{L}_{\mathrm{SE}}^{\mathcal{T}}(W) 
\;=\; -\frac{1}{S}\sum_{j=1}^q g_j \;\log_2\!\left(\frac{q_j}{S}\right).
\label{eq:lse}
\end{align}
In implementation, a small positive value $\varepsilon \xrightarrow{} 0^{+}$ is used to guard the logarithm, i.e., $q_j \leftarrow \max(q_j,\varepsilon)$. The gradients can be expressed as follows:
\begin{align}
& \nabla_{\mathbf{W}} \| \mathbf{X^f} \mathbf{W} - \mathbf{Y}\|_F^2 
= 2\,\mathbf{X^f}^\top(\mathbf{X^f} \mathbf{W} - \mathbf{Y}), \nonumber\\
& \nabla_{\boldsymbol{w}_j}\mathcal{L}_{\mathrm{SE}}^{\mathcal{T}}
= -\frac{1}{S}\Big[h_j\big(\boldsymbol{a}-\mathbf{A}_{\text{sym}}\boldsymbol{w}_j\big)
+ \frac{g_j}{\ln 2\, q_j}\,\boldsymbol{a}\Big],
\label{eq:grad}
\end{align}
where $h_j$ denotes $\log_2\!\Big(\frac{q_j}{S}\Big)$, $\boldsymbol{a}$ denotes $\mathbf{A}^\top\boldsymbol{1}$
, and $\mathbf{A}_{\text{sym}}$ denotes $\mathbf{A}+\mathbf{A}^\top
$. Thus, the optimization subproblem is then solved using the projected gradient descent method:
\begin{align}
& \mathbf{G_W}^{(t)}=\nabla_{\mathbf{W}} J(\mathbf{W}^{(t)}), \nonumber \\
& \widetilde{\mathbf{W}}=\mathbf{W}^{(t)}-\eta_t \mathbf{G_W}^{(t)},\nonumber\\
& \mathbf{W}^{(t+1)}=\Pi_{\Delta}\!\big(\widetilde{\mathbf{W}}\big),
\label{eq:updateW}
\end{align}
where $\Pi_{\Delta}$ projects each row onto the probability simplex $\Delta=\{\boldsymbol{w}\in\mathbb{R}^r:\boldsymbol{w}\ge \boldsymbol{0},\ \boldsymbol{1}^\top\boldsymbol{w}=1\}$. The row-wise projection~\cite{duchi2008efficient} is given by
\vspace{-9pt}
\begin{align}
& \boldsymbol{u}=\operatorname{sort}(\boldsymbol{v})\ \text{(desc.)} ,\quad c_k=\sum_{i=1}^k u_i, \nonumber\\
& \rho=\max\{k:\ u_k-\tfrac{c_k-1}{k}>0\}, \nonumber\\
& \theta=\dfrac{c_\rho-1}{\rho}, \nonumber\\
& [\Pi_{\Delta}(\boldsymbol{v})]_i=\max\{v_i-\theta,0\}.
\end{align}
The step size $\eta_t$ is chosen by Armijo backtracking until
\begin{align}
J\!\big(\Pi_{\Delta}(\mathbf{W}^{(t)}-\eta_t \mathbf{G_W}^{(t)})\big)
\le J(\mathbf{W}^{(t)})-c\,\eta_t \lVert \mathbf{G_W}^{(t)}\rVert_F^2,
\end{align}
and the iteration stops when the relative decrease satisfies
\begin{align}
\frac{\big|J(\mathbf{W}^{(t+1)})-J(\mathbf{W}^{(t)})\big|}{J(\mathbf{W}^{(t)})+\varepsilon}<\tau.
\end{align}

\subsubsection{Update $\boldsymbol{\alpha}_v$}
By fixing other variables, the optimization subproblem for $\boldsymbol{\alpha}_v$ can be written as follows:
\begin{align}
\label{eq:problema}
    \min_{\boldsymbol{\alpha}_v} \; \|\mathbf{X^f} - \mathbf{S}\sum_{v=1}^V\boldsymbol{\alpha}_v\mathbf{X}_v\mathbf{P}_v\|_F^2 + \|\mathbf{S} - \sum_{v=1}^V\boldsymbol{\alpha}_v\mathbf{S}_v\|_F^2 .
\end{align}
Setting $\mathbf{\psi} = \ve(\mathbf{X^f}) \in \mathbb{R}^{nd\times 1}, \mathbf{\Psi} = \ve([\mathbf{S}\mathbf{X}_1\mathbf{P}_1,\dots,\\\mathbf{S}\mathbf{X}_V\mathbf{P}_V]]) \in \mathbb{R}^{nd\times V}$, and likewise, $\mathbf{\omega} = \ve(\mathbf{S}) \in \mathbb{R}^{nn\times 1}, \\\mathbf{\Omega} = \ve([\mathbf{S}_1,\dots,\mathbf{S}_V]) \in \mathbb{R}^{nn\times V}$, thus the optimization Subproblem~\ref{eq:problema} is transformed into:
\begin{align}
    \min_{\boldsymbol{\alpha}_v} \; \boldsymbol{\alpha}^\top \mathbf{M}\boldsymbol{\alpha} - \boldsymbol{\alpha}^\top \mathbf{h},
\label{eq:updateA}
\end{align}         
where $\mathbf{M} = \mathbf{\Psi^\top\Psi} + \mathbf{\Omega^\top\Omega}, \mathbf{h} = 2(\mathbf{\Psi^\top\psi} + \mathbf{\Omega^\top\omega})$. Due to the semidefinite $\mathbf{M}$, Formula~(\ref{eq:updateA}) is a quadratic convex programming problem and can be solved efficiently~\cite{jiang2023adaptive}.

\subsection{Convergence Property of Optimization}
\label{subsec:convergence}
Since the overall objective in Formula~(\ref{eq:overall}) is non-convex with respect to all variables jointly, we adopt an alternating optimization strategy that updates one variable block at a time while fixing the others.

In each iteration, the updates of $\mathbf{X^f}$, $\mathbf{S}$, and $\mathbf{H}_v$ are carried out by multiplicative rules derived under the KKT conditions~\cite{bertsekas1997nonlinear}, which preserve the non-negativity constraints and do not increase the corresponding subproblem objective. The update of $\mathbf{W}$ is solved by projected gradient descent with Armijo backtracking, which guarantees sufficient descent of the subproblem objective. The update of $\boldsymbol{\alpha}_v$ corresponds to a convex quadratic programming subproblem and can therefore be effectively solved. As a result, each block update makes the overall objective value non-increasing, and the objective sequence generated by Algorithm~\ref{alg:algorithm} is monotonically non-increasing.

\newtext{Moreover, all terms in Formula~(\ref{eq:overall}) are non-negative, consisting of the reconstruction losses in Formula~(\ref{eq:h}), the structural entropy regularization in Formula~(\ref{eq:lsew}), and the graph Laplacian regularization term $\Tr(\mathbf{L_Y}^\top \mathbf{SL_Y})$. Therefore, the overall objective is lower bounded by 0. Since the objective value sequence is monotonically non-increasing and lower bounded, it is guaranteed to converge.}

\newtext{Nevertheless, due to the non-convexity caused by the strong coupling among multiple variables, this analysis only guarantees convergence to a stable point, rather than the global optimum. In practice, the risk of poor local optima can be alleviated by reasonable initialization and multiple random restarts.}

\begin{table*}[hb]
\footnotesize
\centering
\caption{The detailed information for the datasets in our experiment}
\label{tab:datasets}
\begin{tabularx}{\textwidth}{l@{\hspace{5pt}}*8{X}}
\toprule
Datasets      & EMOTIONS    & YEAST  & VOC07 & MIRFlickr & SCENE   & OBJECT  & Corel5K     & IAPRTC12    \\ 
\midrule
View-1($d_1$) & Rhythmic(8) & GE(79)   & DH(100)   & DH(100)   & CH(64)  & CH(64)  & DH(100)     & DH(100)     \\
View-2($d_2$) & Timbre(64)  & PP(24)   & GIST(512) & GIST(512) & CM(225) & CM(225) & DHV3H1(300) & DHV3H1(300) \\
View-3($d_3$)  & -   & -    & HH(100)  & HH(100)   & CORR(144) & CORR(144) & GIST(512)   & GIST(512)   \\
View-4($d_4$)  & -   & -    & -        & -         & EDH(73)   & EDH(73)   & HHV3H1(300) & HHV3H1(300) \\
View-5($d_5$)  & -   & -    & -        & -         & WT(128)   & WT(128)   & HH(100)     & HH(100)     \\
\midrule
Instances($n$) & 593 & 2417 & 3817     & 4053    & 4400      & 6047      & 4999        & 4999        \\
Features($d$)  & 72  & 103  & 712      & 712     & 634       & 634       & 1312        & 1312        \\
Labels($q$)    & 6   & 12   & 20       & 38      & 33        & 31        & 260         & 260         \\
\bottomrule
\end{tabularx}
\end{table*}
\begin{table*}[ht]
  \footnotesize
  \centering
  \caption{Experimental results $\uparrow$ (mean$\pm$std) of each method on AP; the best result is shown in bold and the second-best is underlined.}
  \label{tab:AP}
  \begin{tabularx}{\textwidth}{l@{\hspace{5pt}}*8{>{\centering\arraybackslash}X}}
    \toprule
    Datasets & SEHFS & DHLI & GRAFS & MSFS & MIFS & SPLDG & SRFS & MSSL \\
    \midrule
    EMOTIONS & \textbf{.686$\pm$.052} & .591$\pm$.045 & .605$\pm$.036 & .591$\pm$.050 & .637$\pm$.049 & .654$\pm$.049 & \underline{.666$\pm$.052} & .631$\pm$.048 \\
    YEAST & \underline{.670$\pm$.019} & .663$\pm$.022 & .665$\pm$.020 & .651$\pm$.039 & .663$\pm$.024 & .663$\pm$.019 & \textbf{.672$\pm$.021} & .660$\pm$.020 \\
    VOC07 & .592$\pm$.019 & .585$\pm$.017 & .586$\pm$.020 & \underline{.593$\pm$.016} & .576$\pm$.021 & .583$\pm$.015 & \textbf{.598$\pm$.016} & .577$\pm$.016 \\
    MIRFlickr & \textbf{.692$\pm$.022} & .655$\pm$.016 & .659$\pm$.025 & .682$\pm$.019 & .646$\pm$.019 & .682$\pm$.015 & \underline{.687$\pm$.019} & .678$\pm$.020 \\
    SCENE & \textbf{.804$\pm$.019} & .786$\pm$.018 & .801$\pm$.019 & .718$\pm$.046 & .792$\pm$.017 & .788$\pm$.017 & \underline{.801$\pm$.016} & .797$\pm$.018 \\
    OBJECT & \textbf{.454$\pm$.052} & .408$\pm$.051 & \underline{.437$\pm$.056} & .357$\pm$.049 & .436$\pm$.044 & .423$\pm$.050 & .435$\pm$.045 & .433$\pm$.054 \\
    Corel5K & \textbf{.210$\pm$.039} & .191$\pm$.040 & \underline{.196$\pm$.039} & .184$\pm$.035 & .157$\pm$.032 & .173$\pm$.040 & .177$\pm$.039 & .190$\pm$.036 \\
    IAPRTC12 & \textbf{.215$\pm$.015} & .192$\pm$.015 & \underline{.193$\pm$.017} & .184$\pm$.017 & .175$\pm$.018 & .183$\pm$.018 & .182$\pm$.017 & .190$\pm$.017 \\
    \bottomrule
  \end{tabularx}
\end{table*}
\begin{table*}[ht]
  \footnotesize
  \centering
  \caption{Experimental results $\downarrow$ (mean$\pm$std) of each method on Cov; the best result is shown in bold and the second-best is underlined.}
  \label{tab:Cov}
  \begin{tabularx}{\textwidth}{l@{\hspace{5pt}}*8{>{\centering\arraybackslash}X}}
    \toprule
    Datasets & SEHFS & DHLI & GRAFS & MSFS & MIFS & SPLDG & SRFS & MSSL \\
    \midrule
    EMOTIONS & \textbf{.663$\pm$.029} & .775$\pm$.035 & .749$\pm$.029 & .769$\pm$.033 & .722$\pm$.031 & .706$\pm$.033 & \underline{.681$\pm$.026} & .721$\pm$.033 \\
    YEAST & \underline{.679$\pm$.034} & .690$\pm$.037 & .682$\pm$.034 & .729$\pm$.062 & \underline{.679$\pm$.042} & .688$\pm$.039 & \textbf{.675$\pm$.035} & .689$\pm$.034 \\
    VOC07 & \textbf{.441$\pm$.005} & .458$\pm$.005 & .459$\pm$.006 & .454$\pm$.004 & .495$\pm$.006 & .454$\pm$.004 & \underline{.445$\pm$.004} & .474$\pm$.005 \\
    MIRFlickr & \textbf{.582$\pm$.007} & .635$\pm$.007 & .617$\pm$.011 & .592$\pm$.006 & .640$\pm$.007 & \underline{.585$\pm$.005} & .587$\pm$.006 & .592$\pm$.006 \\
    SCENE & \textbf{.430$\pm$.009} & .456$\pm$.008 & .439$\pm$.008 & .499$\pm$.010 & .448$\pm$.008 & .457$\pm$.009 & \underline{.435$\pm$.008} & .446$\pm$.009 \\
    OBJECT & \textbf{.311$\pm$.010} & .338$\pm$.008 & \underline{.319$\pm$.009} & .375$\pm$.008 & .320$\pm$.009 & .331$\pm$.008 & .323$\pm$.007 & .334$\pm$.009 \\
    Corel5K & \textbf{.491$\pm$.018} & .516$\pm$.018 & \underline{.515$\pm$.019} & .527$\pm$.017 & .571$\pm$.015 & .553$\pm$.017 & .545$\pm$.017 & .539$\pm$.017 \\
    IAPRTC12 & \textbf{.409$\pm$.012} & \underline{.427$\pm$.012} & .428$\pm$.012 & .462$\pm$.016 & .511$\pm$.011 & .458$\pm$.015 & .474$\pm$.015 & .432$\pm$.012 \\
    \bottomrule
  \end{tabularx}
\end{table*}
\begin{table*}[ht]
  \footnotesize
  \centering
  \caption{Experimental results $\downarrow$ (mean$\pm$std) of each method on HL; the best result is shown in bold and the second-best is underlined.}
  \label{tab:HL}
  \begin{tabularx}{\textwidth}{l@{\hspace{5pt}}*8{>{\centering\arraybackslash}X}}
    \toprule
    Datasets & SEHFS & DHLI & GRAFS & MSFS & MIFS & SPLDG & SRFS & MSSL \\
    \midrule
    EMOTIONS & \textbf{.243$\pm$.031} & .297$\pm$.032 & .297$\pm$.031 & .332$\pm$.059 & .275$\pm$.031 & .271$\pm$.032 & \underline{.255$\pm$.028} & .274$\pm$.039 \\
    YEAST & \textbf{.223$\pm$.007} & .225$\pm$.009 & .230$\pm$.007 & .254$\pm$.013 & .224$\pm$.009 & .230$\pm$.008 & \underline{.223$\pm$.009} & .230$\pm$.008 \\
    VOC07 & \textbf{.079$\pm$.004} & .085$\pm$.004 & .085$\pm$.005 & .087$\pm$.005 & .085$\pm$.004 & \underline{.084$\pm$.004} & \underline{.084$\pm$.004} & .086$\pm$.004 \\
    MIRFlickr & \textbf{.167$\pm$.010} & .189$\pm$.007 & .184$\pm$.010 & .175$\pm$.009 & .188$\pm$.007 & .173$\pm$.007 & \underline{.172$\pm$.009} & .176$\pm$.009 \\
    SCENE & \textbf{.092$\pm$.007} & .103$\pm$.008 & .098$\pm$.007 & .136$\pm$.019 & .102$\pm$.006 & .100$\pm$.006 & \underline{.097$\pm$.007} & .099$\pm$.006 \\
    OBJECT & \textbf{.052$\pm$.005} & .059$\pm$.004 & .057$\pm$.004 & .069$\pm$.004 & .060$\pm$.005 & .057$\pm$.004 & \underline{.056$\pm$.004} & .057$\pm$.004 \\
    Corel5K & \textbf{.013$\pm$.001} & .014$\pm$.001 & .014$\pm$.001 & .015$\pm$.001 & \underline{.013$\pm$.001} & .014$\pm$.001 & .014$\pm$.001 & .014$\pm$.001 \\
    IAPRTC12 & \textbf{.017$\pm$.002} & .018$\pm$.002 & .018$\pm$.002 & .019$\pm$.002 & \underline{.018$\pm$.002} & .018$\pm$.002 & .018$\pm$.002 & .018$\pm$.002 \\
    \bottomrule
  \end{tabularx}
\end{table*}
\begin{table*}[ht]
  \footnotesize
  \centering
  \caption{Experimental results $\downarrow$ (mean$\pm$std) of each method on RL; the best result is shown in bold and the second-best is underlined.}
  \label{tab:RL}
  \begin{tabularx}{\textwidth}{l@{\hspace{5pt}}*8{>{\centering\arraybackslash}X}}
    \toprule
    Datasets & SEHFS & DHLI & GRAFS & MSFS & MIFS & SPLDG & SRFS & MSSL \\
    \midrule
    EMOTIONS & \textbf{.272$\pm$.051} & .383$\pm$.056 & .362$\pm$.043 & .395$\pm$.062 & .330$\pm$.054 & .316$\pm$.053 & \underline{.291$\pm$.047} & .333$\pm$.057 \\
    YEAST & \underline{.249$\pm$.015} & .255$\pm$.018 & .254$\pm$.015 & .278$\pm$.038 & .254$\pm$.019 & .256$\pm$.015 & \textbf{.247$\pm$.016} & .259$\pm$.016 \\
    VOC07 & \textbf{.194$\pm$.017} & .206$\pm$.015 & .206$\pm$.019 & .204$\pm$.014 & .225$\pm$.018 & .205$\pm$.013 & \underline{.194$\pm$.014} & .215$\pm$.014 \\
    MIRFlickr & \textbf{.149$\pm$.014} & .180$\pm$.010 & .173$\pm$.017 & .159$\pm$.013 & .183$\pm$.011 & .156$\pm$.009 & \underline{.154$\pm$.012} & .159$\pm$.012 \\
    SCENE & \textbf{.085$\pm$.011} & .099$\pm$.011 & .091$\pm$.011 & .136$\pm$.025 & .097$\pm$.011 & .100$\pm$.011 & \underline{.090$\pm$.010} & .093$\pm$.011 \\
    OBJECT & \textbf{.173$\pm$.027} & .199$\pm$.025 & \underline{.182$\pm$.027} & .228$\pm$.025 & .183$\pm$.024 & .192$\pm$.024 & .186$\pm$.022 & .185$\pm$.026 \\
    Corel5K & \textbf{.234$\pm$.043} & .249$\pm$.044 & \underline{.247$\pm$.045} & .257$\pm$.044 & .292$\pm$.047 & .279$\pm$.051 & .272$\pm$.049 & .254$\pm$.044 \\
    IAPRTC12 & \textbf{.167$\pm$.015} & \underline{.178$\pm$.017} & .179$\pm$.015 & .199$\pm$.025 & .223$\pm$.017 & .196$\pm$.025 & .206$\pm$.025 & .183$\pm$.017 \\
    \bottomrule
  \end{tabularx}
\end{table*}
\subsection{Complexity Analysis}
We analyze the time complexity of SEHFS. Let $n$ denote the number of samples, $d = \sum_{v=1}^V d(v)$ the total feature dimension across all $V$ views, and $q$ the number of labels. The primary cost stems from five per-iteration components in the alternating optimization procedure (Algorithm~\ref{alg:algorithm}), preceded by a one-time initialization.

\paragraph{Initialization} Constructing the semantic matrices $\{\mathbf{S}_v\}_{v=1}^V$ and the mutual information matrix $\mathbf{A}$ requires pairwise similarity or information-theoretic computations. Computing $\mathbf{S}_v$ via $k$-nearest neighbors for a single view costs $\mathcal{O}(n^2 d(v))$, totaling $\mathcal{O}(n^2 d)$ over all views. Building the $d \times d$ mutual information matrix $\mathbf{A}$ by evaluating pairwise MI among features costs $\mathcal{O}(n d^2)$. These operations are performed once before iterations.

\paragraph{Iterative updates} Updating the global view matrix $\mathbf{X^f}$ (Formula~(\ref{eq:updateX})) involves multiplications of size $n \times n$ and $n \times d$, yielding a per-iteration cost of $\mathcal{O}(n^2 d)$. Updating the shared semantic matrix $\mathbf{S}$ (Formula~(\ref{eq:updateS})) operates on $n \times d$ matrices and, with precomputation, costs $\mathcal{O}(n^2 d)$ per iteration. Updating the assignment matrix $\mathbf{W}$ (Formula~(\ref{eq:updateW})) requires gradients for the structural entropy term and reconstruction loss; the structural-entropy gradient uses $\mathbf{X^f} \in \mathbb{R}^{n \times d}$ and $\mathbf{W} \in \mathbb{R}^{d \times q}$, costing $\mathcal{O}(n d q)$ per iteration. In the $v$-th view, updating $\mathbf{H}_v$ is dominated by matrix operations with complexity $\mathcal{O}(n d^2)$, while updating $\boldsymbol{\alpha}_v$ is dominated by solving a quadratic program with complexity $\mathcal{O}(n^2)$.

\paragraph{Overall complexity} Assuming convergence in $T$ iterations, the total time complexity is $\mathcal{O}\big(T(n^2 d + n d^2)\big)$, scaling linearly with $T$ and polynomially with $n$ and $d$.

\begin{figure*}[ht]
    \centering
    \hfill
    \begin{subfigure}[b]{\textwidth}
        \centering
        \includegraphics[width=0.85\textwidth]{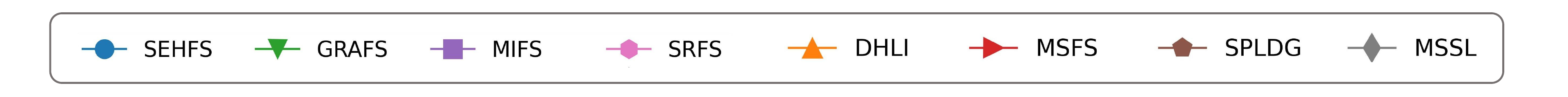}
    \end{subfigure}
    \hfill
    \begin{subfigure}[b]{0.231\textwidth}
        \centering
        \includegraphics[width=\textwidth]{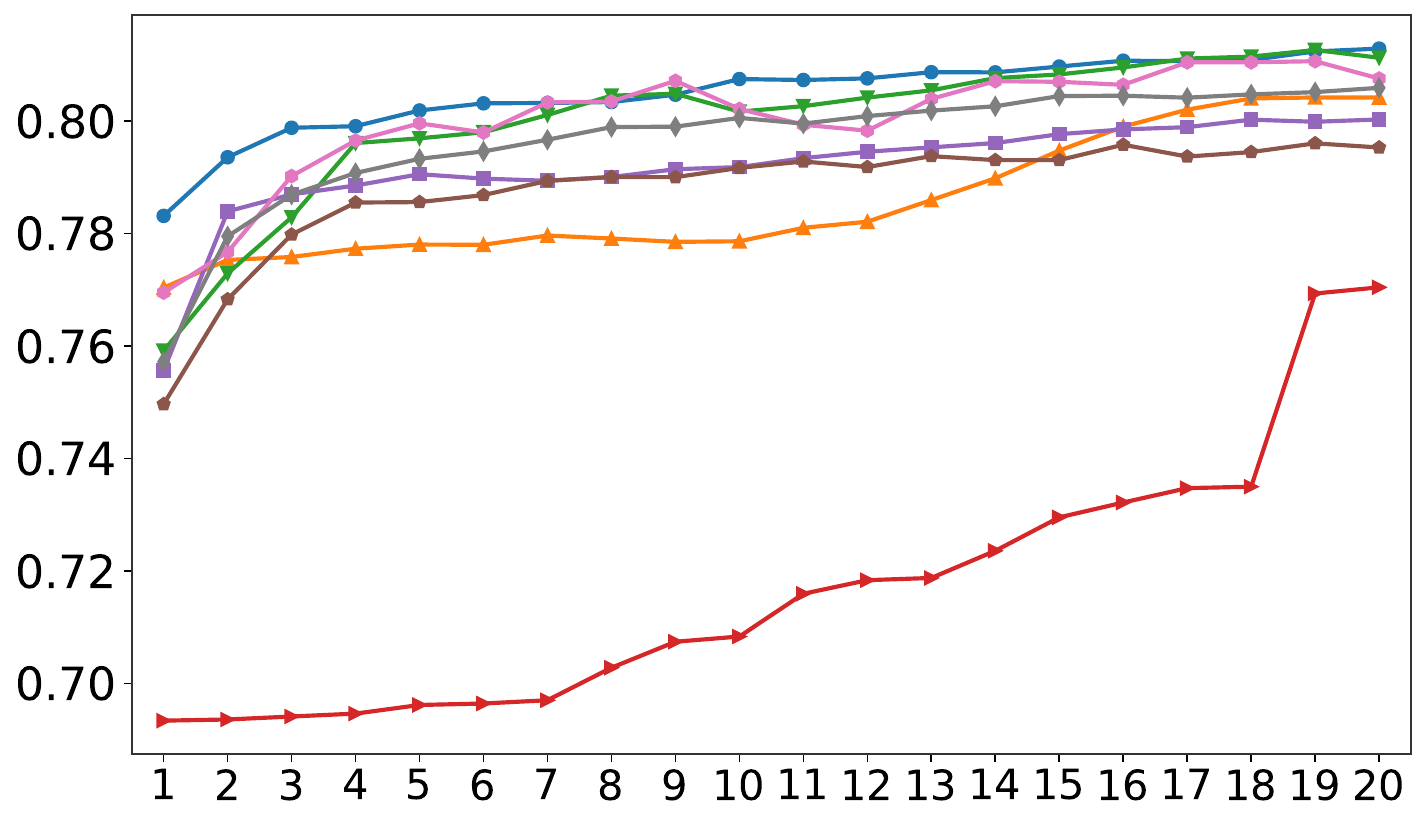}
        \caption{AP}
    \end{subfigure}
    \hfill
    \begin{subfigure}[b]{0.231\textwidth}
        \centering
        \includegraphics[width=\textwidth]{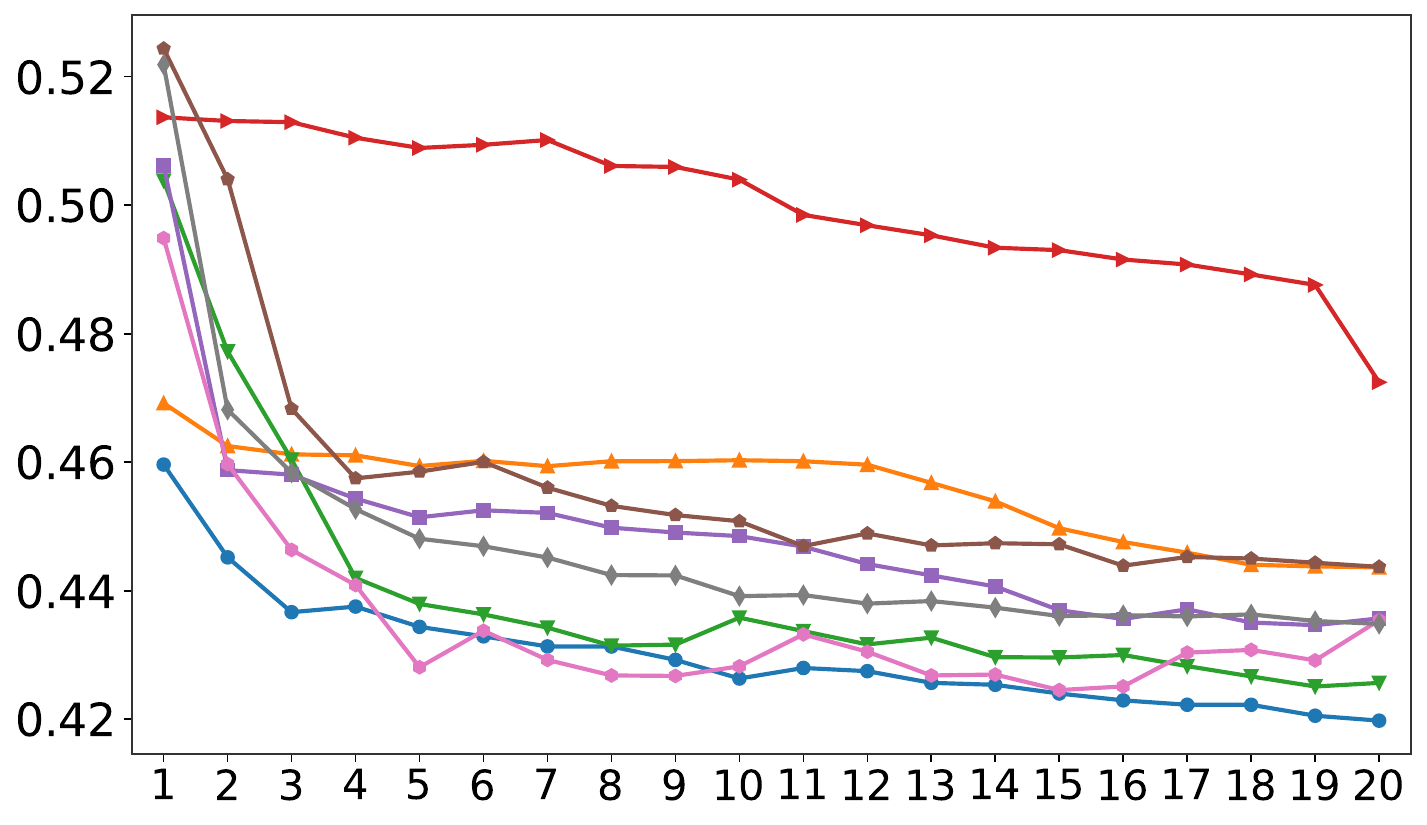}
        \caption{Cov}
    \end{subfigure}
    \hfill
    \begin{subfigure}[b]{0.231\textwidth}
        \centering
        \includegraphics[width=\textwidth]{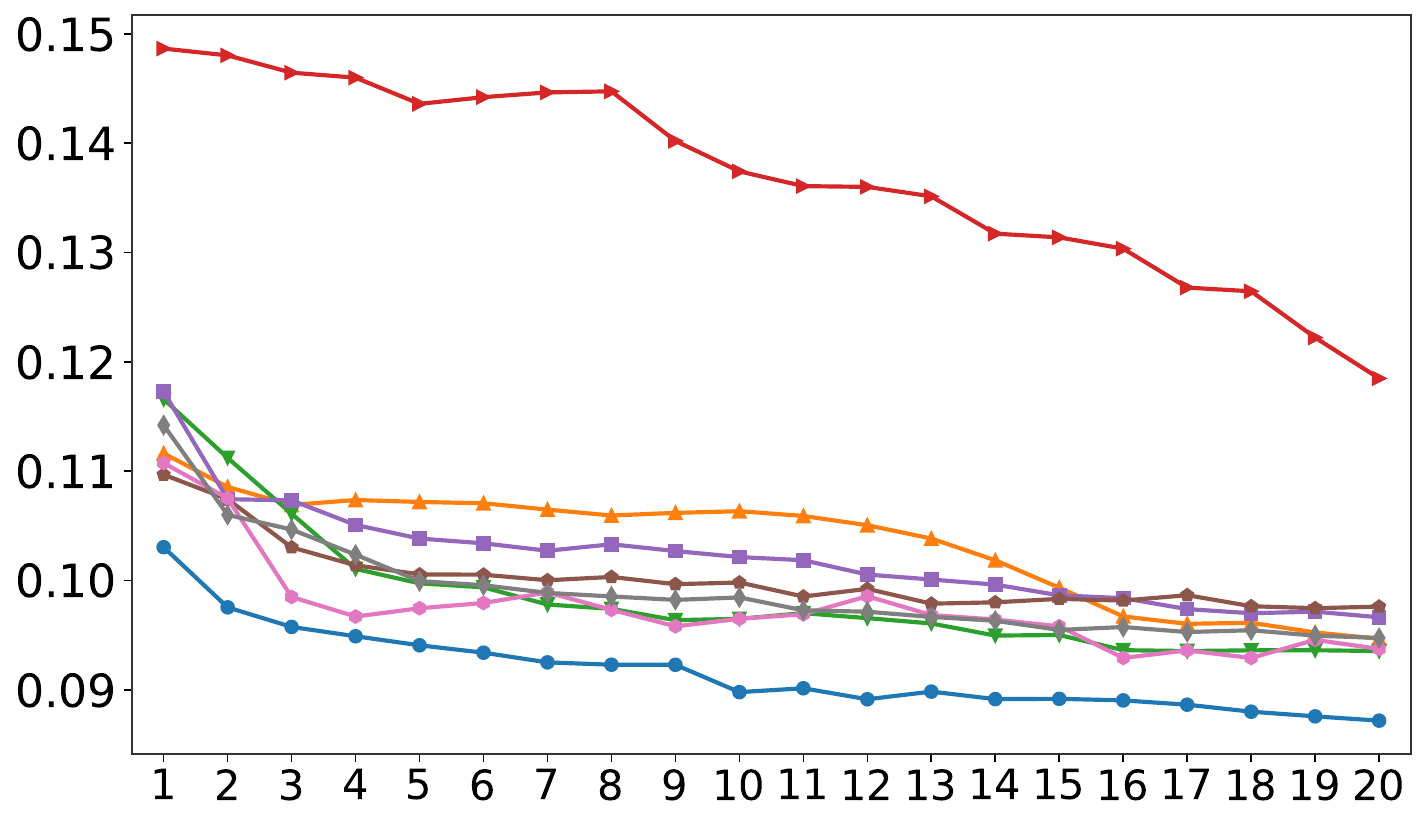}
        \caption{HL}
    \end{subfigure}
    \hfill
    \begin{subfigure}[b]{0.231\textwidth}
        \centering
        \includegraphics[width=\textwidth]{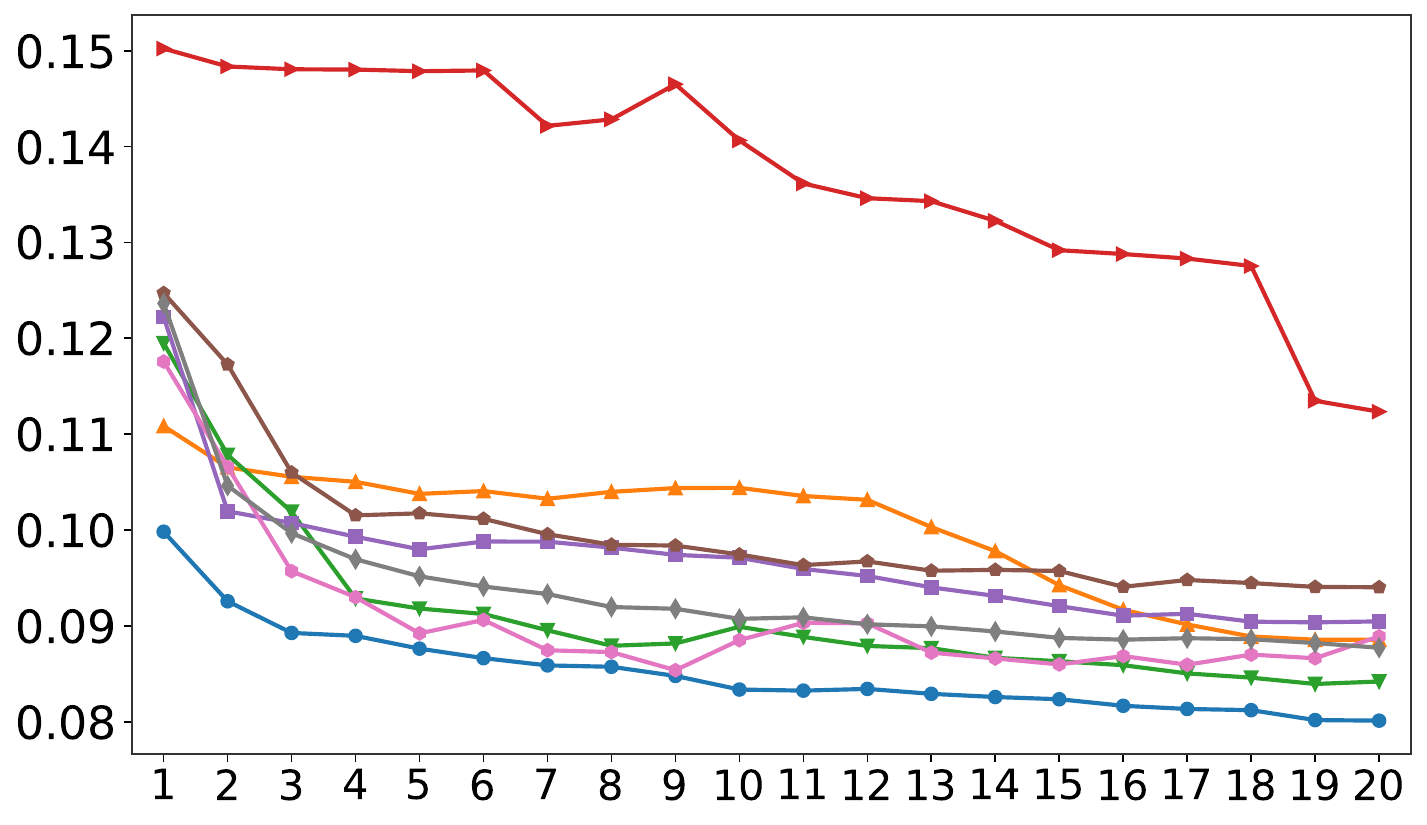}
        \caption{RL}
    \end{subfigure}
    \hfill
    \begin{subfigure}[b]{0.231\textwidth}
        \centering
        \includegraphics[width=\textwidth]{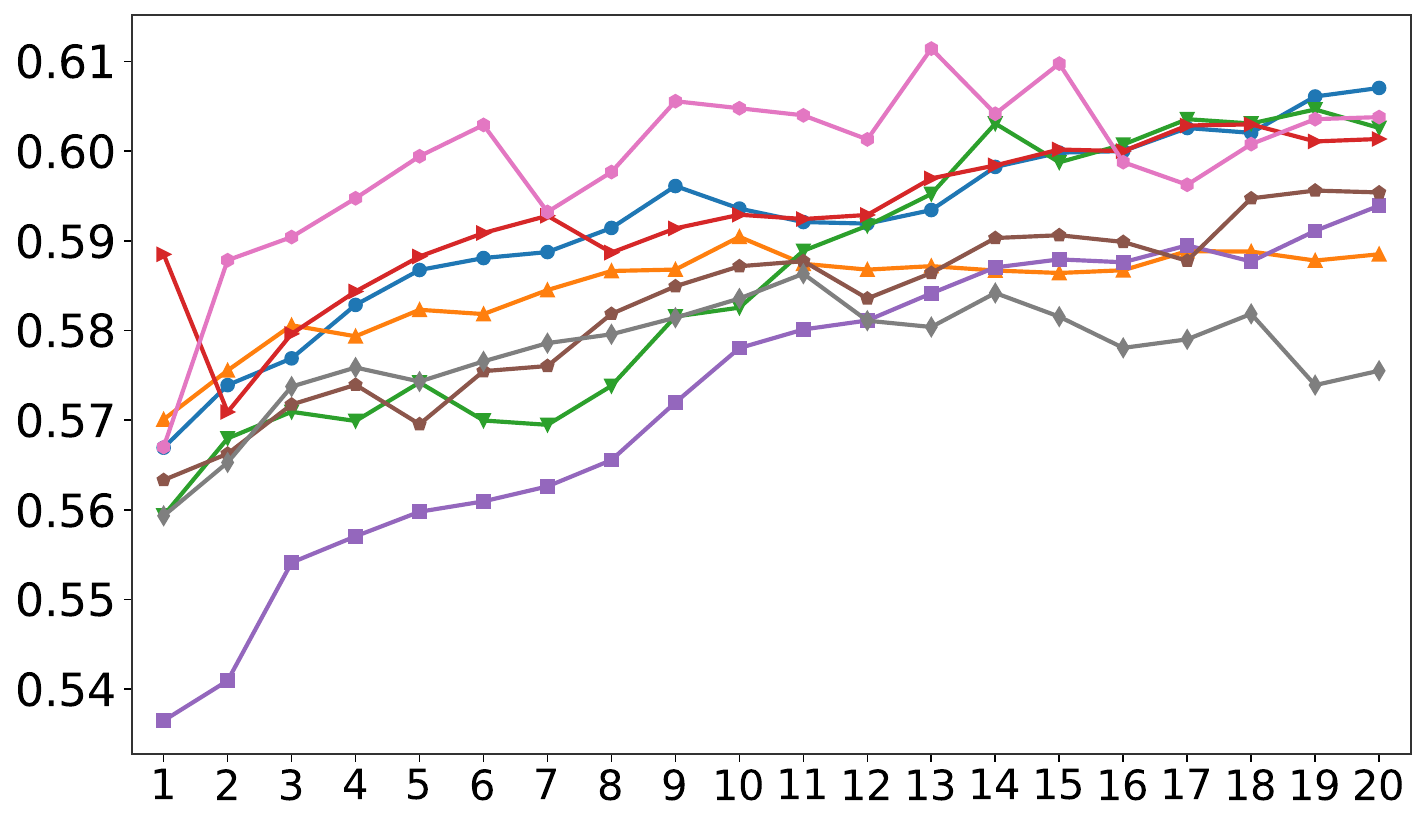}
        \caption{AP}
    \end{subfigure}
    \hfill
    \begin{subfigure}[b]{0.231\textwidth}
        \centering
        \includegraphics[width=\textwidth]{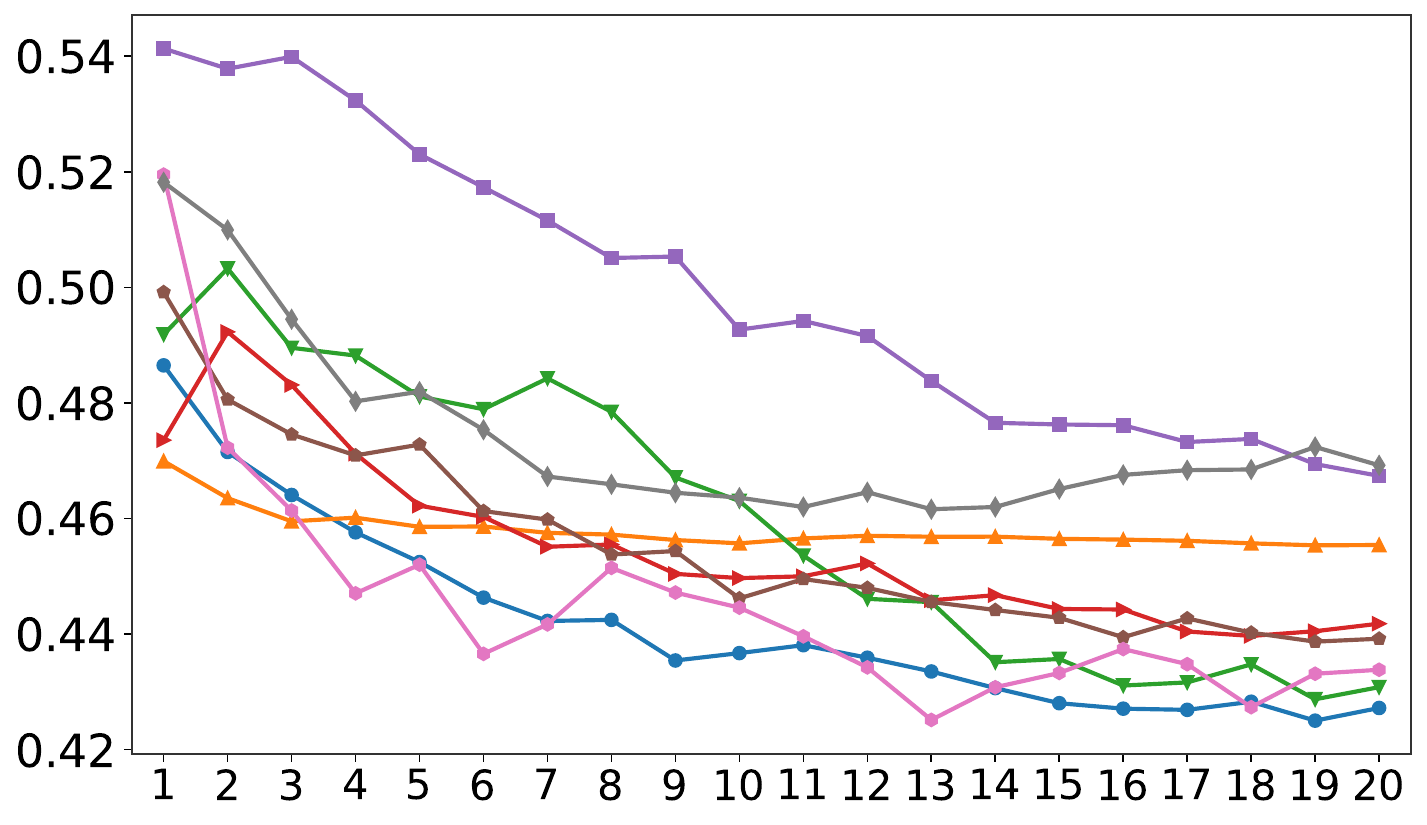}
        \caption{Cov}
    \end{subfigure}
    \hfill
    \begin{subfigure}[b]{0.231\textwidth}
        \centering
        \includegraphics[width=\textwidth]{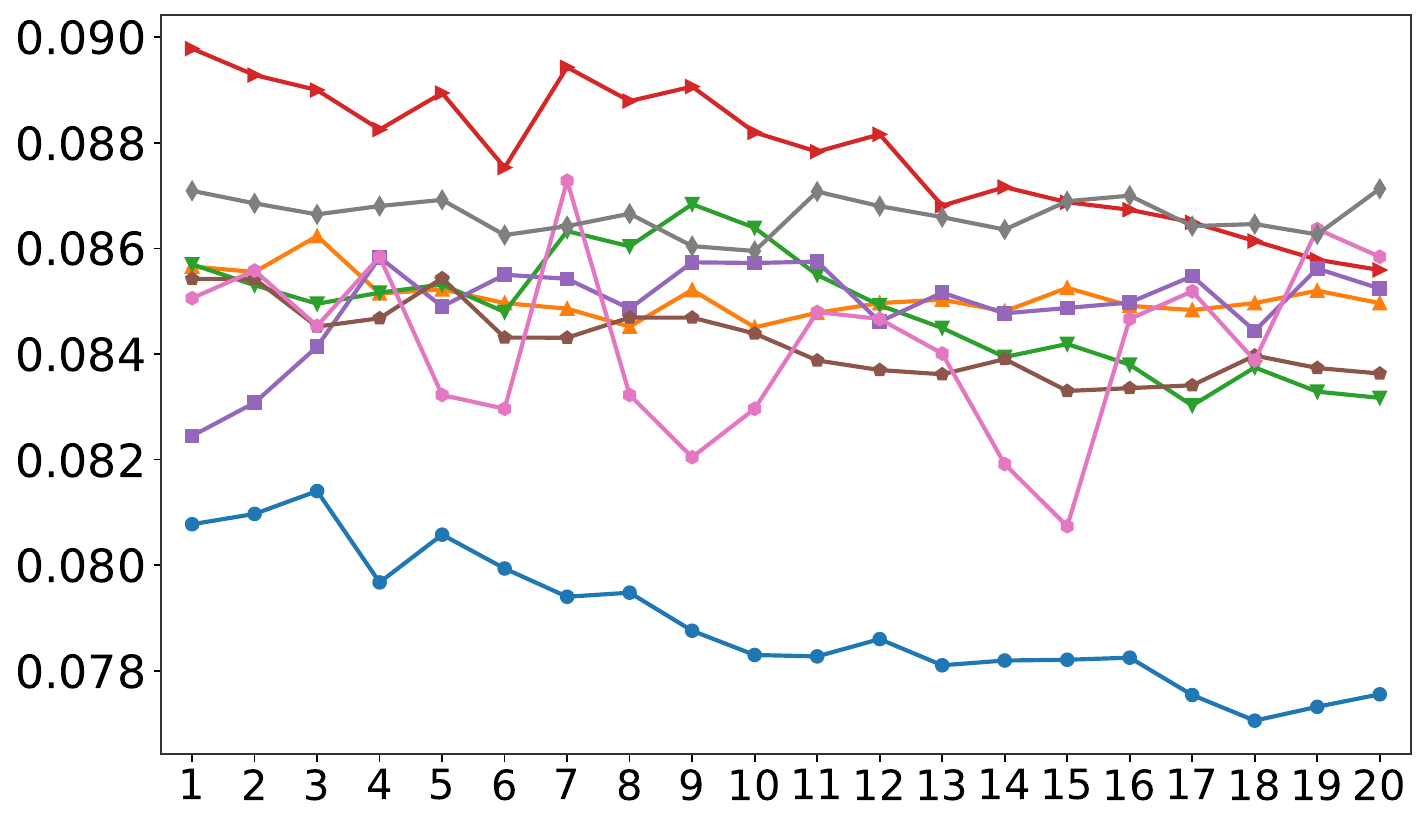}
        \caption{HL}
    \end{subfigure}
    \hfill
    \begin{subfigure}[b]{0.231\textwidth}
        \centering
        \includegraphics[width=\textwidth]{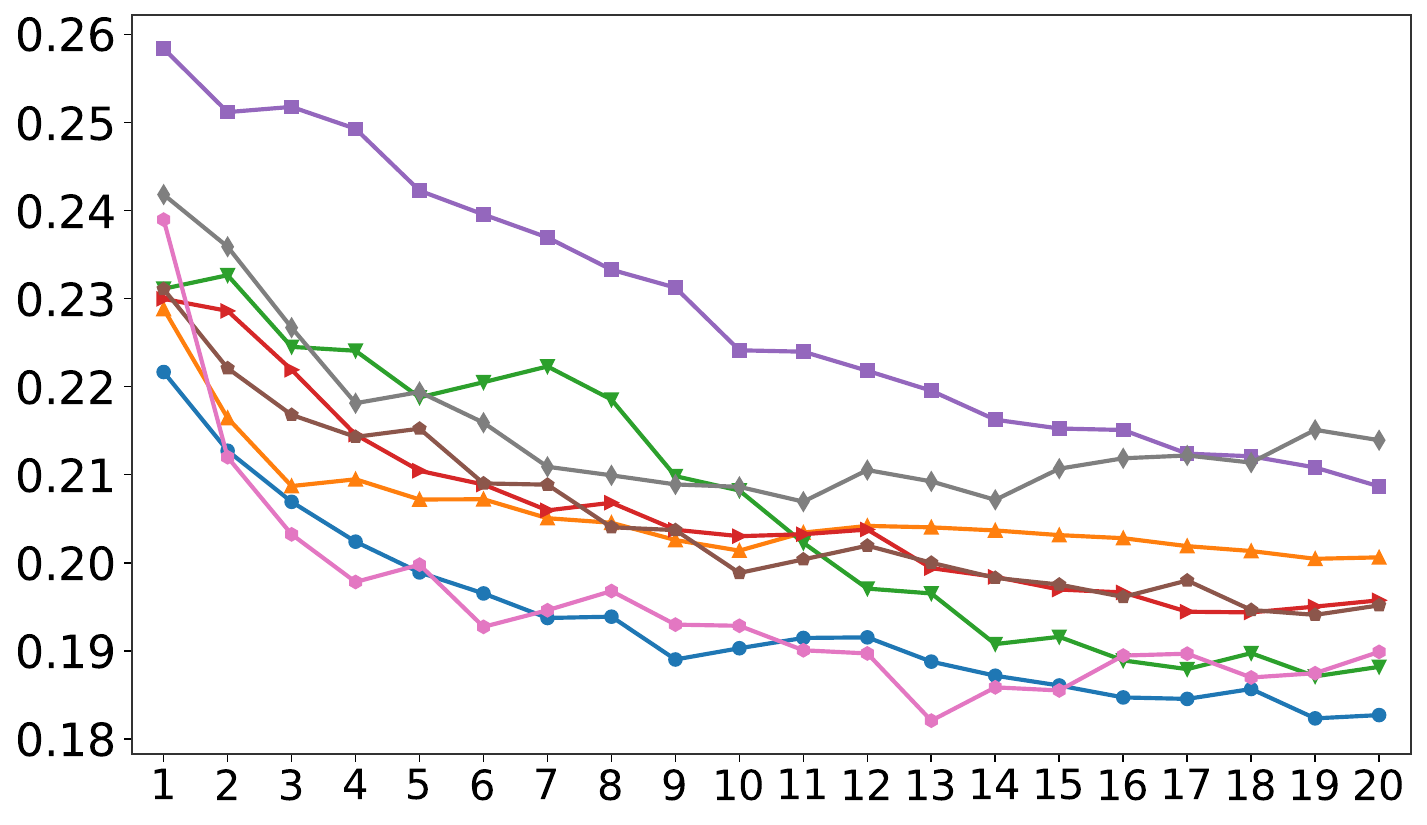}
        \caption{RL}
    \end{subfigure}
    \hfill
    \begin{subfigure}[b]{0.231\textwidth}
        \centering
        \includegraphics[width=\textwidth]{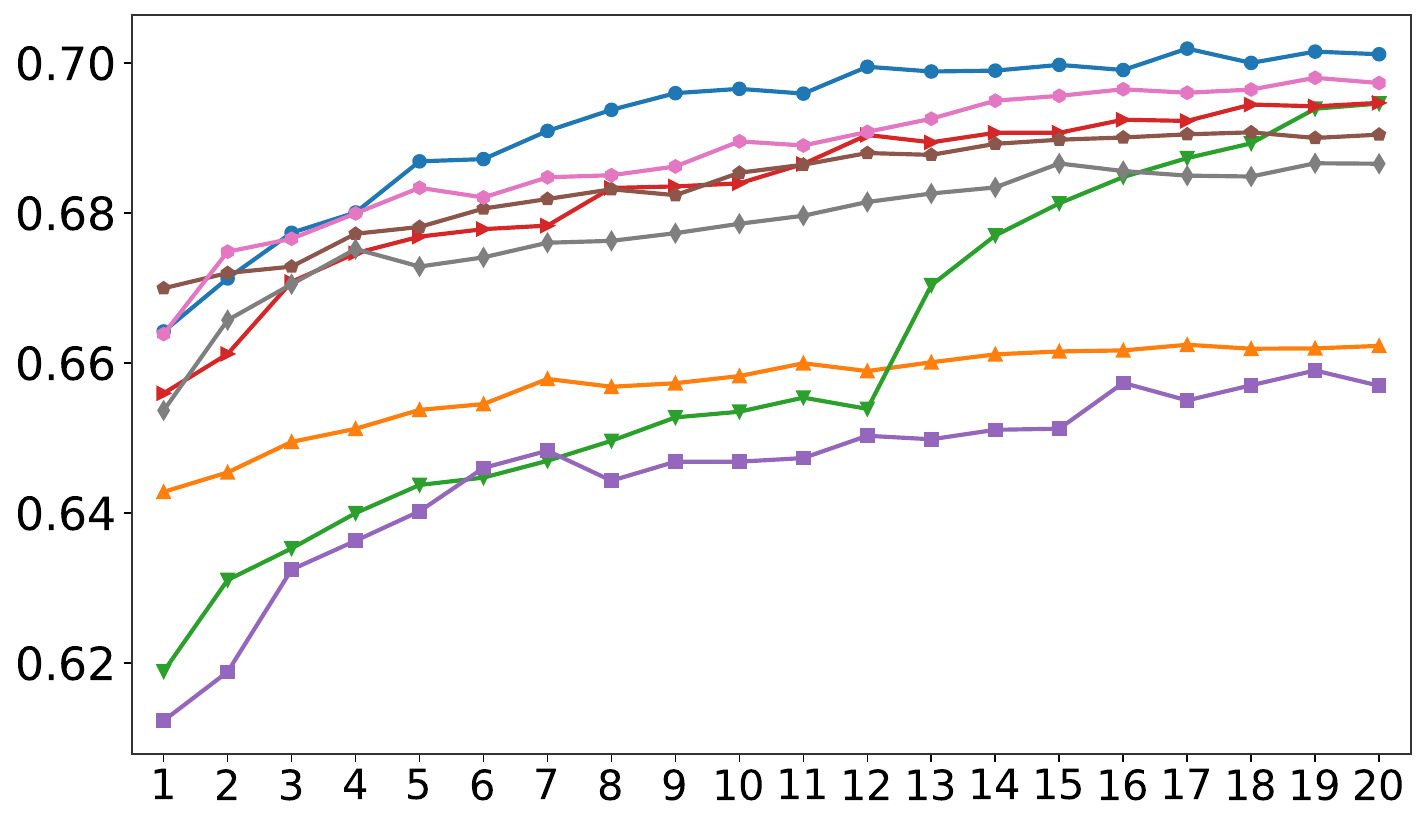}
        \caption{AP}
    \end{subfigure}
    \hfill
    \begin{subfigure}[b]{0.231\textwidth}
        \centering
        \includegraphics[width=\textwidth]{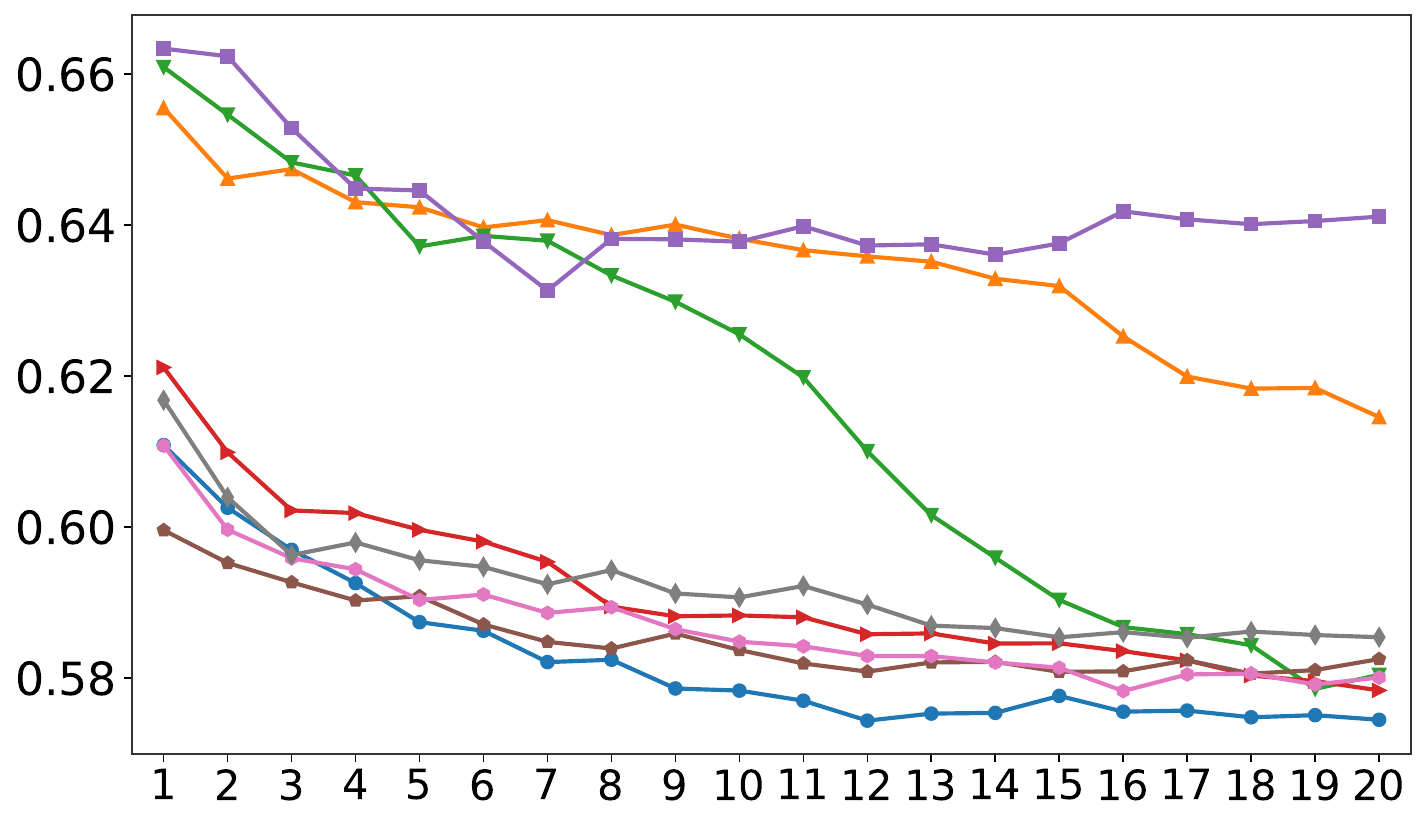}
        \caption{Cov}
    \end{subfigure}
    \hfill
    \begin{subfigure}[b]{0.231\textwidth}
        \centering
        \includegraphics[width=\textwidth]{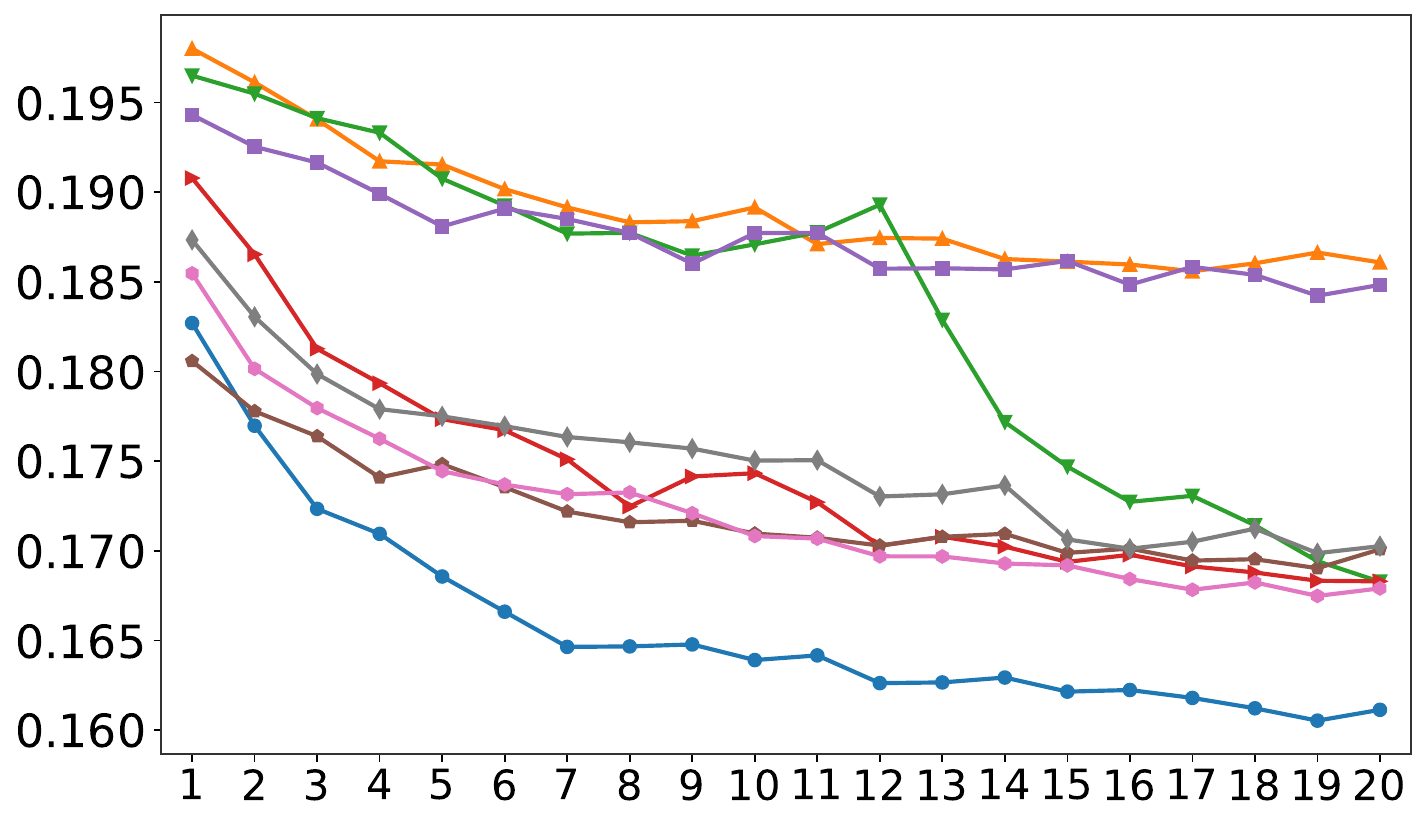}
        \caption{HL}
    \end{subfigure}
    \hfill
    \begin{subfigure}[b]{0.231\textwidth}
        \centering
        \includegraphics[width=\textwidth]{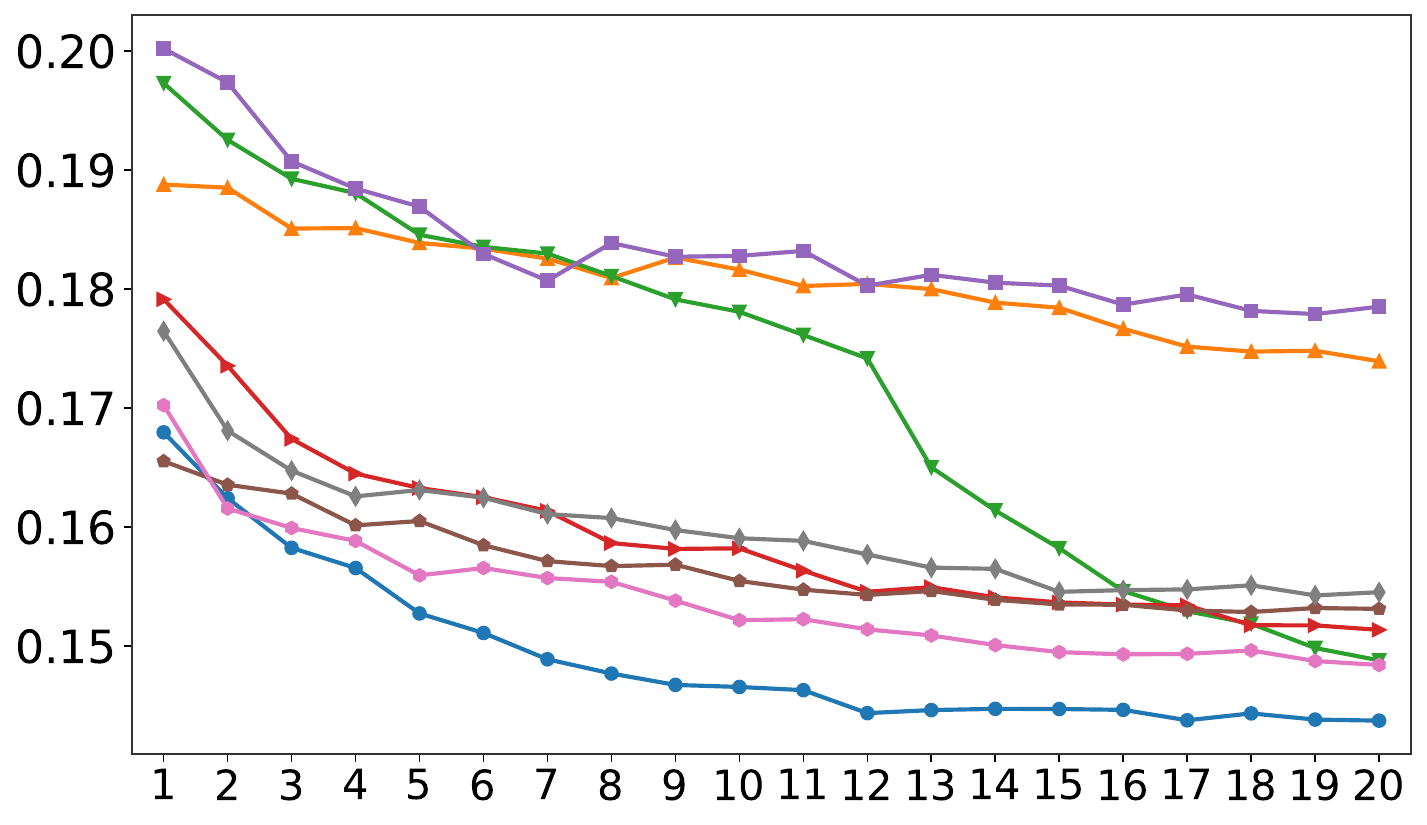}
        \caption{RL}
    \end{subfigure}
    \hfill
    \caption{Performance comparison of SEHFS and seven baseline methods on SCENE (first row), VOC07 (second row), and MIRFlickr (third row). The horizontal axis denotes the number of selected features, and the vertical axis reports AP, Cov, HL, and RL.}
    \label{fig:features}
\end{figure*}
\begin{figure*}[ht]
    \centering
    \begin{subfigure}[b]{0.241\textwidth}
        \centering
        \includegraphics[width=\textwidth]{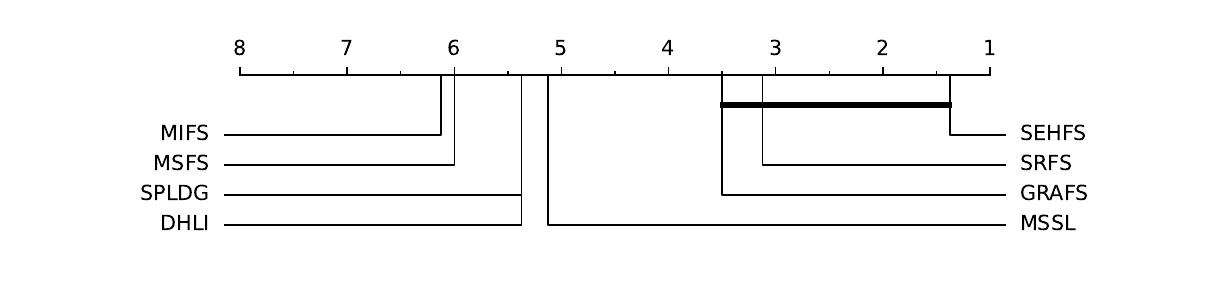}
        \caption{Average Precision}
    \end{subfigure}
    \hfill
    \begin{subfigure}[b]{0.241\textwidth}
        \centering
        \includegraphics[width=\textwidth]{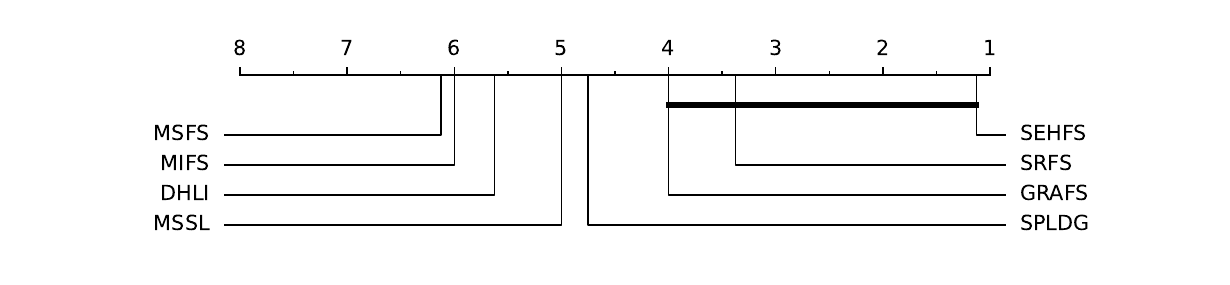}
        \caption{Coverage}
    \end{subfigure}
    \hfill
    \begin{subfigure}[b]{0.241\textwidth}
        \centering
        \includegraphics[width=\textwidth]{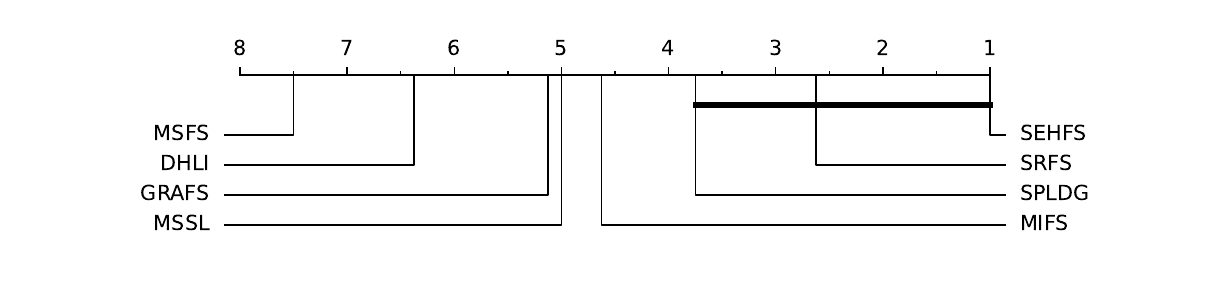}
        \caption{Hamming Loss}
    \end{subfigure}
    \hfill
    \begin{subfigure}[b]{0.241\textwidth}
        \centering
        \includegraphics[width=\textwidth]{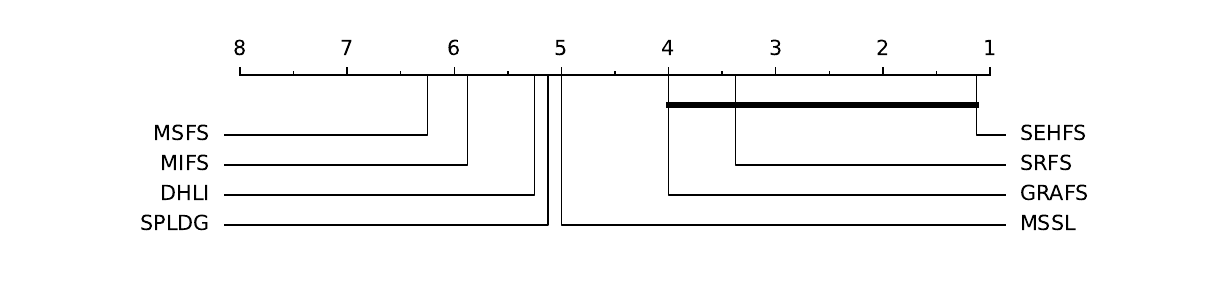}
        \caption{Ranking Loss}
    \end{subfigure}
    \caption{Comparison of SEHFS with other methods using the Bonferroni-Dunn test. Methods not connected to SEHFS indicate a significant difference in performance (CD = 3.2946 at the significance level of 0.05).}
    \label{fig:cd}
\end{figure*}
\section{Experiments}
\label{sec:experiments}
\subsection{Datasets}
Our method is evaluated on eight widely used multi-view multi-label datasets from different domains: EMOTIONS~\cite{tsoumakas2008effective} for music emotion classification, YEAST~\cite{elisseeff2001kernel} for gene function prediction, VOC07~\cite{everingham2010pascal} for image classification, MIRFlickr~\cite{huiskes2008mir} for multimedia annotation, SCENE~\cite{boutell2004learning} and OBJECT~\cite{chua2009nus} for image retrieval, Corel5K~\cite{duygulu2002object} and IAPRTC12~\cite{escalante2010segmented} for image annotation. The datasets contain varying numbers of views, ranging from 2 to 5, and the number of features per view also varies from tens to thousands. The dataset details are presented in Table~\ref{tab:datasets}, where `-' is used to denote the absence of the corresponding view in the dataset.

\subsection{Experimental setup}
\paragraph{Evaluation Metrics}
Four widely used multi-view multi-label evaluation metrics are adopted: Average Precision (AP), Coverage (Cov), Hamming Loss (HL), and Ranking Loss (RL). For AP, larger values indicate better performance and are denoted by $\uparrow$ in the table. For Cov, HL, and RL, smaller values indicate better performance and are denoted by $\downarrow$ in the table.  \newtext{SEHFS contains four trade-off parameters, namely $\alpha$, $\beta$, $\gamma$, and $\lambda$, all of which are tuned within the range of $\{10^{-3}, 10^{-2}, \dots, 10^{3}\}$. For graph construction in Formula~(\ref{eq:tr}), we fix the number of nearest neighbors to $k=50$, and set the Gaussian kernel parameter $\sigma$ as the mean of all non-zero pairwise distances in the corresponding view.}

\paragraph{Baselines}
Seven state-of-the-art methods are introduced as baselines in the experiments: three multi-view multi-label feature selection methods (\textbf{DHLI}~\cite{hao2024double}, \textbf{GRAFS}~\cite{hao2024anchor}, \textbf{MSFS}~\cite{zhang2020multi}) and four multi-label feature selection methods (\textbf{SRFS}~\cite{li2025fusion}, \textbf{SPLDG}~\cite{zhang2025sparse}, \textbf{MSSL}~\cite{cai2018multi}, \textbf{MIFS}~\cite{jian2016multi}). The top $20\%$ of features are uniformly selected for comparison using the MLkNN classifier~\cite{zhang2007ml}. Results are presented as mean evaluation scores with standard deviations obtained by ten-fold cross-validation. The parameters for each method are tuned within the range of $\{10^{-3}, 10^{-2}, \dots, 10^3\}$ to ensure a fair and valid comparison.

\subsection{Experimental result}
The detailed results for AP, Cov, HL, and RL are presented in Table~\ref{tab:AP},~\ref{tab:Cov},~\ref{tab:HL} and~\ref{tab:RL}, where the mean results and standard deviations are recorded. According to the comparisons of 224 statistical results, some key observations are clearly revealed:
\begin{itemize}
    \item Among four evaluation metrics across all eight datasets, the proposed method SEHFS achieved the best evaluation in $87.5\%$ of cases. Among the remaining cases, all are ranked second-best except for the AP metric on VOC07. Regarding the HL metric specifically, SEHFS achieved the best evaluation with a perfect $100\%$ rate. This indicates that minimizing the structural entropy regularization term learns high-order feature correlations and directly selects more suitable and low-redundancy feature support for each label.
    \item Among all baselines, SEHFS is observed to outperform the three multi-view multi-label methods across all metrics on all datasets, except on the AP metric for VOC07. On YEAST and VOC07, which have fewer views and a smaller scale, no advantage over the multi-label learning methods SPLDG and SRFS is observed. In contrast, on SCENE, OBJECT, Corel5K, and IAPRTC12, which have more views and a larger scale, SEHFS achieves the best performance across all metrics and outperforms all seven baselines. These results demonstrate the effectiveness of SEHFS for multi-view learning. Moreover, on these four larger datasets, SEHFS exceeds the best results of baselines by $4.80\%$ on average, indicating that the structural entropy regularization term effectively learns high-order feature correlations from more complex data structures and provides robustness to noise.
    \item A more intuitive comparison among all eight methods on SCENE at different percentages of selected features is presented in Figure~\ref{fig:features}. As the proportion of selected features increases, the superiority of SEHFS and the effectiveness of feature selection are clearly observed.
\end{itemize}
\vspace{-0.05em}
The Friedman test~\cite{demvsar2006statistical} is employed on all four metrics to evaluate the relative performance of the baselines, and Table~\ref{tab:Friedman} reports the Friedman statistic $F_F$ for each metric together with the corresponding critical value. From Table~\ref{tab:Friedman}, it is evident that the null hypothesis, which assumes no difference in model performance between the methods, is rejected at the significance level ($\alpha = 0.05$). Consequently, a significant difference between the methods is indicated.
\begin{table}[t]
\small
\centering
\caption{Friedman statistics $F_F$ over four evaluation metrics, along with the critical value at a 0.05 significance level for SEHFS.}
\label{tab:Friedman}
\begin{tabularx}{0.45\textwidth}{l@{\hspace{20pt}}X@{\hspace{-1pt}}X}
\toprule
Evaluation Metric & $F_F$     & Critical value          \\ \midrule
Average Precision & 6.0485    & \multirow{4}{*}{2.2032} \\
Coverage          & 5.9945    &                         \\
Hamming Loss      & 16.5200   &                         \\
Ranking Loss      & 5.8174    &                         \\ \bottomrule
\end{tabularx}
\end{table}
\vspace{-0.1em}
Following the rejection of the null hypothesis in the Friedman test, the Bonferroni-Dunn test~\cite{demvsar2006statistical}, a non-parametric statistical hypothesis testing method, is employed for pairwise comparisons. The critical value $q_{\mathfrak{a}} = 2.690$ and the Critical Distance $CD = 3.2946$ can be calculated at a significance level of $\mathfrak{a} = 0.05$. If the difference between the average ranks of two methods is greater than $CD$, it indicates a significant difference in their performance. The results of this test are presented in Figure~\ref{fig:cd}, from which it can be inferred that SEHFS outperforms five baselines among them across all four metrics.
\begin{figure}[t]
    \centering
    \includegraphics[width=0.62\linewidth]{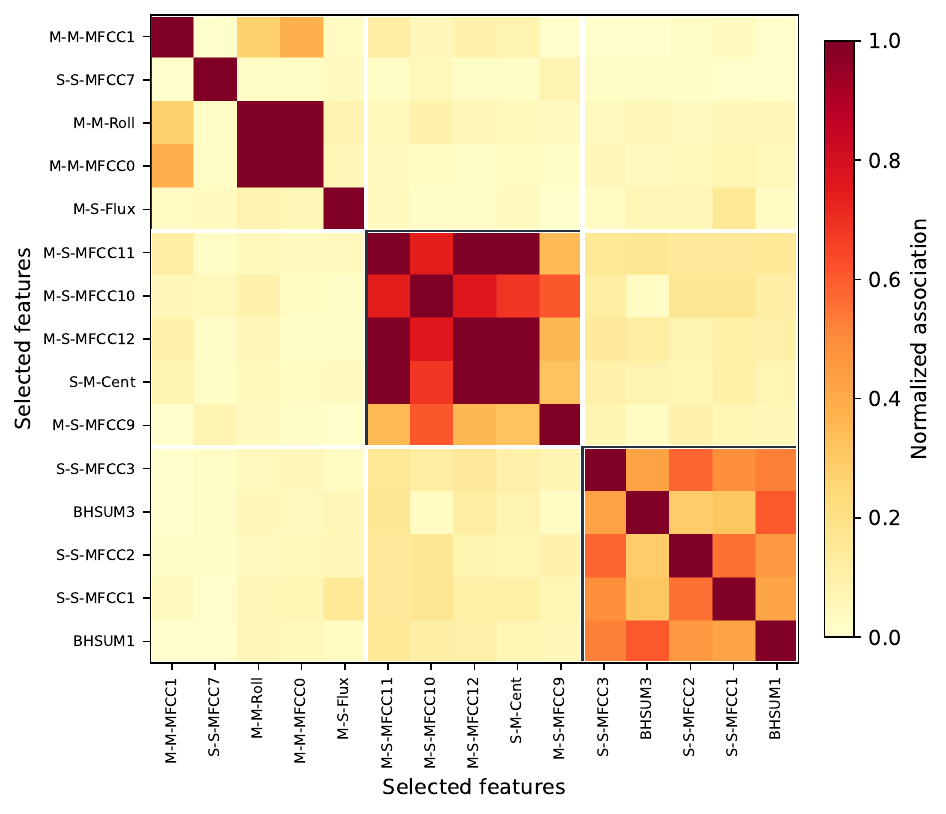}
    \caption{\newtext{Reordered feature graph heatmap according to the encoding-tree clustering learned by SEHFS on the Emotions dataset.}}
    \label{fig:tree_visualization}
\end{figure}
\vspace{-0.05em}
To illustrate the feature organization learned by SEHFS, we visualize the encoding-tree clustering results on the Emotions dataset. As shown in Fig.~\ref{fig:tree_visualization}, the selected features are reordered according to their assignments to the intermediate encoding-tree nodes. The reordered feature graph shows a clear block-wise pattern, where features within the same cluster have stronger associations than those across different clusters. This indicates that the structural entropy term encourages coherent modular organization in the feature graph.

We further inspect one representative cluster, which contains $S$-$S$-MFCC7, $M$-$M$-MFCC1, $M$-$M$-Rolloff, $M$-$M$-MFCC12, and $M$-$M$-MFCC0. These features are acoustically related: MFCCs describe the spectral envelope and timbre characteristics at different cepstral orders, while spectral rolloff reflects high-frequency energy distribution. Their grouping suggests that SEHFS learns data-driven structural dependencies rather than manually predefined feature categories.

\begin{figure*}[ht]
    \centering
    \begin{subfigure}[b]{0.231\textwidth} 
        \centering
        \includegraphics[width=\textwidth]{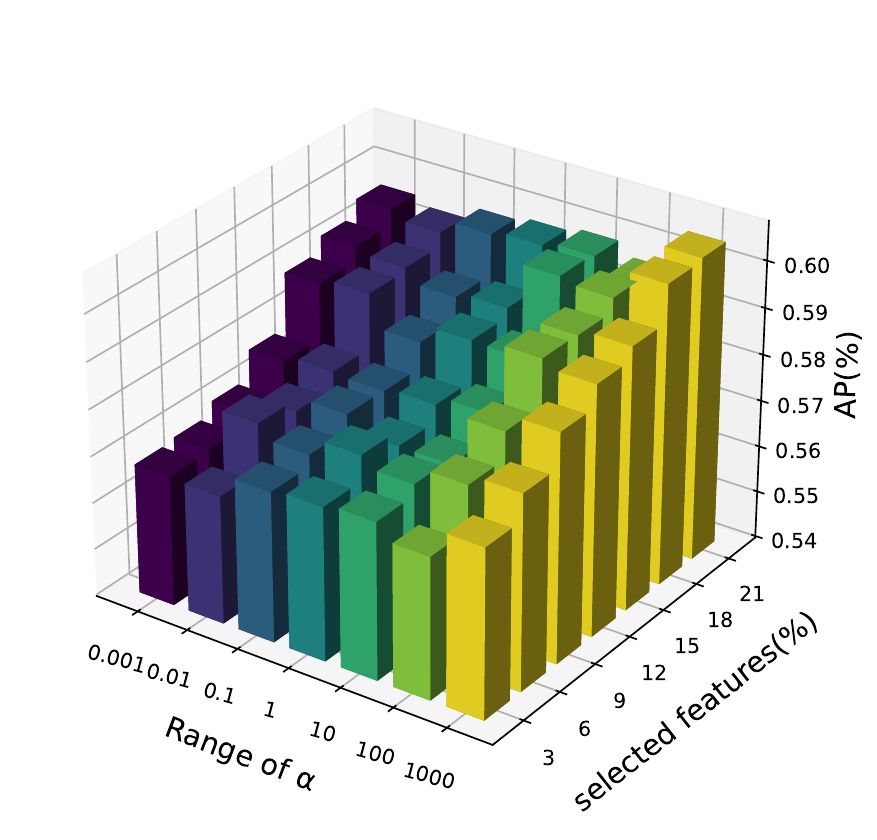}
        \caption{$\alpha$}
    \end{subfigure}
    \hfill
    \begin{subfigure}[b]{0.231\textwidth}
        \centering
        \includegraphics[width=\textwidth]{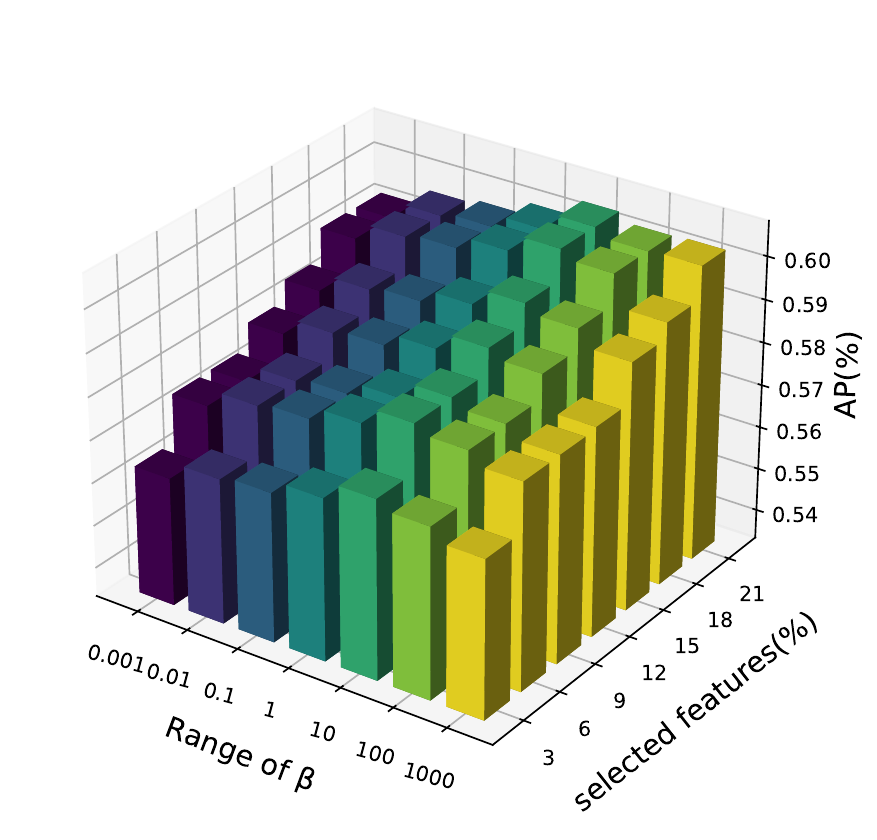}
        \caption{$\beta$}
    \end{subfigure}
    \hfill
    \begin{subfigure}[b]{0.231\textwidth}
        \centering
        \includegraphics[width=\textwidth]{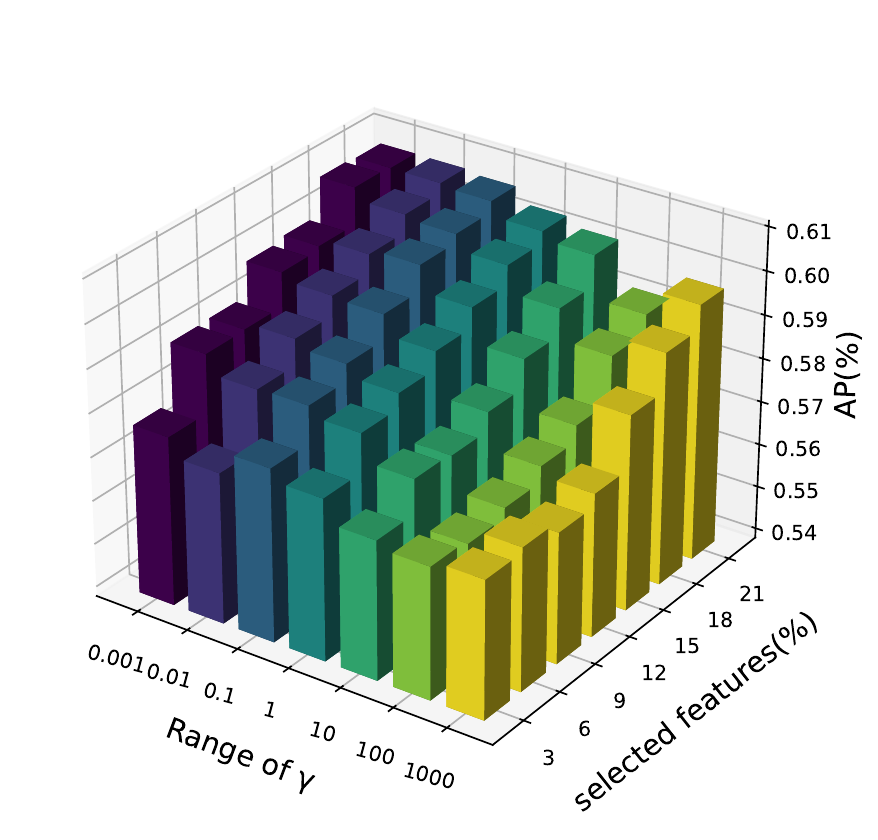}
        \caption{$\gamma$}
    \end{subfigure}
    \hfill
    \begin{subfigure}[b]{0.231\textwidth}
        \centering
        \includegraphics[width=\textwidth]{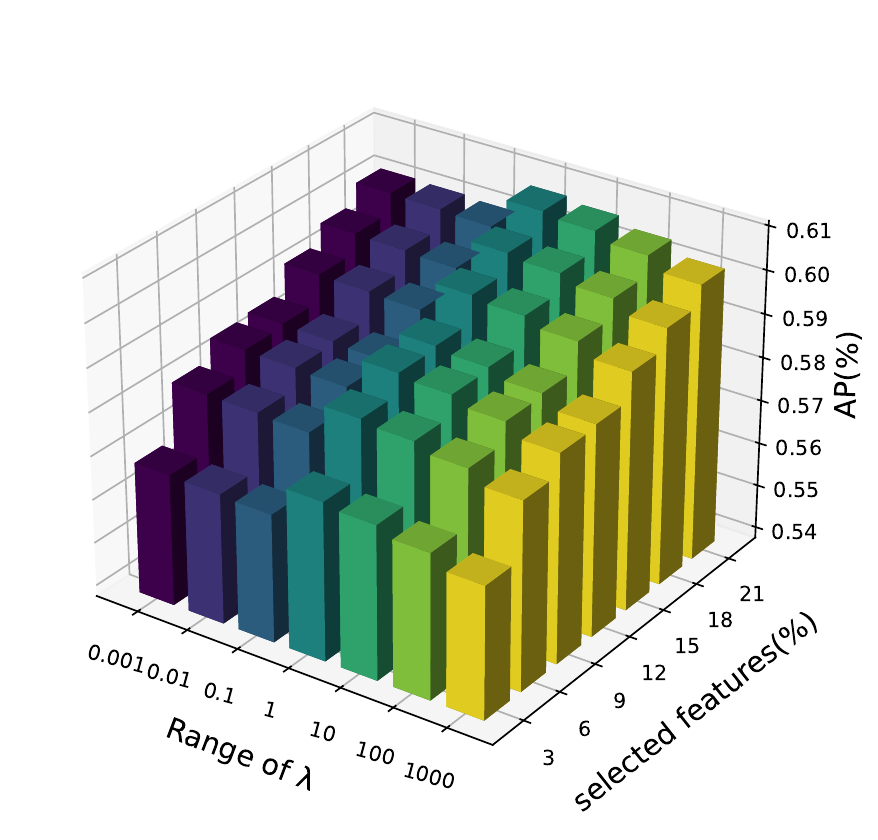}
        \caption{$\lambda$}
    \end{subfigure}
    \hfill
    \begin{subfigure}[b]{0.231\textwidth} 
        \centering
        \includegraphics[width=\textwidth]{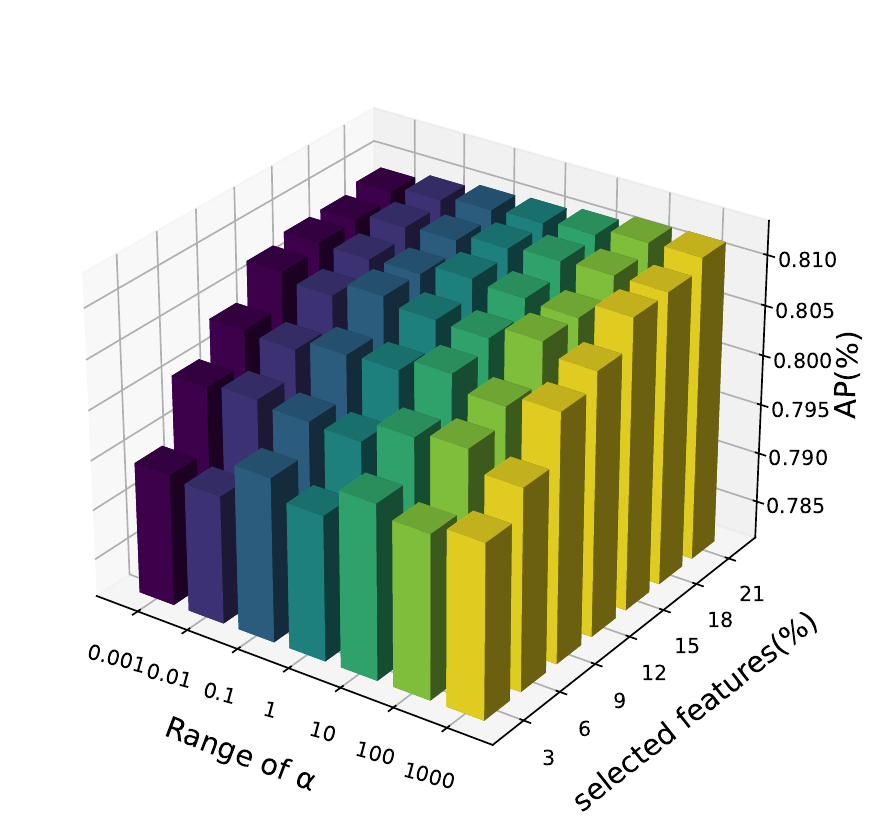}
        \caption{$\alpha$}
    \end{subfigure}
    \hfill
    \begin{subfigure}[b]{0.231\textwidth}
        \centering
        \includegraphics[width=\textwidth]{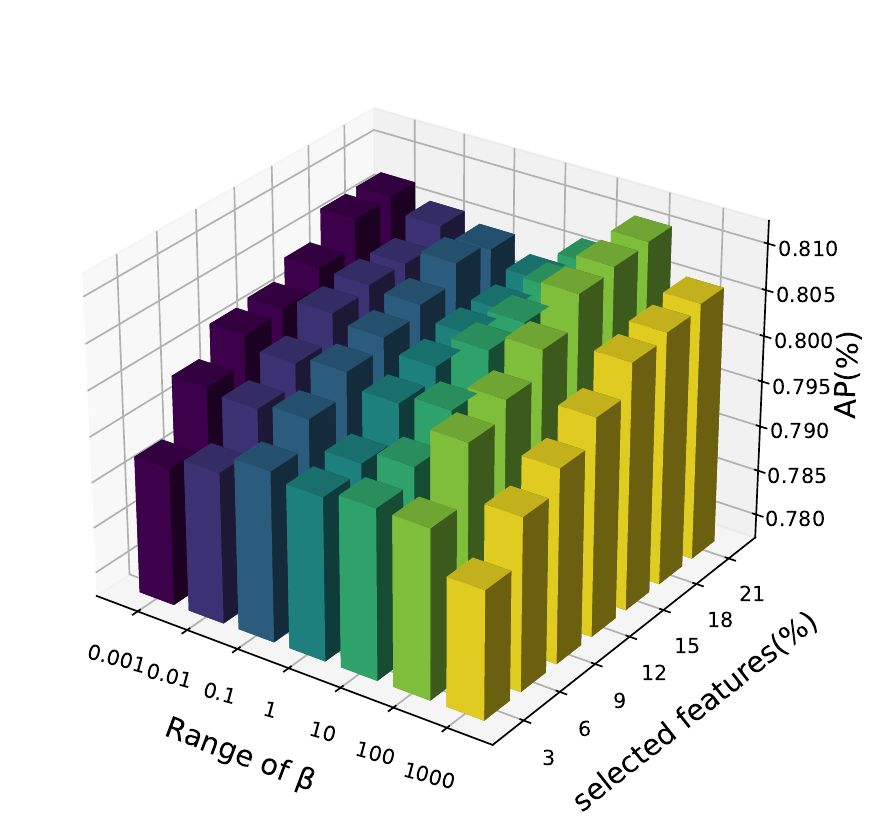}
        \caption{$\beta$}
    \end{subfigure}
    \hfill
    \begin{subfigure}[b]{0.231\textwidth}
        \centering
        \includegraphics[width=\textwidth]{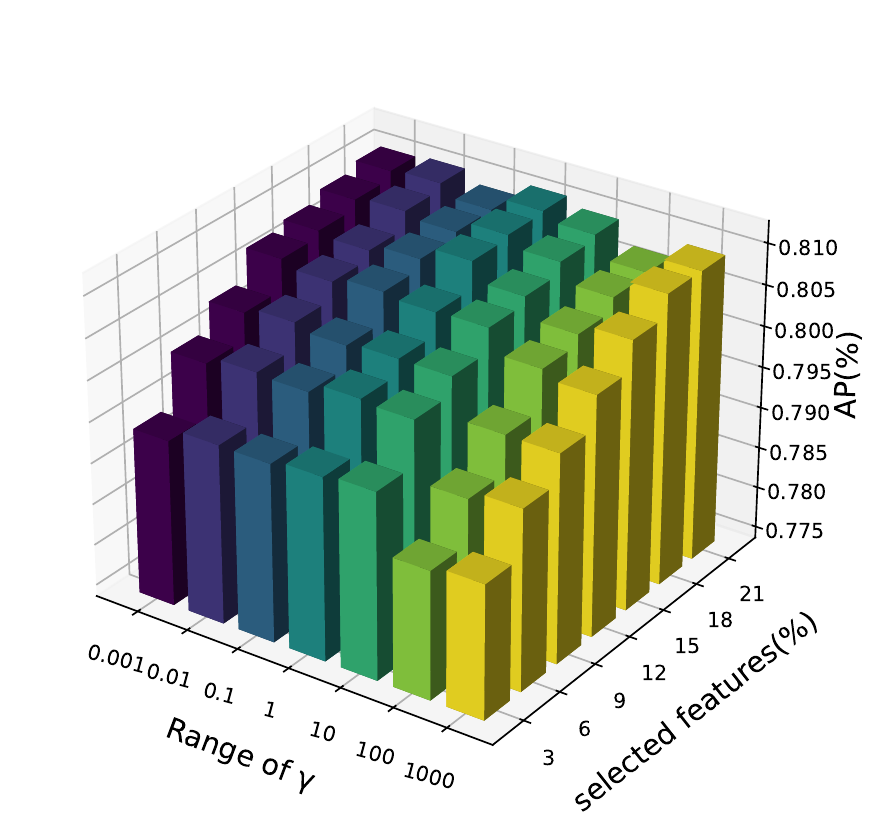}
        \caption{$\gamma$}
    \end{subfigure}
    \hfill
    \begin{subfigure}[b]{0.231\textwidth}
        \centering
        \includegraphics[width=\textwidth]{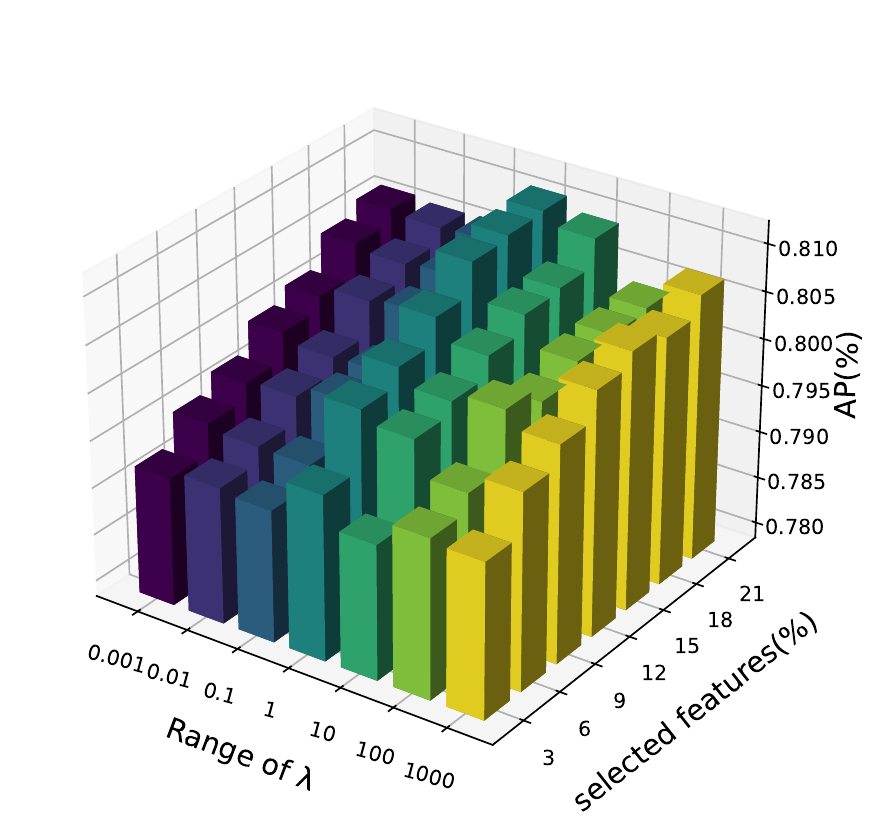}
        \caption{$\lambda$}
    \end{subfigure}
    \hfill
    \begin{subfigure}[b]{0.231\textwidth} 
        \centering
        \includegraphics[width=\textwidth]{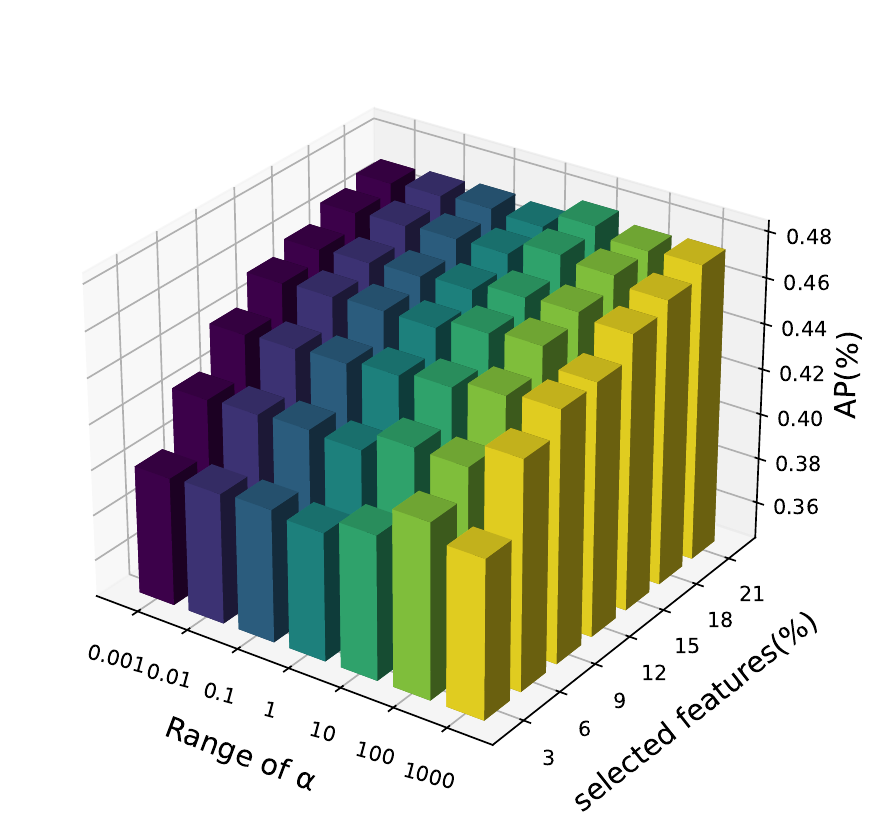}
        \caption{$\alpha$}
    \end{subfigure}
    \hfill
    \begin{subfigure}[b]{0.231\textwidth}
        \centering
        \includegraphics[width=\textwidth]{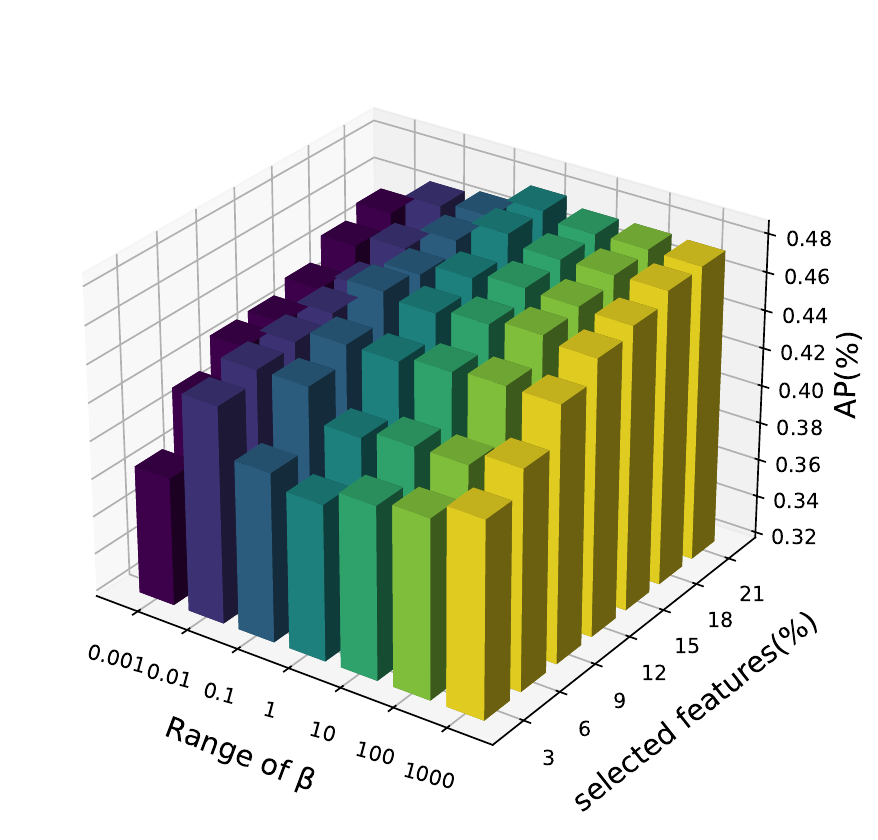}
        \caption{$\beta$}
    \end{subfigure}
    \hfill
    \begin{subfigure}[b]{0.231\textwidth}
        \centering
        \includegraphics[width=\textwidth]{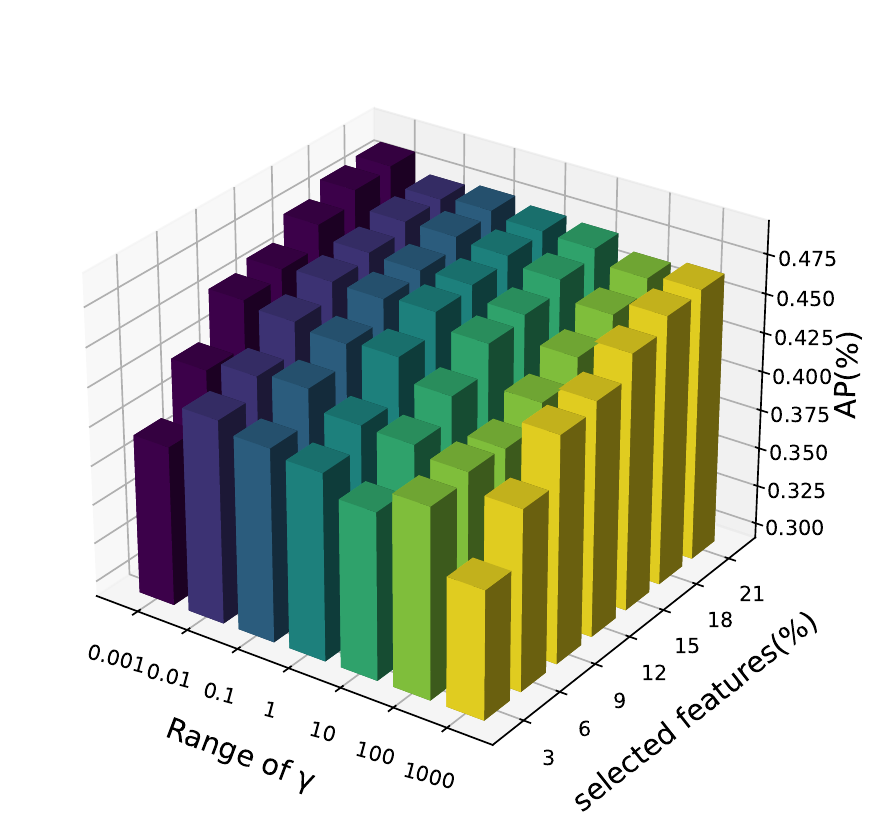}
        \caption{$\gamma$}
    \end{subfigure}
    \hfill
    \begin{subfigure}[b]{0.231\textwidth}
        \centering
        \includegraphics[width=\textwidth]{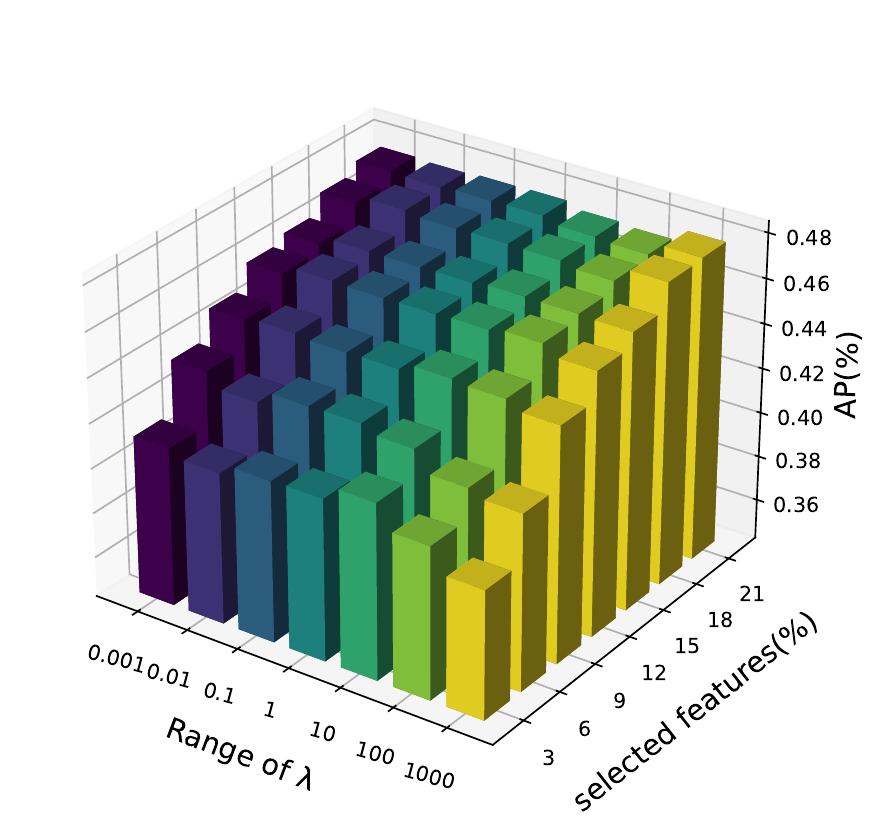}
        \caption{$\lambda$}
    \end{subfigure}
    \caption{Parameter sensitivity analysis of SEHFS on VOC07 (first row), SCENE (second row), and OBJECT (third row), where the horizontal axis indicates the parameter range, the vertical axis shows the number of selected features, and the depth axis represents AP.}
    \label{fig:param}
\end{figure*}
\begin{figure*}[ht]
    \centering
    \hfill
    \begin{subfigure}[b]{0.231\textwidth} 
        \centering
        \includegraphics[width=\textwidth]{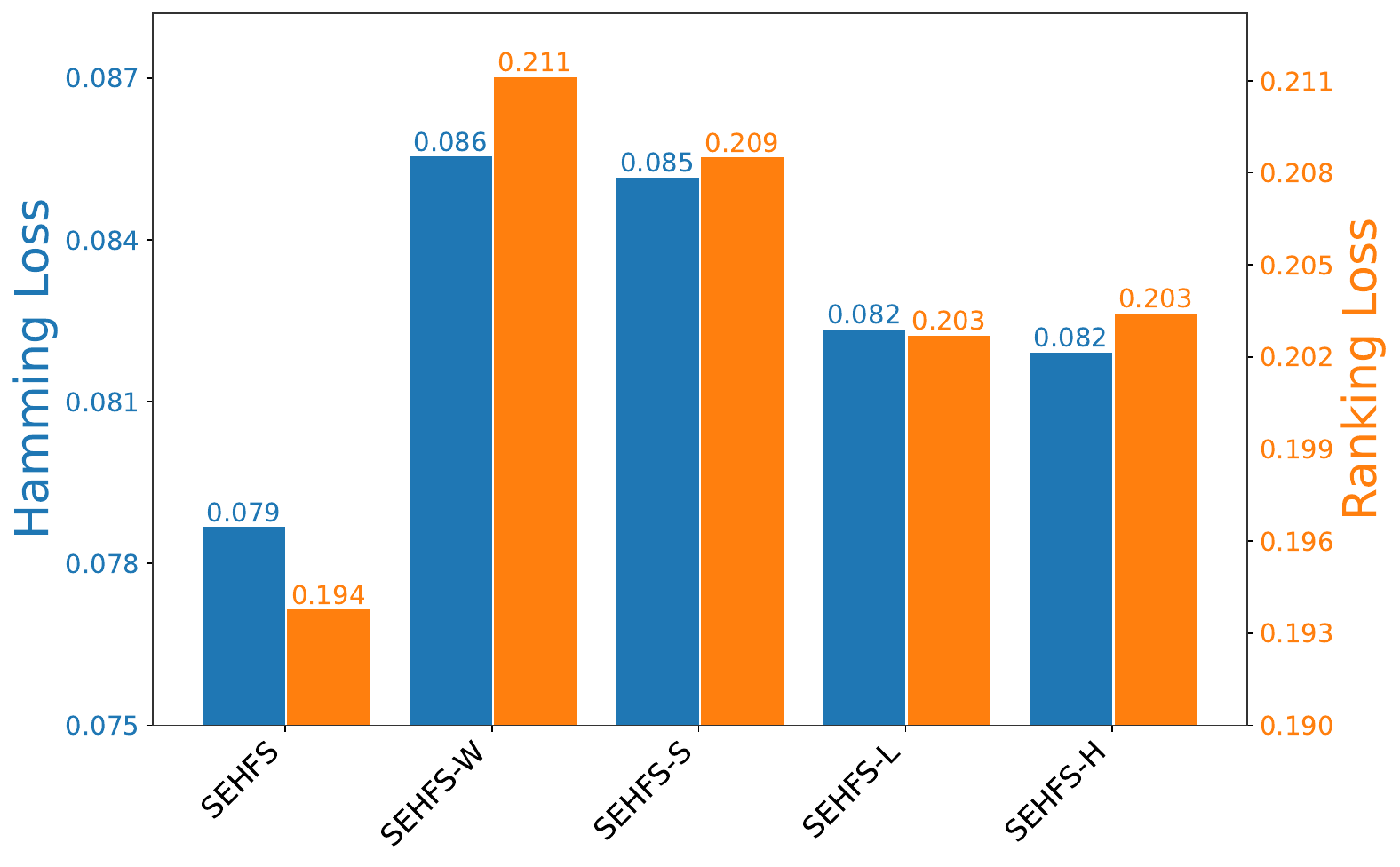}
        \caption{VOC07}
    \end{subfigure}
    \hfill
    \begin{subfigure}[b]{0.231\textwidth}
        \centering
        \includegraphics[width=\textwidth]{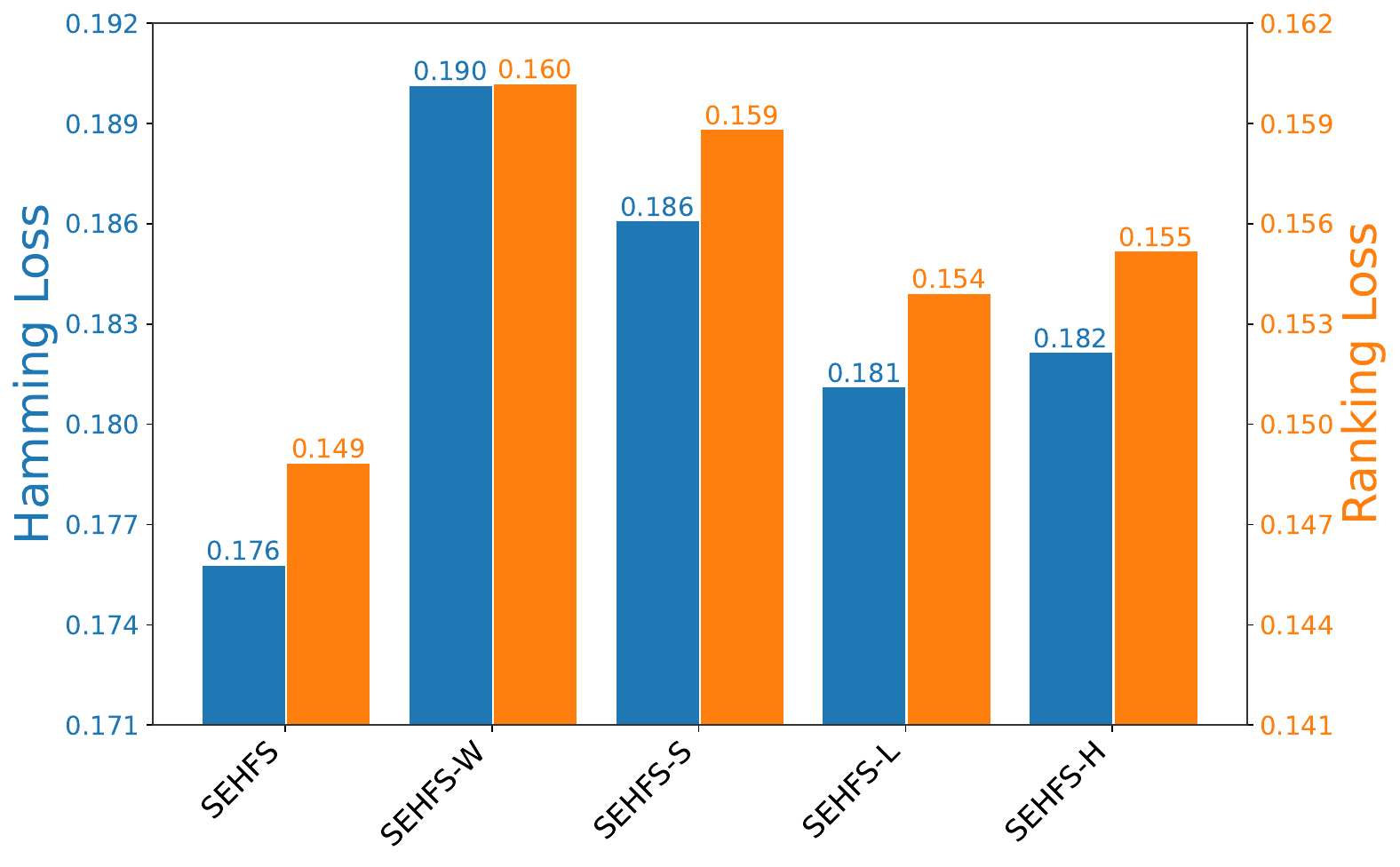}
        \caption{MIRFlickr}
    \end{subfigure}
    \hfill
    \begin{subfigure}[b]{0.231\textwidth} 
        \centering
        \includegraphics[width=\textwidth]{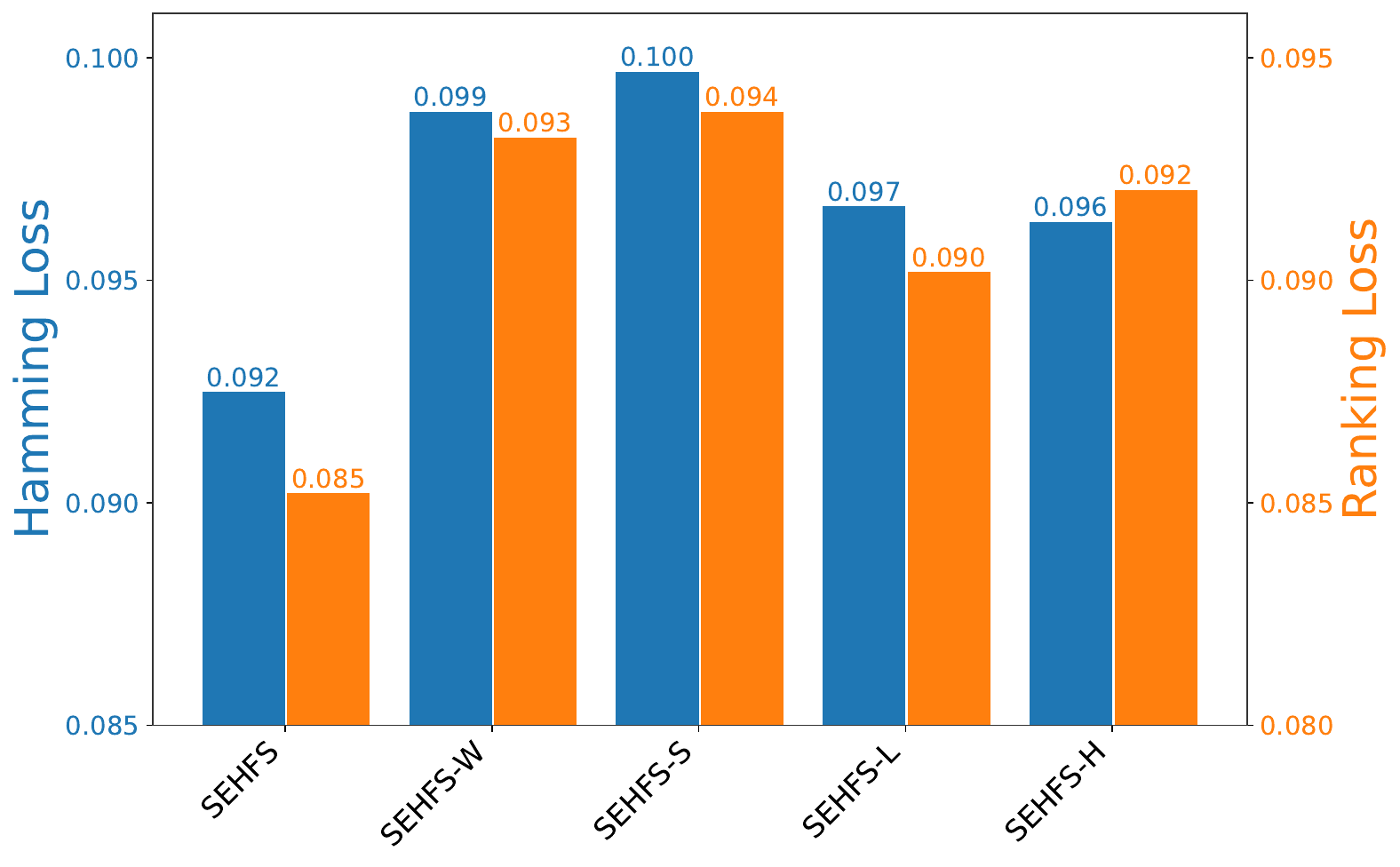}
        \caption{SCENE}
    \end{subfigure}
    \hfill
    \begin{subfigure}[b]{0.231\textwidth}
        \centering
        \includegraphics[width=\textwidth]{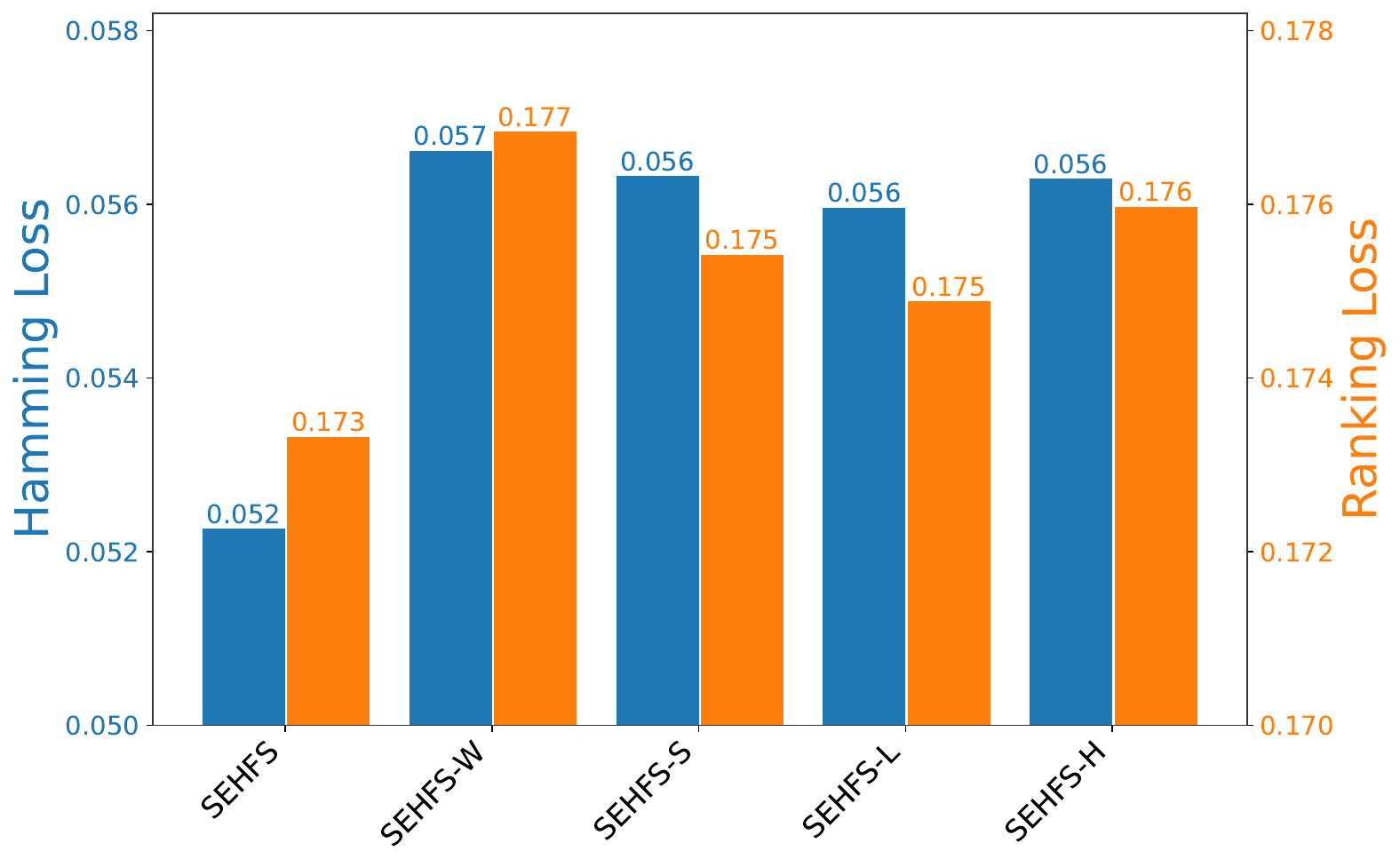}
        \caption{OBJECT}
    \end{subfigure}
    \hfill
    \caption{Ablation study results of SEHFS on VOC07, MIRFlickr, SCENE, and OBJECT, where each subfigure lists SEHFS and its four variants on the horizontal axis and reports the evaluation metric on the vertical axis, with HL on the left and RL on the right.}
    \label{fig:ablation}
\end{figure*}
\begin{figure*}[ht]
    \centering
    \hfill
    \begin{subfigure}[b]{0.231\textwidth} 
        \centering
        \includegraphics[width=\textwidth]{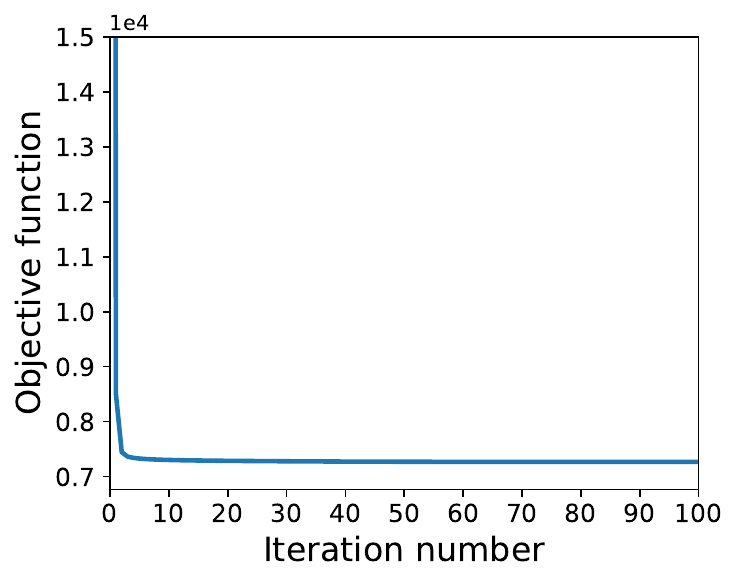}
        \caption{VOC07}
    \end{subfigure}
    \hfill
    \begin{subfigure}[b]{0.231\textwidth}
        \centering
        \includegraphics[width=\textwidth]{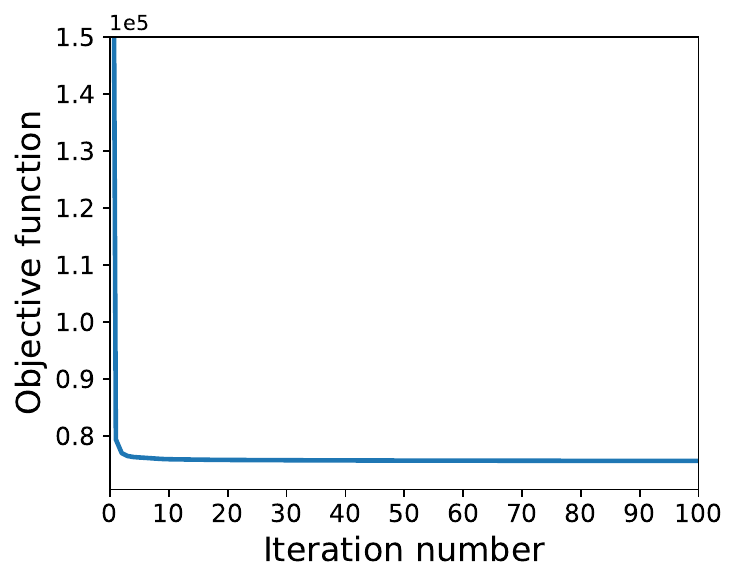}
        \caption{MIRFlickr}
    \end{subfigure}
    \hfill
    \begin{subfigure}[b]{0.231\textwidth} 
        \centering
        \includegraphics[width=\textwidth]{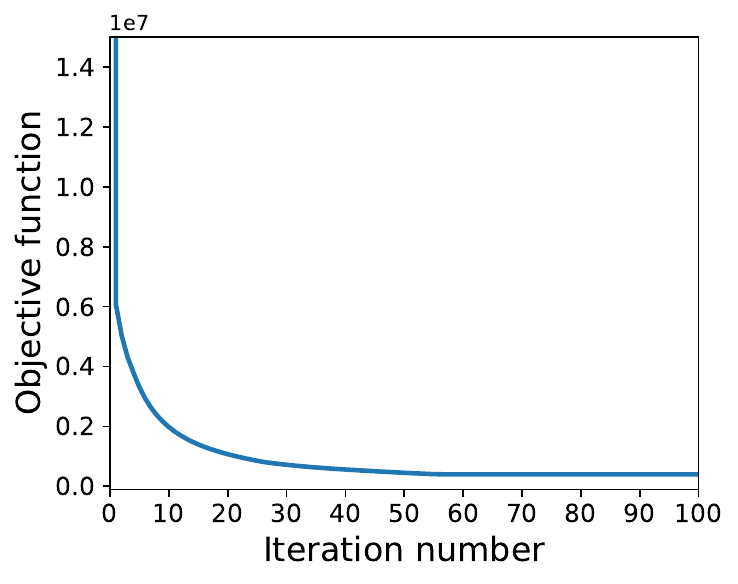}
        \caption{Corel5K}
    \end{subfigure}
    \hfill
    \begin{subfigure}[b]{0.231\textwidth}
        \centering
        \includegraphics[width=\textwidth]{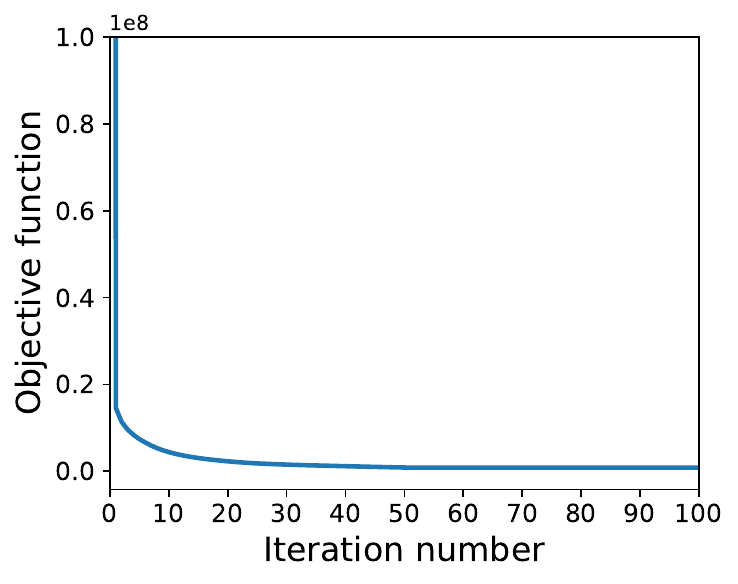}
        \caption{IAPRTC12}
    \end{subfigure}
    \hfill
    \caption{Convergence curve of SEHFS on VOC07, MIRFlickr, Corel5K, and IAPRTC12.}
    \label{fig:con}
\end{figure*}
\subsection{Parameter analysis}
SEHFS comprises four parameters: $\alpha,\,\beta,\,\gamma,\,\lambda$. The influence of each parameter on the performance of SEHFS in terms of the AP metric on VOC07, SCENE, and OBJECT is presented in Figure~\ref{fig:param}. In the figure, the X-axis represents the range of parameters, and the Y-axis represents the percentage of selected features. The results show that SEHFS exhibits low sensitivity to these parameters and that performance remains stable when the proportion of selected features ranges from $3\%$ to $21\%$.

Furthermore, when the weight parameter $\alpha$ of the structural entropy regularization term is increased, SEHFS attains improved performance, as shown in the first column of Figure~\ref{fig:param}. This indicates that the structural entropy regularization term effectively learns high-order feature correlations and that its guidance aligns closely with the optimization objective.

\subsection{Ablation study}
\newtext{To further justify the design rationale of each module, we analyze not only the performance variation of the ablated models, but also the specific information loss caused by removing each component. For this purpose, four variants of the original method are designed: (1) SEHFS-W removes the structural entropy regularization term and directly learns the feature selection matrix $\mathbf{W}$ from the classifier; (2) SEHFS-S discards learning the shared semantic matrix $\mathbf{S}$ and reconstructs the global view matrix solely via the view-specific contribution matrix $\mathbf{H}^v$ of each view; (3) SEHFS-L removes the graph Laplacian regularization term, thereby ignoring the relationship between the shared semantic matrix $\mathbf{S}$ and the label matrix $\mathbf{Y}$; (4) SEHFS-H ignores the view-specific contribution matrix $\mathbf{H}^v$ and learns only the shared semantic matrix $\mathbf{S}$ to reconstruct the global view matrix. The ablation study adopts HL and RL as evaluation metrics on VOC07, MIRFlickr, SCENE, and OBJECT, and the results are presented in Figure~\ref{fig:ablation}.}

\newtext{As shown in Fig.~\ref{fig:ablation}, SEHFS-W exhibits an average degradation of $8.51\%$ in terms of HL relative to the original SEHFS, which confirms that the structural entropy regularization term is essential for redundancy-aware feature organization. SEHFS-L degrades by $7.42\%$, indicating that the graph Laplacian regularization plays an important role in aligning the shared semantic structure with the label manifold and preserving discriminative information in the reconstructed global view.}

\newtext{More importantly, both SEHFS-S and SEHFS-H underperform the original model, with average degradations of $4.94\%$ and $4.81\%$, respectively. This shows that neither shared semantic modeling nor view-specific compensation alone is sufficient for reconstructing an effective global view. The larger degradation of SEHFS-S suggests that $\mathbf{S}$ is the primary carrier of cross-view consensus information, while the degradation of SEHFS-H indicates that $\mathbf{H}^v$ preserves view-specific complementary cues. Overall, $\mathbf{S}$ captures common semantics, $\mathbf{H}^v$ compensates residual view-specific information, and the Laplacian regularization enhances label relevance, jointly yielding a consistent, complementary, and discriminative global view.}

\subsection{Convergence Analysis}
To further demonstrate the convergence property of SEHFS, we also implement experiments on four datasets, VOC07, MIRFlickr, Corel5K, and IAPRTC12. \newtext{Figure~\ref{fig:con} shows the objective value curves of SEHFS during the alternating optimization process. The curves exhibit a clear monotonic decreasing trend and stabilize within a few iterations, which is consistent with the discussion on the convergence property of the optimization algorithm in Section~\ref{subsec:convergence}. These results empirically support that the proposed optimization procedure is numerically stable in practice, although only convergence to a stable point can be guaranteed due to the non-convexity of the overall objective.}

\section{Conclusion}
\label{sec:conclusion}
\newtext{In this paper, we proposed SEHFS, a structural entropy-guided framework for multi-view multi-label feature selection. This method employs structural entropy as a new measure for learning high-order nonlinear feature correlations and reducing redundancy, while fusing information theory and matrix operations to balance global-local generalization. Moreover, SEHFS reconstructs a global view matrix with a shared semantic matrix and view-specific contribution matrices to achieve a balance between consistency and complementarity among views. Experimental results on multiple datasets demonstrate that SEHFS achieves competitive and superior performance compared with seven state-of-the-art methods.}

\newtext{Despite these encouraging results, the current SEHFS framework still has three limitations: (1) since the feature graph $G^f$ is constructed from pairwise mutual information, its characterization of true joint dependency remains indirect; (2) to ensure optimization tractability, SEHFS adopts a fixed three-layer encoding tree, which may limit its ability to capture richer hierarchical structures; (3) the current framework assumes relatively complete and reliable view and label information, while real-world multi-view multi-label data often involve missing views and noisy labels. In future work, we will explore more effective structural representations for multivariate dependencies and extend SEHFS to more challenging multi-view multi-label scenarios with incomplete and noisy data.}

\newpage
\bibliographystyle{IEEEtran}
\bibliography{IEEEref}

\begin{IEEEbiography}[{\includegraphics[width=1in,height=1.25in,clip,keepaspectratio]{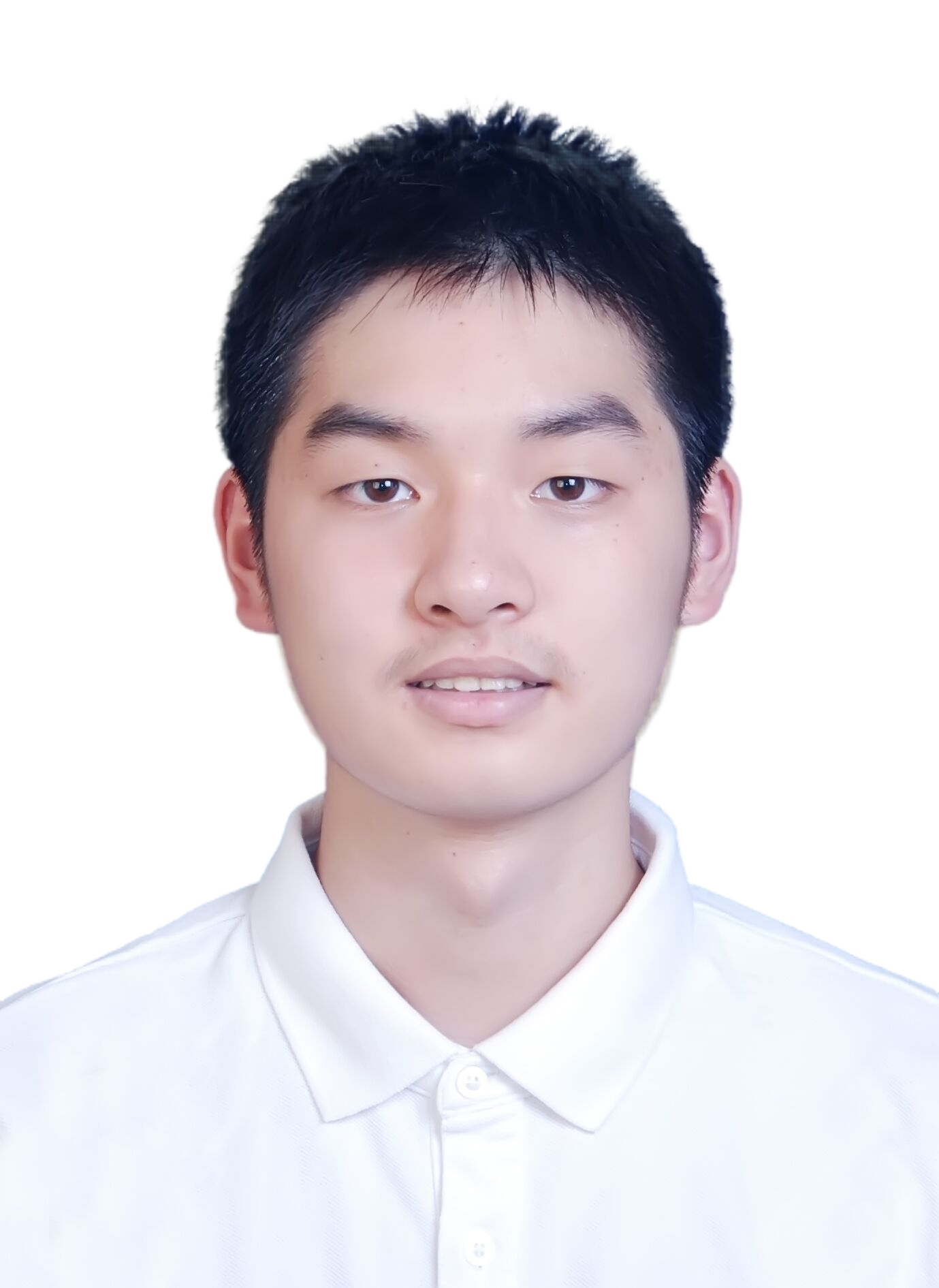}}]{Cheng Peng} is currently working toward his Bachelor's degree in the College of Software, Jilin University, Changchun, China. His main research interests include feature selection, multi-label learning, multi-view learning, and information theory.
\end{IEEEbiography}
\begin{IEEEbiography}[{\includegraphics[width=1in,height=1.25in,clip,keepaspectratio]{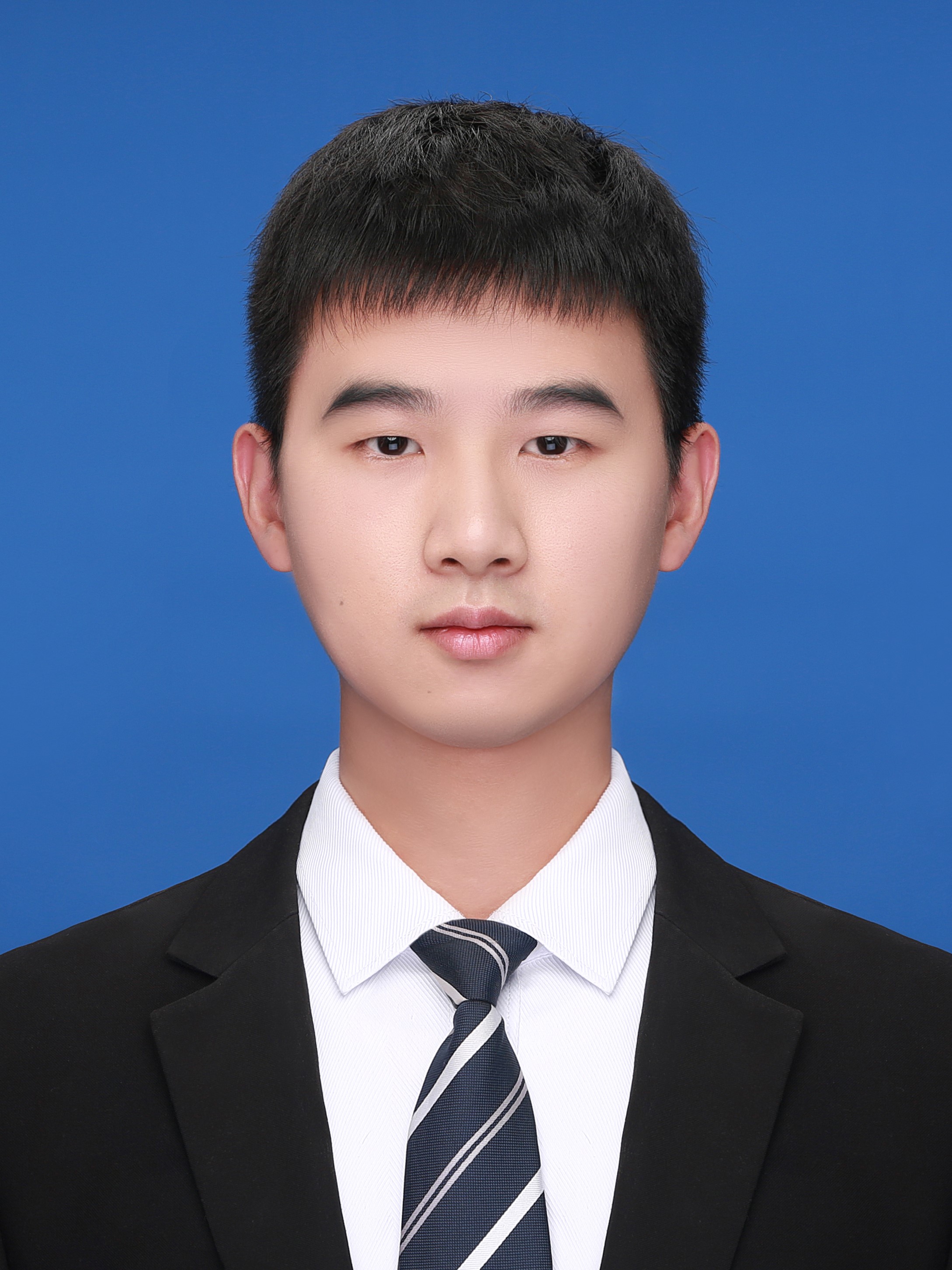}}]{Yonghao Li} received the Ph.D. degree from the College of Computer Science and Technology, Jilin University, in 2023. He joined the School of Computing and Artificial Intelligence, Southwestern University of Finance and Economics (SWUFE) in 2024. His research interests include feature selection, causal learning and federated learning. He has published more than 30 papers in renowned journals/conferences such as IEEE Transactions on Knowledge and Data Engineering, IEEE Transactions on Neural Networks and Learning Systems, IEEE Transactions on Artificial Intelligence, Pattern Recognition, Information Fusion, Knowledge and Information Systems, Information Processing \& Management, KDD and IJCAI. He also serves as a reviewer for several journals and conferences, including TKDE, TNNLS, TKDD, TMM, TSMC, TCYB, TCSVT, TFS, TETCI, TCE, Pattern Recognition, Information Processing \& Management, CVPR, ECCV, NeurIPS, AAAI and ACM MM, etc.
\end{IEEEbiography}
\begin{IEEEbiography}[{\includegraphics[width=1in,height=1.25in,clip,keepaspectratio]{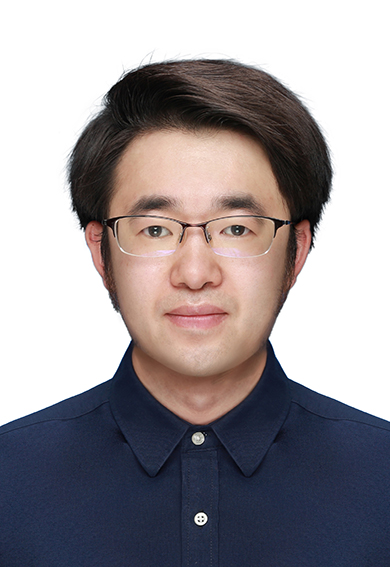}}]{Wanfu Gao}(Senior Member, IEEE) received the B.S. and Ph.D. degrees from the College of Computer Science and Technology, Jilin University, Changchun, China, in 2013 and 2019, respectively. He completed his postdoctoral research at the School of Artificial Intelligence, Jilin University, under the supervision of Prof. Yi Chang. He is an Associate Professor with the College of Computer Science and Technology, Jilin University. He has published more than 60 papers on feature selection, including SIGIR, ACL, AAAI, IJCAI, IEEE Transactions on Systems, Man, and Cybernetics: Systems, IEEE Transactions on Neural Networks and Learning Systems, IEEE Transactions on Artificial Intelligence, and Pattern Recognition. His research interests include feature selection, multi-label learning, and information theory. He serves on the program committees and as a reviewer for NeurIPS, ICML, AAAI, IJCAI, ICDM, TKDE, TNNLS, IEEE TCYB, and other conferences/journals. Dr. Gao received the 2019 ACM Changchun Doctoral Dissertation Award and the Postdoctoral Innovative Talents Support Program and was selected as a Top 2\% worldwide scientist.
\end{IEEEbiography}
\begin{IEEEbiography}[{\includegraphics[width=1in,height=1.25in,clip,keepaspectratio]{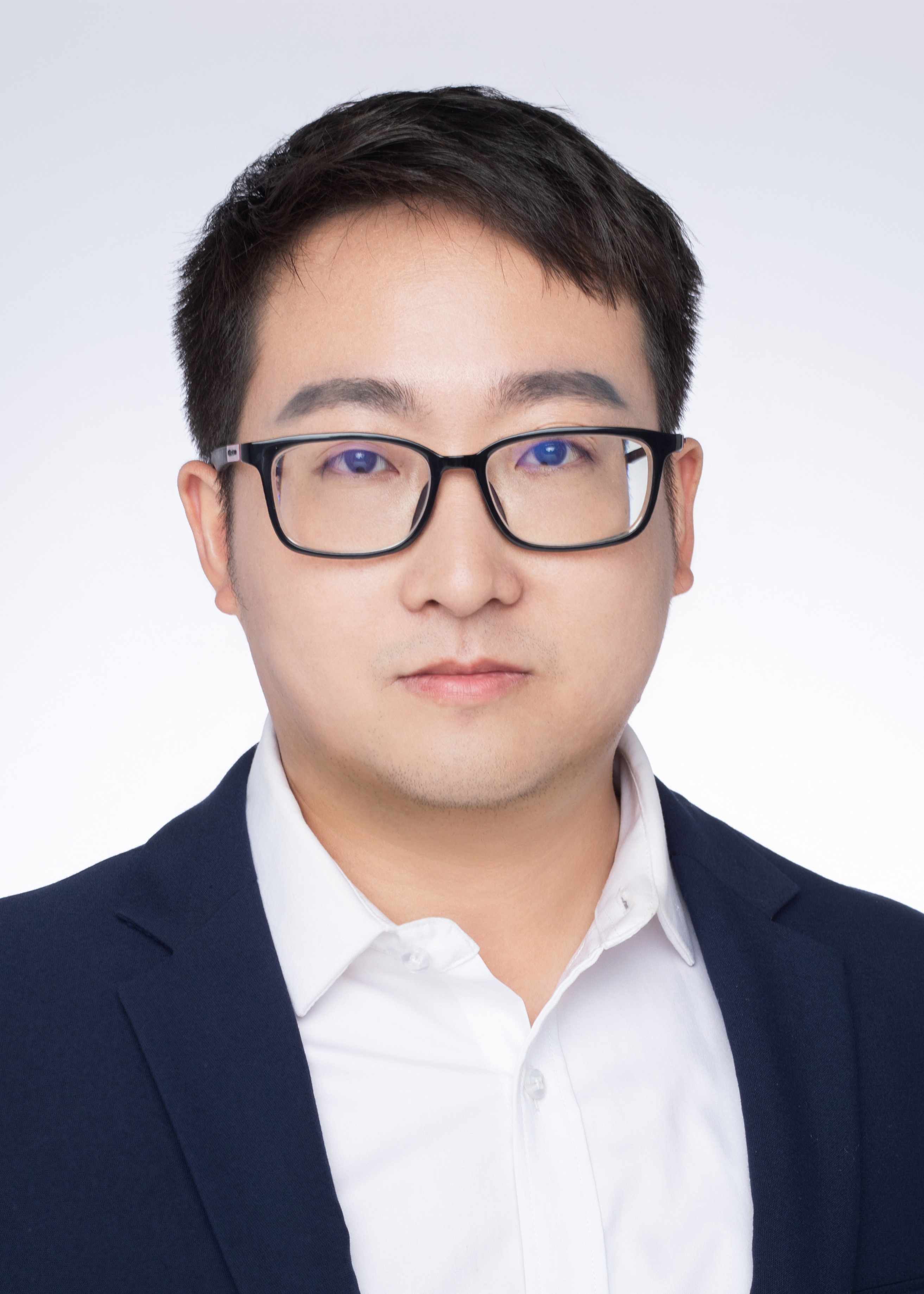}}]{Jie Wen} (Senior Member, IEEE) received the Ph.D. degree in Computer Science and Technology at Harbin Institute of Technology, Shenzhen in 2019. He is currently an Associate Professor at the School of Computer Science and Technology, Harbin Institute of Technology, Shenzhen. His research interests include image and video enhancement, pattern recognition, and machine learning. He has authored or co-authored more than 100 technical papers at prestigious international journals and conferences. He serves as an \textbf{Associate Editor} of \textit{IEEE Transactions on Pattern Analysis and Machine Intelligence}, \textit{IEEE Transactions on Image Processing}, \textit{IEEE Transactions on Information Forensics and Security}, \textit{IEEE Transactions on Circuits and Systems for Video Technology}, \textit{Pattern Recognition}, and \textit{International Journal of Image and Graphics}, an \textbf{Area Editor} of \textit{Information Fusion}. He is on the \textbf{Editorial Board} of \textit{Discover Artificial Intelligence} and \textit{Mathematics}, and \textbf{Youth Editorial Board} of \textit{CAAI Transactions on Intelligence Technology}. He also served as the \textbf{Area Chair} of \textit{NeurIPS}, \textit{ICLR}, \textit{ACM MM}, and \textit{ICML}, as well as the SPC of \textit{AAAI} and \textit{IJCAI}. He was selected for the `World's Top 2\% Scientists List' in 2021-2025. One paper received the `distinguished paper award’ from AAAI’23. For more information, please refer to the homepage: https://sites.google.com/view/jerry-wen-hit/home.
\end{IEEEbiography}
\begin{IEEEbiography}[{\includegraphics[width=1in,height=1.25in,clip,keepaspectratio]{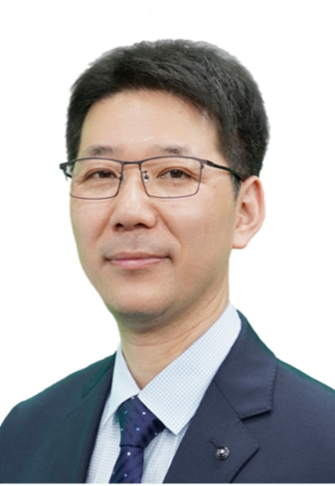}}]{Weiping Ding}(Senior Member, IEEE) received the Ph.D. degree in Computer Science, Nanjing University of Aeronautics and Astronautics, Nanjing, China, in 2013. From 2014 to 2015, he was a Postdoctoral Researcher at the Brain Research Center, National Chiao Tung University, Hsinchu, Taiwan, China. In 2016, he was a Visiting Scholar at National University of Singapore, Singapore. From 2017 to 2018, he was a Visiting Professor at University of Technology Sydney, Australia. His main research directions involve deep neural networks, granular data mining, and multimodal machine learning. He ranked within the top 2\% Ranking of Scientists in the World by Stanford University (2020-2025). He has published over 480 articles, including over 280 IEEE Transactions papers. His twenty-five authored/co-authored papers have been selected as ESI Highly Cited Papers. He served as an Associate Editor/Area Editor/Editorial Board member of more than 10 international prestigious journals, such as IEEE Transactions on Neural Networks and Learning Systems, IEEE Transactions on Fuzzy Systems, IEEE/CAA Journal of Automatica Sinica, IEEE Transactions on Emerging Topics in Computational Intelligence, IEEE Transactions on Intelligent Transportation Systems, IEEE Transactions on Artificial Intelligence, Information Fusion, Information Sciences, Neurocomputing, Applied Soft Computing, Engineering Applications of Artificial Intelligence, Swarm and Evolutionary Computation, et al.
\end{IEEEbiography}

\end{document}